\documentclass{article}

\usepackage{arxiv}
\usepackage[utf8]{inputenc}
\usepackage[T1]{fontenc}
\usepackage{hyperref}
\usepackage{url}
\usepackage{booktabs}
\usepackage{amsmath,amssymb}
\usepackage{amsfonts}
\usepackage{nicefrac}
\usepackage{microtype}
\usepackage{cleveref}
\usepackage{graphicx}
\usepackage{natbib}
\usepackage{doi}
\usepackage{array}
\usepackage{float}
\usepackage{placeins}
\usepackage[algo2e,ruled]{algorithm2e}
\newcolumntype{P}[1]{>{\raggedright\arraybackslash}p{#1}}

\graphicspath{{figures/}}
\title{Response Renormalization for Critical Deep Equilibrium Models}
\author{Jose Luis Lima de Jesus Silva\\
Federal University of Bahia, Department of Geophysics\\
Salvador, BA 40170-115, Brazil\\
Grupo de Estudos e Aplica\c{c}\~ao de Intelig\^encia Artificial em Geof\'isica (GAIA)\\
Federal University of Bahia, Salvador, BA 40170-115, Brazil\\
\texttt{jseluis.silva@gmail.com}}
\date{}

\renewcommand{\shorttitle}{Response Renormalization for Critical DEQs}
\hypersetup{
  colorlinks=true,
  linkcolor=black,
  citecolor=black,
  urlcolor=black,
  pdftitle={Response Renormalization for Critical Deep Equilibrium Models},
  pdfauthor={Jose Luis Lima de Jesus Silva},
  pdfkeywords={deep equilibrium models, implicit differentiation, response theory, singular values, renormalization}
}

\newcommand{\R}{\mathbb R}
\newcommand{\cL}{\mathcal L}

\newcommand{\diag}{\operatorname{diag}}

\newcommand{\ii}{\mathrm i}

\begin{document}
\raggedbottom

\maketitle

\begin{abstract}
Deep Equilibrium Models (DEQs) compute predictions by finding a hidden
representation that remains unchanged under the model's update. Training
through this equilibrium uses implicit differentiation, which requires solving
an adjoint system built from the residual Jacobian. If this Jacobian is nearly
singular along directions to which the loss is sensitive, small perturbations
can be strongly amplified in the adjoint response, producing large and highly
sensitive gradients that can make optimization unreliable. We introduce
Response Renormalization, a backward-pass framework that lifts selected
near-pole denominators while leaving unlifted response channels unchanged.
Collective Mode Response Renormalization (CMR) applies this correction in a
low-dimensional critical subspace, while Phi-adaptive CMR computes a bounded
response mass from a prescribed positive susceptibility rule. We derive dense
and matrix-free collective formulations, distinguish exact gradients of a
modified frozen-anchor residual from backward-response surrogates, and extend
the construction to Structured Implicit Layers and Vector Attractors (SILVA).
Across 23 multiphysics families spanning partial differential equations,
three-dimensional fields, operator maps, complex geometries, and particle
systems, models trained with CMR and Phi-CMR have test errors no more than five
percent higher than those trained with exact implicit differentiation in more
than 98\% of static and 95\% of transient family--seed comparisons.
Solver-index experiments additionally show finite responses converging toward
the static adjoint, while physical-time rollouts retain predictive fidelity
under the evaluated conditions. These results demonstrate that selective response
renormalization can control
near-critical adjoint amplification without globally damping well-conditioned
sensitivity. Therefore, the method can make parameter updates more reliable
while preserving the useful gradient information needed for learning.
\end{abstract}


\begingroup
\linespread{0.94}\selectfont

\section{Introduction}
\label{sec:introduction}

Modern implicit layers replace a finite computation graph by the solution of an
equation, and Deep Equilibrium Models (DEQs) instantiate this idea through the
hidden fixed point \(z^\star=f_\theta(z^\star,x)\). Effective depth is therefore
determined by a solver rather than by an explicitly stored stack of layers
\citep{bai2019deep}, placing DEQs within the broader implicit deep learning
framework of \citet{elghaoui2021implicit}. The same fixed-point principle now supports
multiscale vision models \citep{bai2020multiscale}, steady-state neural
operators for partial differential equation (PDE) solution maps
\citep{marwah2023steady}, and equilibrium algorithmic reasoners for iterative
computation \citep{georgiev2024algorithmic}, making implicit layers attractive
because useful features can emerge from many recurrent interactions without
declaring a finite depth in advance.

This modeling advantage creates a training bottleneck because fixed-point
learning still depends on a linearized sensitivity problem whose conditioning
can dominate parameter updates. Existing approaches use Jacobian penalties
\citep{bai2021stabilizing}, monotone networks \citep{winston2020monotone},
initialization analysis \citep{agarwala2022sensitive}, phantom gradients
\citep{geng2021training}, standardized TorchDEQ solvers
\citep{geng2023torchdeq}, or reversibility \citep{mccallum2025reversible}, but
do not directly control the inverse response while preserving informative
equilibrium sensitivities.

The unresolved issue in this setting is geometric because implicit
differentiation avoids storing the forward solver trajectory by replacing
unrolled backpropagation with the adjoint equation, whose conditioning depends
on how the residual Jacobian stretches, compresses, and rotates directions in
the hidden-state space.
Let \(z^\star\) denote the equilibrium hidden state for input \(x\),
\(f_\theta\) the model update with parameters \(\theta\), and \(\mathcal L\) the
loss. The residual Jacobian \(K\), equilibrium-state loss gradient \(g\), and
adjoint \(v\) satisfy
\begin{equation}
  K^\top v=g,\qquad
  K=I-\partial_z f_\theta(z^\star,x),\qquad
  g=\nabla_{z^\star}\cL .
  \label{eq:adjoint}
\end{equation}
In the singular-value decomposition \(K=U\Sigma V^\top\), the columns \(u_i\)
of \(U\) and \(v_i\) of \(V\) are the left and right singular vectors, and the
diagonal entries \(\sigma_i\) of \(\Sigma\) are the singular values. The inverse
response is
\begin{equation}
  v=U\Sigma^{-1}V^\top g
  =
  \sum_i \frac{v_i^\top g}{\sigma_i}u_i,
  \label{eq:modal-adjoint}
\end{equation}
Equation~\eqref{eq:modal-adjoint} identifies \(v_i^\top g\) as the loss-source
projection onto mode \(i\) and \(1/\sigma_i\) as its amplification factor. For
a non-normal linearized operator with \(JJ^\top\neq J^\top J\),
eigenvalues do not capture this source-pole geometry because distinct input and
output directions can strongly amplify perturbations even when the eigenvalues appear well behaved
\citep{trefethen2005spectra}. The same geometry connects DEQ training to
classical ill-posed inverse problems and numerical linear algebra, since
pseudoinverses make singular linear responses well defined
\citep{benisrael2003generalized}, Tikhonov regularization gives a finite inverse
by filtering every singular direction with a common regularization parameter
\citep{tikhonov1963solution}, and
Krylov methods such as the generalized minimal residual method (GMRES) make
nonsymmetric adjoint solves practical without forming dense matrices
\citep{saad1986gmres}. In this DEQ setting, the resulting finite adjoint
response may also attenuate informative long-range sensitivity together with
the near-singular component.

We introduce Response Renormalization by identifying \(G_0(0)=K^{-1}\) as the
zero-frequency response of the discrete fixed-point trajectory and
\(v=G_0(0)^\top g\) as its adjoint action to control loss-visible near-singular
amplification while preserving the remaining response directions.
Appendix~\ref{app:silva-transient},
Eqs.~\eqref{eq:solver-transient-linearization}--\eqref{eq:physical-rollout-metric},
derives the finite-horizon limit and separates solver index from physical time.
Response theory names the loss gradient the
source, small singular denominators response poles, and a controlled
denominator change a counterterm \citep{martin1973statistical}. The
source-response constructions of \citet{janssen1976lagrangean} and
\citet{dedominicis1976techniques}, together with their use in analyses of neural
dynamics \citep{schuecker2016functional}, motivate this terminology. Applied to
a DEQ, they provide a finite and explicit procedure that identifies loss-aligned
singular response directions and modifies only their unstable denominators,
while retaining explicit forms in \(K\), \(K^\top v=g\), singular vectors,
source projections, and modified denominators.

Response Renormalization aims to preserve the original implicit response in
stable directions while making only source-visible critical response
finite. It implements this objective through Collective Mode Response
Renormalization (CMR), a pole-selective denominator lift, and through
Phi-adaptive CMR (Phi-CMR) and Delta-Phi (\(\Delta\Phi\)) structured
finite-response and source-conditioned gates, with dense and matrix-free forms
distinguishing exact from surrogate gradients. We test selectivity through
controlled mechanisms and source rotations, learned physical prediction
through Darcy and native SILVA experiments, and transfer across equations and
representations through 13 public datasets.\footnote{The extension includes
PDEArena \citep{gupta2023pdearena} (Navier--Stokes 2D, Shallow Water 2D,
Maxwell 3D, and Kuramoto--Sivashinsky 1D), DynaBench
\citep{dulny2023dynabench} (Advection, Burgers, Gas Dynamics,
Kuramoto--Sivashinsky, Reaction--Diffusion, and Wave), PDEGym
\citep{herde2024poseidon} (Poisson--Gauss), CFDBench
\citep{luo2023cfdbench} (Cylinder Geometry), and LagrangeBench
\citep{toshev2023lagrangebench} (Taylor--Green Vortex 2D).} Consistent with this objective, the
CMR and Phi-CMR error ratios are at most 1.05 in all 65 static comparisons
and in 59 and 57 of 60 horizon-seven comparisons, respectively.

\section{Methods}
\label{sec:method}

Response Renormalization treats the DEQ backward pass as a controlled response
in which, at equilibrium \(z^\star\in\R^d\), the adjoint propagates loss source
\(g\in\R^d\) through the transpose inverse of the residual Jacobian. By leaving
stable channels exact and lifting only resolved critical denominators, the
construction recovers the original implicit adjoint when no critical response
is present. Appendix~\ref{app:implicit-derivation},
Eqs.~\eqref{eq:forward-sensitivity-scalar}--\eqref{eq:compatibility-condition},
derives the adjoint and singular compatibility condition, and
Appendix~\ref{app:response-field} derives the causal response construction.

Vectors are columns, \(A^\top\) denotes transpose, \(\|q\|_2\) the Euclidean
norm, \(\diag(q_i)\) a diagonal matrix, and \([q]_+=\max\{q,0\}\). All SVDs use
the fixed state representation and train-only normalization, while
Appendix~\ref{app:coordinate-metric} gives the coordinate-aware metric form. Indices
\(n\), \(i\), and \(j\) label solver steps, full singular modes, and computed
low-rank modes. Euclidean interval projection is denoted by
\(\mathcal P_{[\ell,u]}(q)=\arg\min_{y\in[\ell,u]}\tfrac12(y-q)^2\), with its
piecewise form and the full symbol and index registry in Appendix~\ref{app:notation},
Table~\ref{tab:notation}.

\paragraph{Response poles in DEQ adjoints.}
Let \(z_n\in\R^d\) be the hidden state at solver index \(n\), let \(x\) be the
input, and let \(\theta\) contain the trainable parameters. After adding a
small source \(h_n\in\R^d\), write \(z_n=z^\star+\phi_n\) and define
\(J=\partial_zf_\theta(z^\star,x)\in\R^{d\times d}\). Linearization and a
discrete Fourier transform give
\begin{align}
  z_{n+1}&=f_\theta(z_n,x)+h_n,
  \label{eq:sourced-iteration}\\
  \phi_{n+1}&=J\phi_n+h_n,
  \label{eq:linearized-iteration}\\
  G_0(\omega)&=(e^{\ii\omega}I-J)^{-1},
  \qquad
  G_0(0)=(I-J)^{-1}=K^{-1}.
  \label{eq:frequency-response}
\end{align}
Here \(\phi_n\) is the equilibrium displacement, \(\omega\in[-\pi,\pi]\) is
frequency, and the bare propagator \(G_0(\omega)\) maps the transformed source
\(h(\omega)\) to \(\phi(\omega)\). Thus the residual Jacobian \(K=I-J\) is the
inverse static response. The loss source
\(g=\nabla_{z^\star}\cL\in\R^d\) enters \(K^\top v=g\), and the adjoint
\(v\in\R^d\) converts state sensitivity into every parameter derivative. In
particular,
\begin{equation}
  v=G_0(0)^\top g=K^{-\top}g.
  \label{eq:adjoint-response}
\end{equation}
Appendix Eqs.~\eqref{eq:appendix-sourced-iteration}--\eqref{eq:appendix-zero-frequency-response}
and Eqs.~\eqref{eq:forward-sensitivity-scalar}--\eqref{eq:parameter-gradient-adjoint}
give the Taylor, Fourier, and adjoint substitutions, including why one adjoint
solve replaces one direct solve per parameter, while response-field
theory supplies this source-response language
\citep{martin1973statistical,janssen1976lagrangean,dedominicis1976techniques}.
For a DEQ, \(g\) is the source, \(K^{-1}\) is the response operator, and the
singular values of \(K\) are its denominators. Appendix~\ref{app:response-field}
connects the same DEQ derivatives
\(F^{(p)}=\partial_z^p f_\theta(z^\star,x)\) to direct perturbations, the
generating functional, Feynman expansion, Dyson dressing, and the finite
counterterm. Equations~\eqref{eq:appendix-static-source-series}--\eqref{eq:appendix-cmr-matching-condition}
then project these insertions onto the singular channels of \(K\). The method
implements the resulting finite denominator through
Eq.~\eqref{eq:main-counterterm}, without numerically summing diagrams.

For the singular-value decomposition \(K=U\Sigma V^\top\), columns \(u_i\)
and \(v_i\) are output and source directions, while
\(\Sigma=\diag(\sigma_1,\ldots,\sigma_d)\) has
\(0\leq\sigma_1\leq\cdots\leq\sigma_d\). Equation~\eqref{eq:modal-adjoint}
therefore assigns channel \(i\) the source amplitude \(v_i^\top g\), gain
\(1/\sigma_i\), and output direction \(u_i\), with its exact, regularized, and
singular limits given by Appendix
Eqs.~\eqref{eq:compatibility-condition}--\eqref{eq:baseline-exact}.

The following non-normal example shows why eigenvalues alone do not contain
this source-to-response geometry,
\begin{equation}
  J=\begin{pmatrix}1-\varepsilon_{\rm gap}&M\\0&1-\varepsilon_{\rm gap}\end{pmatrix},
  \qquad
  K=I-J=\begin{pmatrix}\varepsilon_{\rm gap}&-M\\0&\varepsilon_{\rm gap}\end{pmatrix},
  \label{eq:nonnormal-example}
\end{equation}
the spectral gap \(\varepsilon_{\rm gap}>0\) fixes both eigenvalues at
\(1-\varepsilon_{\rm gap}\), while the scalar non-normal coupling \(M\)
creates the response scale \(M\varepsilon_{\rm gap}^{-2}\) in \(K^{-1}\), a
susceptibility exposed by the singular vectors \citep{trefethen2005spectra}.

\paragraph{Response-renormalization construction.}
At threshold \(\kappa\), the critical set
\(\mathcal C=\{i\mid\sigma_i<\kappa\}\) contains modes with potentially large
exact gains \(1/\sigma_i\), whereas all remaining gains are bounded by
\(1/\kappa\). A critical mode is loss-visible
if \(v_i^\top g\neq0\), and affects parameter \(a\) only if additionally
\(u_i^\top B_a\neq0\). CMR obtains this
partition from the singular-value decomposition of \(K\)
\citep{golub1965calculating} and chooses positive effective denominators
\(\sigma_i^{\rm eff}\) only for \(i\in\mathcal C\).
Appendix~\ref{app:baseline-equations} derives all responses in common singular
coordinates and separates the exact modified-forward gradient from the
backward surrogate.
\begin{equation}
  \sigma_i^{\rm eff} =
  \begin{cases}
  \sigma_i, & i\notin\mathcal C,\\
  \max(\sigma_i,m_i), & i\in\mathcal C,
  \end{cases}
  \qquad m_i>0.
  \label{eq:cmr-denominator}
\end{equation}
Here \(\kappa\) is the critical resolution, \(m_i\) the response mass, and
\(r_\kappa=|\mathcal C|\) the active pole rank. Dense evaluation finds
\(\mathcal C\) directly, while matrix-free evaluation requests \(r\geq
r_\kappa\) smallest triplets. With
\(U_{\mathcal C}=[u_i]_{i\in\mathcal C}\) and
\(V_{\mathcal C}=[v_i]_{i\in\mathcal C}\), the response lift is
\begin{align}
  \delta_i&=\sigma_i^{\rm eff}-\sigma_i,\qquad
  \Delta K_{\mathcal C}
  =U_{\mathcal C}\diag(\delta_i)V_{\mathcal C}^\top,
  \label{eq:main-counterterm}\\
  R^R_\theta(z,x\mid\mathcal A)
  &=z-f_\theta(z,x)+\Delta K_{\mathcal C}(z-z_{\rm ref}),
  \qquad K^R=\partial_zR^R_\theta=K+\Delta K_{\mathcal C},
  \label{eq:main-renormalized-residual}\\
  (K^R)^\top v^R&=g,\qquad
  \nabla_\theta^R\cL=\partial_\theta\cL+
  (\partial_\theta f_\theta)^\top v^R.
  \label{eq:main-renormalized-gradient}
\end{align}
Here \(\mathcal A=(z_{\rm ref},U_{\mathcal C},V_{\mathcal C},\delta)\) is
held fixed during one update. At \(z_{\rm ref}\), substitution of
Eq.~\eqref{eq:main-counterterm} into Eq.~\eqref{eq:main-renormalized-residual}
gives \(K^R=U\Sigma_{\rm eff}V^\top\), and hence
\begin{equation}
  v_{\rm CMR}=U\diag((\sigma_i^{\rm eff})^{-1})V^\top g.
  \label{eq:cmr-adjoint}
\end{equation}
Equations~\eqref{eq:main-counterterm}--\eqref{eq:main-renormalized-gradient}
define the modified DEQ and its exact frozen-anchor gradient, whereas applying
only Eq.~\eqref{eq:cmr-adjoint} at the original equilibrium gives the
separately evaluated surrogate. Appendix
Eqs.~\eqref{eq:appendix-cmr-counterterm}--\eqref{eq:appendix-cmr-substitution-chain}
give the finite operator matching, while
Eqs.~\eqref{eq:objective-consistent-adjoint}--\eqref{eq:appendix-modified-parameter-gradient}
give the exact and surrogate boundaries, and
Algorithm~\ref{alg:deq-cmr-exact} implements the complete construction.
For comparison, generalized inverses define the singular reference
\citep{benisrael2003generalized}, while Tikhonov regularization
\citep{tikhonov1963solution} uses
\begin{equation}
  F_\mu(\sigma)=\frac{\sigma}{\sigma^2+\mu^2},
  \label{eq:tikhonov-filter}
\end{equation}
Stable-critical filtering leaves the exact gain on stable modes and applies
ridge only within \(\mathcal C\), providing a critical-only ridge comparator.
CMR instead lifts each critical denominator and leaves stable response exact.

If \(\sigma_i^{\rm eff}\geq m_{\min}>0\) on \(\mathcal C\) and
\(\sigma_i\geq\kappa\) off \(\mathcal C\), then
\begin{equation}
  \|v_{\rm CMR}\|_2
  \leq
  \left(
  \sum_{i\notin\mathcal C}\frac{|v_i^\top g|^2}{\sigma_i^2}
  +
  \sum_{i\in\mathcal C}\frac{|v_i^\top g|^2}{m_{\min}^2}
  \right)^{1/2}.
  \label{eq:cmr-bound}
\end{equation}
The stable-sector response is unchanged, while the induced bias remains
confined to the critical sector,
\begin{equation}
  v_{\rm CMR}-v
  =
  \sum_{i\in\mathcal C}
  (v_i^\top g)
  \left((\sigma_i^{\rm eff})^{-1}-\sigma_i^{-1}\right)u_i.
  \label{eq:cmr-bias}
\end{equation}

\paragraph{Collective and matrix-free algorithm.}
Dense singular-value decompositions (SVDs) \citep{golub1965calculating} expose
the full-state mechanism, whereas a structured specialization acts on a small
collective denominator when a local--global factorization is available. For
\begin{equation}
  K=K_L-AB^\top,
  \label{eq:woodbury-form}
\end{equation}
with inexpensive \(K_L^{-1}\) products and low-rank \(A,B\), the Woodbury
identity gives \citep{woodbury1950inverting}
\begin{equation}
  K^{-1}=K_L^{-1}+K_L^{-1}A(I-B^\top K_L^{-1}A)^{-1}B^\top K_L^{-1}.
  \label{eq:woodbury}
\end{equation}
The small matrix
\begin{equation}
  \Gamma=I-B^\top K_L^{-1}A
  \label{eq:collective-denominator}
\end{equation}
is the collective denominator whose critical singular values CMR lifts. Its
singular values need not equal those of the full \(K\) because Woodbury only
localizes the low-rank correction. The rank-one identity of
\citet{sherman1950adjustment} is the one-mode case.
Jacobian-vector products (JVPs), vector-Jacobian products (VJPs), and the
generalized minimal residual method (GMRES) \citep{saad1986gmres} then require
only \(z^\star\), products with \(K\) and \(K^\top\), and a low-rank critical
subspace.
Appendix~\ref{app:matrix-free} derives the matrix-free products and collective
adjoint in Eq.~\eqref{eq:appendix-woodbury-adjoint}.
Appendix~\ref{app:algorithm} gives
Algorithms~\ref{alg:cmr-backward}--\ref{alg:silva-cmr-exact}, covering the
conceptual pass, full DEQ procedure, and Structured Implicit Layers and Vector
Attractors (SILVA) local--global specialization
\citep{silva2026silvanetworks}. Dense or matrix-free low-rank estimates
\citep{golub1965calculating,lehoucq1998arpack} and bounded parameters
\(m_0,\kappa,r,\alpha_{\max},\lambda,c_{\max}\) complete the numerical specification.

\paragraph{Phi-CMR and Delta-Phi.}
CMR still needs a finite response scale, which fixed CMR declares directly and
Phi-CMR parameterizes through a finite-response target,
\begin{equation}
  G_i(0)=\chi_R,
  \qquad
  m_i^\Phi=\frac{1}{\chi_R},
  \qquad
  \delta m_i^\Phi=[m_i^\Phi-\sigma_i]_+.
  \label{eq:phi-law}
\end{equation}
Here \(G_i(0)\), \(\chi_R>0\), \(m_i^\Phi=\chi_R^{-1}\), and
\(\delta m_i^\Phi\) are the channel response, target susceptibility, inverse
response mass, and nonnegative lift above \(\sigma_i\), respectively.
The positive monotone per-mode and collective parameterizations used in the experiments
are
\begin{align}
  m_i^\Phi
  &=m_0\left(1+\frac{m_0}{\sigma_i+m_0}\right),
  \qquad m_0\leq m_i^\Phi\leq2m_0,
  \label{eq:main-phi-mode-law}\\
  m_{\mathcal C}^\Phi
  &=m_0\left[1+(\alpha_{\max}-1)p_{\mathcal C}\right],
  \qquad
  p_{\mathcal C}=\mathcal P_{[0,1]}\!\left(
  \frac{\kappa-\sigma_{\mathcal C}}{\kappa}\right),
  \label{eq:main-phi-collective-law}
\end{align}
where \(\sigma_{\mathcal C}=\min_{i\in\mathcal C}\sigma_i\),
\(p_{\mathcal C}\in[0,1]\), \(m_0>0\), and \(\alpha_{\max}\geq1\) are the
smallest selected denominator, pole pressure, base mass, and maximum
multiplier. For a structured operator, \(\sigma_{\mathcal C}\) is the smallest
selected singular value of \(\Gamma\). These are prescribed positive DEQ
parameterizations, not unique derivations or automatically learned corrections.
Appendix~\ref{app:phi-derivation} gives their Hartree-style motivation
\citep{baym1962self,cornwall1974effective} and proves positivity, boundedness,
and monotonicity.

Delta-Phi adds the source geometry that CMR uses diagnostically through a gate
that measures the critical source fraction \(a_{\mathcal C}\), pole pressure
\(p_{\mathcal C}\), and signed source gate \(s_{\mathcal C}\) before projecting
the response mass onto a positive admissible interval.
\begin{align}
  a_{\mathcal C}
  &=
  \frac{\sum_{i\in\mathcal C}(v_i^\top g)^2}
       {\sum_{i=1}^d(v_i^\top g)^2+\varepsilon_{\rm den}},
  \label{eq:source-critical-fraction}\\
  p_{\mathcal C}
  &=\mathcal P_{[0,1]}\!\left(\frac{\kappa-\sigma_{\mathcal C}}{\kappa}\right),
  \qquad
  s_{\mathcal C}=\mathcal P_{[-1,1]}(2a_{\mathcal C}-1),\nonumber\\
  m^{\Delta\Phi}
  &=
  \mathcal P_{[m_0,c_{\max}m_0]}\!\left(
  m^\Phi + m_0\lambda p_{\mathcal C}s_{\mathcal C}
  \right).
  \label{eq:delta-phi}
\end{align}
\begin{sloppypar}
Here \(a_{\mathcal C}\in[0,1]\), \(s_{\mathcal C}\in[-1,1]\),
\(\lambda\geq0\), and \(c_{\max}\geq1\) are critical source energy, signed
activation, correction strength, and upper-mass multiplier, while
\(\varepsilon_{\rm den}>0\) prevents division by zero. This bounded
source-conditioned rule
replaces a free multilayer perceptron (MLP), and
Appendix~\ref{app:delta-phi-derivation} derives its source-conditioned
potential, bounds, and activation limits.
The per-mode formulation uses
\(m_i=\max(m_i^\Phi,m^{\Delta\Phi})\), while the collective formulation uses the
scalar \(m^{\Delta\Phi}\) directly. The per-mode field formulation is therefore
lift-only relative to Phi-CMR, whereas the collective formulation can reduce the Phi
mass toward the positive floor when the source avoids the pole. In either case
Eq.~\eqref{eq:cmr-denominator} turns the selected mass into the effective
denominator.
Appendix~\ref{app:quantitative-metrics} defines the metrics and parameters,
and Algorithms~\ref{alg:deq-cmr-exact} and~\ref{alg:silva-cmr-exact} give the
per-mode DEQ and scalar collective SILVA procedures.
\end{sloppypar}

\linespread{0.86}\selectfont

\begin{figure}[!t]
\centering
\includegraphics[width=0.66\textwidth]{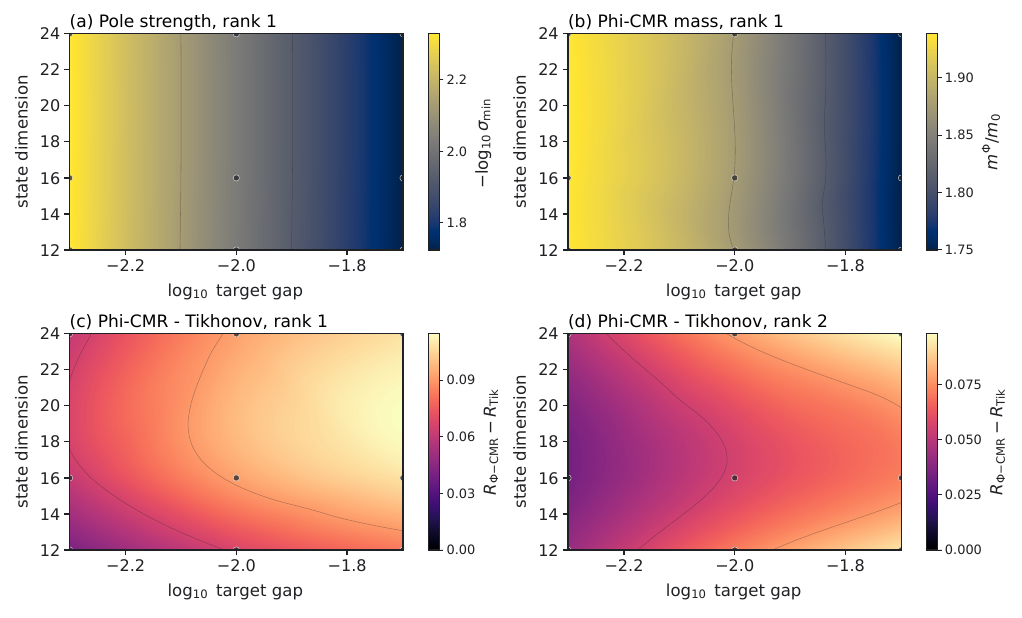}
\small
\caption{Pole-response maps with (a) dense pole strength
\(-\log_{10}\sigma_{\min}\), (b) the Phi-CMR mass multiplier, (c)
\(R_{\Phi{\rm -CMR}}-R_{\rm Tik}\) at critical rank one, and (d) the same
difference at critical rank two over target gap and state dimension.}
\label{fig:pole-maps}
\end{figure}

\section{Results and Discussion}
\label{sec:experiments}

\begin{figure}[!t]
\centering
\includegraphics[width=0.70\textwidth]{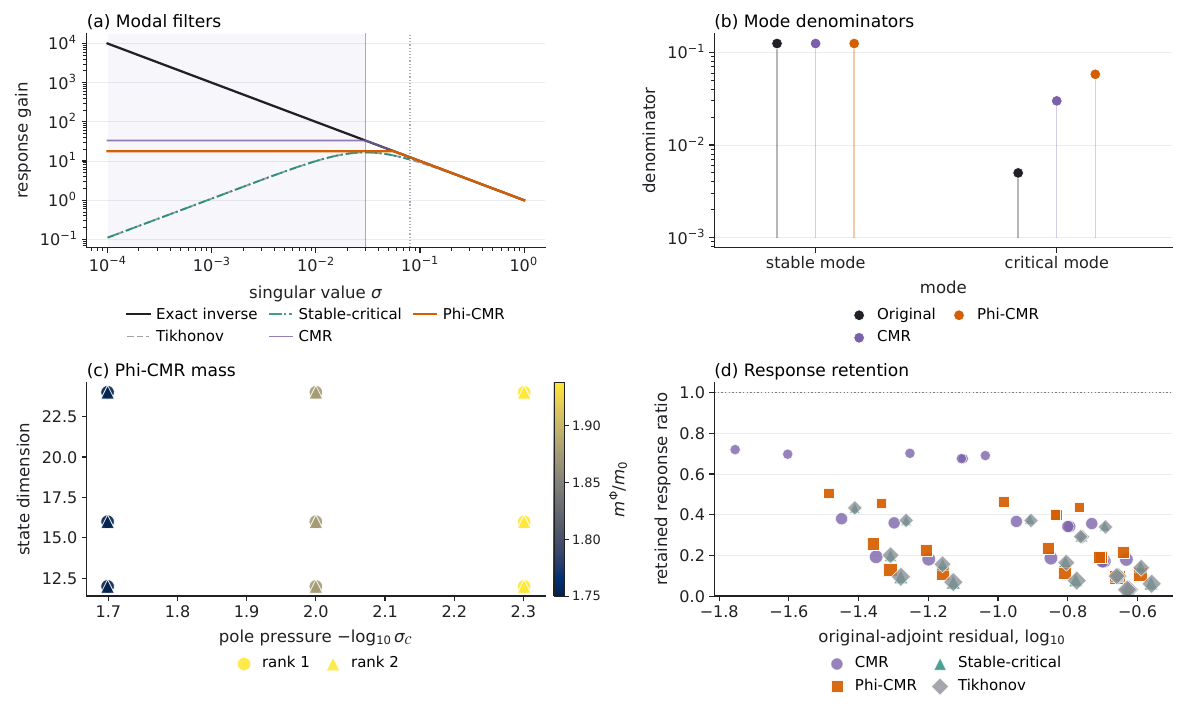}
\small
\caption{Selective renormalization of the near-pole response with (a) exact and
filtered response amplification versus singular value, (b) original and
renormalized denominators for stable and critical modes, (c) Phi-CMR mass versus
pole pressure by state dimension and critical rank, and (d) retained response
versus residual against the original adjoint equation.}
\label{fig:mechanism}
\end{figure}

The experiments test whether pole strength predicts the adjoint while a
selective lift preserves resolved stable response, whether source rotation
activates the effect through \(v_i^\top g\), and whether transfer retains fidelity through
physical-time rollouts and new representations. Table
\ref{tab:main-quantitative-evidence} quantifies the first two mechanisms, with
Phi-CMR improving amplitude and direction over ridge on the collective grid and
CMR improving norm ratio and relative error under source alignment.
Appendix~\ref{app:quantitative-metrics} and
Table~\ref{tab:protocol-grid} give the metrics, protocols, extended results,
and validity conditions, while native SILVA links the controlled tests to
physical data. Together they evaluate directional selectivity and predictive
fidelity for deliberately biased backward interventions.
\begin{table}[H]
\centering
\scriptsize
\setlength{\tabcolsep}{3.2pt}
\caption{Directional dense-adjoint evidence. Entries are mean $\pm$ standard deviation over configurations. $R_m$ is the norm ratio, $C_m$ is directional cosine, $P_m$ is signed projection onto the exact adjoint, and $E_m$ is relative adjoint error.}
\label{tab:main-quantitative-evidence}
\begin{tabular}{@{}llrcccc@{}}
\toprule
Block & Method & $n$ & $R_m$ & $C_m$ & $P_m$ & $E_m$\\
\midrule
Collective grid & Phi-CMR & 36 & 0.303 $\pm$ 0.239 & 0.947 $\pm$ 0.061 & 0.290 $\pm$ 0.234 & 0.726 $\pm$ 0.202\\
Collective grid & Tikhonov & 36 & 0.231 $\pm$ 0.250 & 0.815 $\pm$ 0.191 & 0.206 $\pm$ 0.245 & 0.812 $\pm$ 0.208\\
Source rotation & CMR & 48 & 0.471 $\pm$ 0.283 & 0.999 $\pm$ 0.001 & 0.471 $\pm$ 0.283 & 0.530 $\pm$ 0.283\\
Source rotation & Tikhonov & 48 & 0.242 $\pm$ 0.307 & 0.973 $\pm$ 0.055 & 0.241 $\pm$ 0.308 & 0.760 $\pm$ 0.308\\
Held-out gate & Delta-Phi & 20 & 0.440 $\pm$ 0.351 & 0.999 $\pm$ 0.002 & 0.440 $\pm$ 0.351 & 0.560 $\pm$ 0.351\\
Held-out gate & Phi-CMR & 20 & 0.432 $\pm$ 0.350 & 0.997 $\pm$ 0.005 & 0.432 $\pm$ 0.350 & 0.569 $\pm$ 0.350\\
\bottomrule
\end{tabular}
\end{table}
The first sweep tests Eq.~\eqref{eq:modal-adjoint} against the exact inverse,
ridge \citep{tikhonov1963solution}, stable-critical filtering, CMR, and
Phi-CMR. In Fig.~\ref{fig:pole-maps}, yellow at the smallest gaps in panel (a)
marks the \(1/\sigma_i\) pole, while the matching yellow in panel (b) is the
bounded rise of \(m_{\mathcal C}^\Phi\) prescribed by
Eq.~\eqref{eq:main-phi-collective-law}. The positive dark-to-yellow fields in
panels (c,d) show greater response retention than Tikhonov at critical ranks
one and two, and Table~\ref{tab:main-quantitative-evidence} confirms that this
gain maintains higher mean directional alignment with the exact adjoint than
Tikhonov. Appendix~\ref{app:extra-figures}, Fig.~\ref{fig:rank2-pole-extra},
provides the rank-two pole and mass maps.

Figure~\ref{fig:mechanism} summarizes a controlled singular-spectrum experiment
at target gaps \(0.02\), \(0.01\), and \(0.005\), state dimensions 12, 16, and
24, and critical ranks one and two, comparing how each backward
method treats stable and near-singular modes. In panel (a), the exact inverse diverges as the singular value
approaches zero, Tikhonov suppresses that direction toward zero, and CMR and
Phi-CMR instead cap it at a finite nonzero value before recovering the exact
response outside the critical region. Panel (b) makes this selectivity explicit
because the critical denominator increases from 0.005 to 0.030 under CMR and
0.0581 under Phi-CMR, while the stable denominator remains fixed at 0.125.
Panel (c) visualizes the prescribed dependence of Phi-CMR mass on pole pressure,
and its consistency across dimensions and critical ranks verifies that the
prescribed response law does not explicitly depend on model size. Panel (d) then
shows that CMR and Phi-CMR retain more of the exact response than
stable-critical filtering and Tikhonov as a function of residual against the
original equation after the deliberate lift, while the renormalized-equation
residual stays at solver tolerance, confirming that the displacement records
the intended modification rather than solver failure.
Together, panels (a--d) show that Response Renormalization removes singular
amplification without globally damping the backward signal, consistent with
Eq.~\eqref{eq:cmr-bound} and Table~\ref{tab:main-quantitative-evidence}.

\FloatBarrier

\begin{figure}[!t]
\centering
\includegraphics[width=0.58\textwidth]{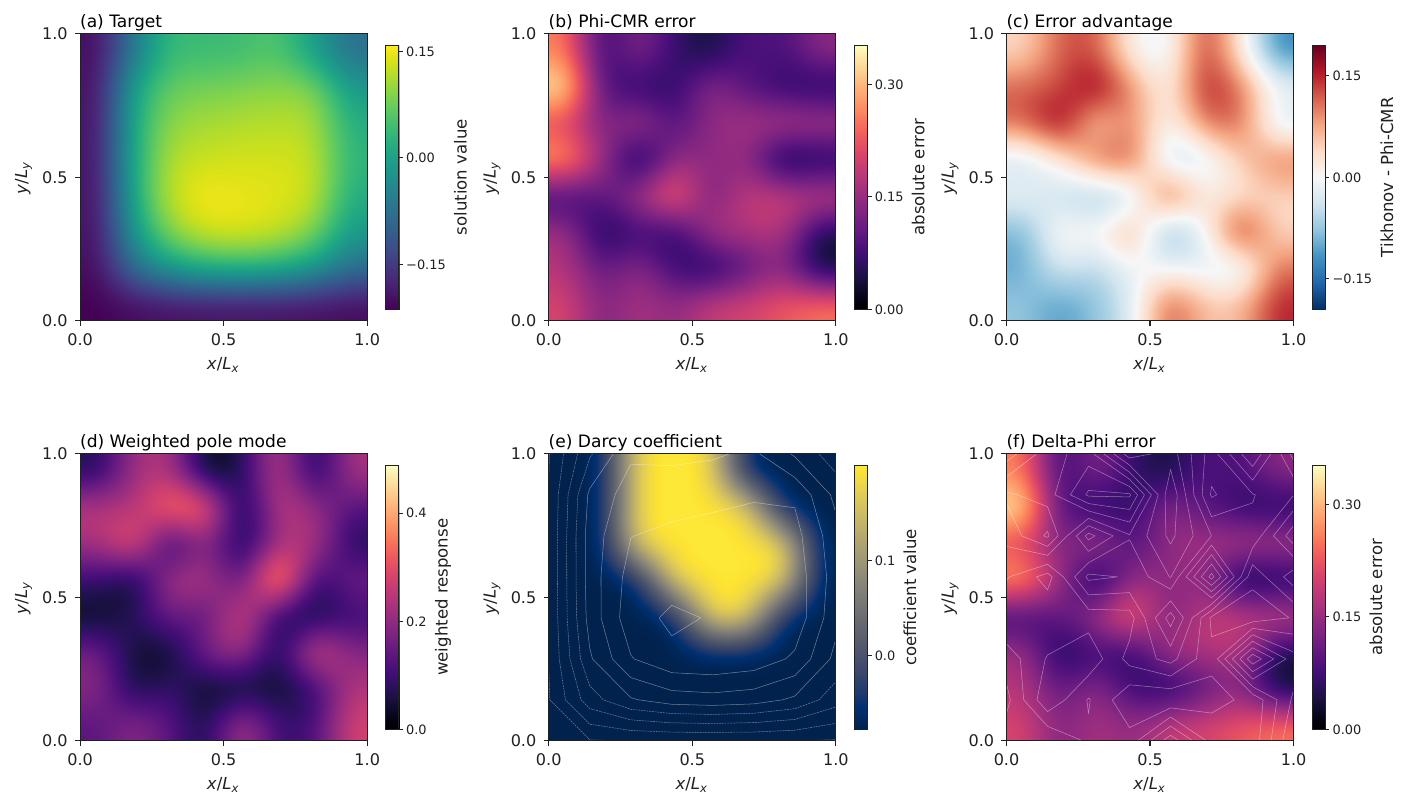}
\scriptsize
\caption{Darcy field bridge with (a) target field, (b) Phi-CMR absolute error,
(c) Tikhonov-minus-Phi-CMR absolute-error difference, (d) source-weighted pole
mode, (e) coefficient field with solution contours, (f) Delta-Phi absolute
error with pole contours.}
\label{fig:darcy-atlas}
\end{figure}

At fixed operator, the gain over Tikhonov increases with \(v_i^\top g\), and the
controlled source intervention directly tests this loss visibility (Appendix
Fig.~\ref{fig:source-falsification}). Structured holdouts further identify pole
geometry for CMR, monotone mass for Phi-CMR, and bounded source activation for
Delta-Phi as the necessary components, whereas free multilayer perceptrons
drift off-axis (Appendix Figs.~\ref{fig:delta-phi-extra}
and~\ref{fig:delta-ablation-extra}).

Figure~\ref{fig:darcy-atlas} connects the modal law to a physical field by
aligning the solution and coefficient in panels (a,e), showing that Phi-CMR error is
spatially localized, and panel (c) uses red for locations where Tikhonov error
is larger and blue where Phi-CMR error is larger. Panel (d) visualizes the
source visibility \(|v_i^\top g|\) applied to the near-pole output structure in
Eq.~\eqref{eq:modal-adjoint}, while the contours in panel (f) relate the bounded
correction in Eq.~\eqref{eq:delta-phi} to that geometry. The corresponding
relative errors are 1.21 for Phi-CMR and Delta-Phi, 1.31 for CMR, 1.56 for
Tikhonov, and 1.58 for stable-critical filtering. The response--suppression
analysis places CMR and Phi-CMR between exact response and stronger spectral
damping, confirming selective rather than global attenuation
(Table~\ref{tab:extended-quantitative-evidence},
Figs.~\ref{fig:source-falsification}--\ref{fig:delta-phi-extra}). Errors above
the unit zero-prediction reference indicate a shared forward approximation
limit, whereas spatial differences identify the source-aligned backward effect
tested in Appendix~\ref{app:silva-forward-capacity}.

The same response law transfers to SILVA \citep{silva2026silvanetworks}
on PDEBench and The Well \citep{takamoto2022pdebench,ohana2024well}.
Appendix~\ref{app:native-silva-details} gives the matched design, baseline
roles, and shared forward protocol, with CMR and Phi-CMR alone modifying the
small collective denominator while Delta-Phi remains the source-conditioned
ablation. Figure~\ref{fig:silva-multiphysics} and
Table~\ref{tab:silva-main-summary} quantify fidelity, activation, convergence,
and cost: CMR has the lowest one-step error, JFB the lowest cost, and Phi-CMR
lower observed error than Tikhonov in 27 of 50 paired comparisons.

\begin{table}[!t]
\centering
\scriptsize
\setlength{\tabcolsep}{3.5pt}
\caption{All native SILVA baselines over ten one-step families and nine dynamic
families. Errors and time are normalized by exact implicit differentiation.}
\label{tab:silva-main-summary}
\begin{tabular}{@{}lccccc@{}}
\toprule
Method & One-step error & Ratio $\leq1.05$ & Time & Horizon-8 error & Ratio $\leq1.05$\\
\midrule
Implicit & 1.000 & 50/50 & 1.00 & 1.000 & 45/45\\
Phantom & 0.988 & 48/50 & 0.65 & 0.997 & 43/45\\
Neumann & 1.015 & 43/50 & 0.58 & 1.007 & 42/45\\
JFB & 1.011 & 43/50 & 0.52 & 0.988 & 42/45\\
SHINE & 1.060 & 35/50 & 0.55 & 1.052 & 31/45\\
Tikhonov & 1.011 & 47/50 & 2.49 & 1.016 & 43/45\\
CMR & 0.986 & 49/50 & 2.03 & 0.986 & 45/45\\
Phi-CMR & 1.039 & 48/50 & 2.02 & 1.025 & 43/45\\
\bottomrule
\end{tabular}
\end{table}

Appendix~\ref{app:silva-transient} distinguishes and derives the solver-index
and physical-time limits, while the Phi-CMR error ratio is at most 1.05 in 43
of 45 horizon-eight comparisons and its finite
causal-response error decreases
from \(0.411\) at one solver step to \(2.50\times10^{-4}\) at 64 steps without
assigning the modified backward endpoint a new PDE time scale.

Across 13 further datasets, the Phi-CMR error ratio is at most 1.05 in all 65
static and 57 of 60 horizon-seven comparisons, with sparse activation in every
dataset (Appendix~\ref{app:silva-cross-suite}). It also lowers error in all 30
high-error capacity cases (Appendix~\ref{app:silva-forward-capacity}), and
Appendices~\ref{app:extended-discussion-boundary} and
\ref{app:silva-family-atlas} give the validity conditions and field--pole
realizations.

\section{Conclusion}
Response Renormalization formulates near-critical DEQ/SILVA sensitivity as a
source-visible pole problem by decomposing the adjoint into local and
low-dimensional collective terms. CMR lifts only critical denominators,
Phi-CMR selects their positive masses, and Delta-Phi bounds source conditioning.
The implicit response is recovered without a source-visible critical mode,
while dense CMR preserves resolved stable singular channels of \(K\) and SILVA
preserves unlifted \(\Gamma\) modes. Controlled source rotations verify the
predicted \(v_i^\top g\) dependence, which the Darcy field bridge localizes
against the shared forward error.

Pole sweeps, source rotations, and eight-method comparisons establish
directional fidelity, predictive preservation, cost, and solver-index
convergence across equations and representations. Capacity scaling associates
part of predictive error with forward capacity, while joint adjoint--rollout
analyses distinguish solver from PDE time. Source-visible activation and
\((\sigma_{\mathcal C},a_{\mathcal C},\rho_R)\) respectively diagnose backward
criticality and its pole pressure, loss coupling, and response consistency.

Together, these results define an operating regime where modifying only
loss-visible poles addresses near-criticality without damping the full adjoint
response or disturbing resolved modes. Across 23 families of PDE grids,
three-dimensional fields, operator maps, irregular geometries, and particles,
capacity extensions reduce field error while solver-index and physical-time
tests separate backward response from forward dynamics, thereby making
singular inversion a finite response problem shared by dense DEQs and
SILVA.

\FloatBarrier

\endgroup

\clearpage

\section*{Data and Software Availability}
The physical benchmark data used in this study are publicly available. Native
experiments use PDEBench and The Well
\citep{takamoto2022pdebench,ohana2024well}, whereas cross-benchmark experiments use
PDEArena, DynaBench, PDEGym Poisson--Gauss, CFDBench, and LagrangeBench
\citep{gupta2023pdearena,dulny2023dynabench,herde2024poseidon,luo2023cfdbench,toshev2023lagrangebench}.
Controlled mechanism studies use procedurally generated synthetic systems as
specified in the experimental descriptions. Dataset versions, data partitions,
temporal offsets, train-only normalization, random seeds, model settings, and
solver tolerances are specified in Appendix~\ref{app:data-specification}. SILVA Networks is archived on
Zenodo \citep{silva2026software}.

\section*{Acknowledgments}
The broader research program from which this study emerged was initiated
within the Swedish National Infrastructure for Computing (SNIC) Small Compute
project \emph{Artificial Intelligence for Physics and Engineering, Modeling
and Simulation}, Project No.~SNIC 2022/22-843, conducted at Link\"oping
University under the author's principal investigatorship. This work was
supported by the Brazilian National Council for Scientific and Technological
Development (CNPq) under grant No.~445344/2024-5. The author also acknowledges
financial support provided through the Program Talentos Brasil, as Project
Investigator.

\bibliographystyle{plainnat}
\bibliography{references}

\appendix
\small

\section{SILVA Experimental Design}
\label{app:native-silva-details}

The SILVA study compares backward response geometry within a shared forward
architecture and numerical protocol. CMR and Phi-CMR enter only as automatic-differentiation
backward operators, with matrix-free VJP/GMRES products for local adjoints and
a modification confined to the small collective denominator. A no-lift CMR
case therefore recovers the original implicit gradient, whereas a lifted case
is explicitly labeled as a backward-response surrogate. The eight matched
modes are exact implicit, Phantom, Neumann, JFB, SHINE, full-operator Tikhonov,
CMR, and Phi-CMR, allowing predictive differences to be interpreted as
consequences of the backward rule under this controlled design.

The evaluation contains five Darcy regimes and nine additional families,
namely Advection, Burgers, Reaction-\allowbreak Diffusion 1D,
Diffusion-\allowbreak Sorption 1D, Diffusion-\allowbreak Reaction 2D,
Shallow Water 2D, compressible CFD 1D, incompressible
Navier--Stokes 2D, and The Well turbulent radiative layer 2D. Five Darcy
regimes count as one family, giving ten physical families, five seeds, and
560 method evaluations. The selected data subsets occupy 120.685~GB in total,
with the individual storage footprints reported in
Table~\ref{tab:dataset-storage-footprints}.

\begin{table}[H]
\centering
\footnotesize
\setlength{\tabcolsep}{3pt}
\renewcommand{\arraystretch}{0.96}
\caption{Storage footprints of the selected data subsets. Values are decimal
gigabytes ($1~\mathrm{GB}=10^9$ bytes) and exclude unused portions of the
complete source collections.}
\label{tab:dataset-storage-footprints}
\begin{tabular}{@{}P{0.36\textwidth}rP{0.36\textwidth}r@{}}
\toprule
Native dataset & GB & Cross-benchmark dataset & GB\\
\midrule
PDEBench Darcy 2D (five regimes) & 6.554 & PDEArena Navier--Stokes 2D & 0.826\\
PDEBench Advection 1D & 8.233 & PDEArena Shallow Water 2D & 0.089\\
PDEBench Burgers 1D & 8.233 & PDEArena Maxwell 3D & 1.887\\
PDEBench Reaction--Diffusion 1D & 4.137 & PDEArena Kuramoto--Sivashinsky 1D & 0.931\\
PDEBench Diffusion--Sorption 1D & 4.217 & DynaBench Advection & 2.558\\
PDEBench Diffusion--Reaction 2D & 13.244 & DynaBench Burgers & 5.090\\
PDEBench Shallow Water 2D & 6.626 & DynaBench Gas Dynamics & 10.156\\
PDEBench compressible CFD 1D & 12.411 & DynaBench Kuramoto--Sivashinsky & 2.558\\
PDEBench incompressible Navier--Stokes 2D & 9.906 & DynaBench Reaction--Diffusion & 5.090\\
The Well turbulent radiative layer 2D & 7.149 & DynaBench Wave & 5.090\\
& & PDEGym Poisson--Gauss & 2.624\\
& & CFDBench Cylinder Geometry & 2.621\\
& & LagrangeBench Taylor--Green Vortex 2D & 0.454\\
\cmidrule(lr){1-2}\cmidrule(l){3-4}
Native subtotal & 80.710 & Cross-benchmark subtotal & 39.975\\
\midrule
\multicolumn{3}{r}{Combined total} & 120.685\\
\bottomrule
\end{tabular}
\end{table}

Every method receives the same initialization, ordered
16 minibatches, a fixed 64-update direct or 128-update residual budget, and
identical optimizer, forward-solver, and tolerance settings. Because the
backward rule changes parameter updates, trained parameters and equilibria may
differ across methods. Training and test ranges are disjoint, and The Well uses
its dataset-provided training and test partitions, and normalization is fitted only on
training data. The error metric inverse-standardizes the
prediction and target, flattens all stored output entries per sample, and
computes
\(\|{\widehat y}_i-y_i\|_2/\max(\|y_i\|_2,10^{-12})\), then averages samples.
For mixed-unit channels this is a descriptive numerical score that can be
dominated by larger-scale variables rather than a dimensionally homogeneous
physical norm.

\subsection{All-method baselines and transient derivations}
\label{app:silva-transient}

Tikhonov is the primary \emph{mechanistic} reference because it regularizes
the same inverse response, exact implicit differentiation supplies the
fidelity reference, and Phantom and Neumann \citep{geng2021training}, JFB \citep{fung2022jfb}, and SHINE
\citep{ramzi2022shine} provide published
inexact-backward or Jacobian-free training references for comparison with the
CMR and Phi-CMR structured inverses. Table~\ref{tab:silva-main-summary} separates the
resulting accuracy--cost profiles, with CMR giving the lowest mean one-step
error ratio (0.986) and JFB remaining substantially faster than the structured
methods. This distinction matters because agreement with Tikhonov tests the
regularization mechanism, whereas agreement with the broader baseline set
tests whether that mechanism remains competitive as a training rule.

\begin{figure}[H]
\centering
\includegraphics[width=0.98\textwidth]{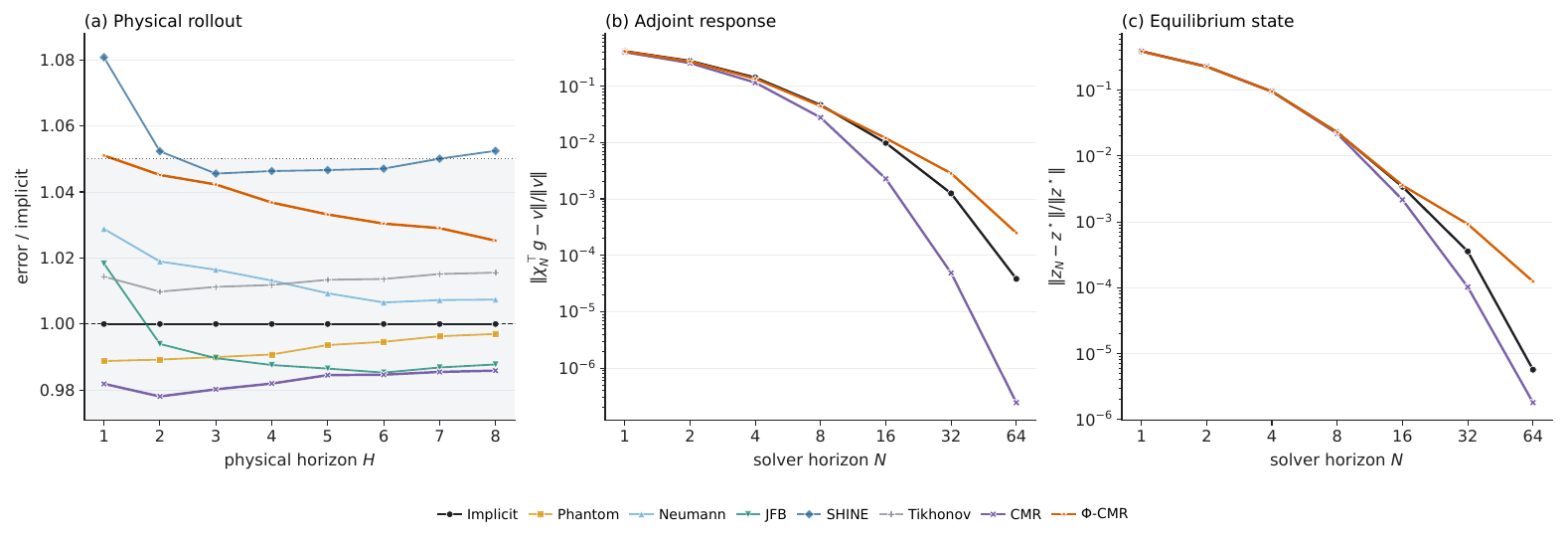}
\small
\caption{Native SILVA transient analysis with (a) family-weighted free
physical-time rollout error relative to models trained with exact implicit
differentiation on nine dynamic datasets and five seeds, (b) finite
causal-response error relative to the converged static adjoint, and (c)
fixed-point state error relative to equilibrium, where
panels (b,c) contain 45 dataset--seed models per method. Shading in panel (a)
denotes error ratios at most 1.05.}
\label{fig:silva-transient}
\end{figure}

Two transient scales leave the static SILVA backward operator unchanged, with
the first describing convergence along the \emph{solver index}. Linearizing
the original fixed-point map at a converged state gives
\begin{equation}
  \delta z_{n+1}=J\,\delta z_n,
  \qquad J=\partial_z f_\theta(z^\star,x),
  \qquad K=I-J.
  \label{eq:solver-transient-linearization}
\end{equation}
For a state-space loss source $g$, define the causal finite-horizon adjoint by
the zero-initialized recurrence
\begin{equation}
  v_0=0,
  \qquad v_{n+1}=g+J^\top v_n.
  \label{eq:finite-adjoint-recurrence}
\end{equation}
Substitution gives $v_1=g$ and $v_2=g+J^\top g$. Induction therefore yields
\begin{equation}
  v_N=\sum_{n=0}^{N-1}(J^\top)^n g,
  \qquad
  v_N\xrightarrow[N\to\infty]{}(I-J^\top)^{-1}g=K^{-\top}g=:v_\infty,
  \label{eq:finite-solver-response}
\end{equation}
when the iteration is stable. The convergence statement follows directly from
the finite geometric-series identity
\begin{align}
  K^\top v_N
  &=(I-J^\top)\sum_{n=0}^{N-1}(J^\top)^n g
    =\bigl[I-(J^\top)^N\bigr]g,\nonumber\\
  v_\infty-v_N
  &=(J^\top)^N v_\infty,
  \label{eq:finite-solver-gap}\\
  K^\top v_N-g
  &=-(J^\top)^N g.
  \label{eq:finite-solver-residual}
\end{align}
Consequently,
\begin{equation}
  \frac{\|v_N-v_\infty\|_2}{\|v_\infty\|_2}
  \leq \|(J^\top)^N\|_2,
  \qquad
  \frac{\|K^\top v_N-g\|_2}{\|g\|_2}
  \leq \|(J^\top)^N\|_2.
  \label{eq:finite-solver-bounds}
\end{equation}
These bounds also explain why a non-normal $J$ can show finite-$N$ transient
amplification even when its spectral radius is below one. Equations
\eqref{eq:finite-adjoint-recurrence}--\eqref{eq:finite-solver-bounds} are the
quantities evaluated in Fig.~\ref{fig:silva-transient}(b,c). The independent
causal-kernel derivation appears in
Eqs.~\eqref{eq:appendix-causal-response-recurrence}--\eqref{eq:appendix-zero-frequency-response}.

SILVA enters only at the static backward endpoint, where
\(b=K_L^{-\top}g\), \(Y=K_L^{-\top}B\),
\(h=A^\top b\), and collective lift \(\Gamma_{\rm eff}\) give the CMR/Phi-CMR response
\begin{equation}
  v^R=b+Y\Gamma_{\rm eff}^{-\top}h,
  \qquad
  \Gamma=I-B^\top K_L^{-1}A.
  \label{eq:silva-static-transient-boundary}
\end{equation}
Because this expression is a collective backward surrogate unless no lift
occurs, it does not define a full-state \(U\Sigma_{\rm eff}V^\top\), while the
transient evaluation instead constructs $v_N$ and $v_\infty$ from the
\emph{original} $J$ for each trained model. By neither iterating a renormalized
$J^R=I-K^R$ nor interpreting an effective denominator as a physical time
constant, the evaluation tests the causal approach to the converged static
adjoint rather than a new finite-frequency SILVA law.

The second scale concerns \emph{physical PDE time}. Let $\mathcal P$ be
the reference one-step evolution and let $\mathcal P_\theta$ be the learned
one-step map, evaluated with the inner SILVA equilibrium solved to the stated
tolerance. A free
rollout is
\begin{equation}
  y_{t+1}=\mathcal P(y_t),
  \qquad
  \widehat y_{t+1}=\mathcal P_\theta(\widehat y_t),
  \qquad
  e_t=\widehat y_t-y_t.
  \label{eq:physical-rollout-definitions}
\end{equation}
Define the local one-step modeling defect
$d_t=\mathcal P_\theta(y_t)-\mathcal P(y_t)$ and tangent
$A_t=D\mathcal P_\theta(y_t)$. Taylor expansion around the reference state
gives the first-order error recurrence
\begin{equation}
  e_{t+1}=A_t e_t+d_t+r_t,
  \qquad \|r_t\|_2=\mathcal O(\|e_t\|_2^2).
  \label{eq:physical-rollout-error-recursion}
\end{equation}
For $\mathcal T_{b:a}:=A_{b-1}\cdots A_a$ and
$\mathcal T_{a:a}:=I$, repeated substitution produces
\begin{align}
  e_H
  &=\mathcal T_{H:0}e_0
    +\sum_{j=0}^{H-1}\mathcal T_{H:j+1}(d_j+r_j),
  \label{eq:physical-rollout-error-unrolled}\\
  \|e_H\|_2
  &\leq \|\mathcal T_{H:0}\|_2\|e_0\|_2
    +\sum_{j=0}^{H-1}\|\mathcal T_{H:j+1}\|_2
      \left(\|d_j\|_2+\mathcal O(\|e_j\|_2^2)\right).
  \label{eq:physical-rollout-error-bound}
\end{align}
This separates local one-step error from its amplification by the learned
physical-time tangent products. With a shared reference initial condition,
$e_0=0$, only the accumulated defect terms remain. The horizon-$h$
metric is
\begin{equation}
  \epsilon_h^{(m)}
  =\frac{\|\widehat y_{t+h}^{(m)}-y_{t+h}\|_2}
  {\|y_{t+h}\|_2+\varepsilon},
  \qquad
  R_h^{(m)}=\frac{1}{|\mathcal D|}
  \sum_{(f,s)\in\mathcal D}
  \frac{\epsilon_{h,f,s}^{(m)}}
  {\epsilon_{h,f,s}^{(\mathrm{implicit})}},
  \label{eq:physical-rollout-metric}
\end{equation}
where $\mathcal D$ contains the equally weighted family--seed pairs. This is
the quantity plotted in Fig.~\ref{fig:silva-transient}(a), and the backward method
changes the learned model, while every rollout evaluates its inner SILVA
equilibrium with the same solver and stated tolerance.

The physical-time study contains $9\times5\times8\times8=2880$ repeated
evaluations from 45 independently seeded family--model units and shows no
consistent long-horizon penalty from backward response control under the
evaluated conditions. At horizon eight, Phi-CMR has mean error ratio 1.025,
with ratios at most 1.05 in 43 of 45 family--seed pairs, compared with
all 45 for CMR, 43 of 45 for Phantom and Tikhonov, 42 of 45 for Neumann and JFB,
and 31 of 45 for SHINE. Because Reaction--Diffusion 1D is the hardest long-horizon case,
both mean and median are shown. Independently, 945 solver-horizon evaluations
test the causal response identity, with the Phi-CMR mean finite-response error
falling from 0.411 at $N=1$ to $2.50\times10^{-4}$ at $N=64$ as state error
falls from 0.387 to $1.24\times10^{-4}$. The sampled final-state sources do not
activate a lift, although training traces activate in every physical family,
which supports the precise interpretation that CMR/Phi-CMR changes the static
backward endpoint when a pole is active, while physical evolution and the
causal approach to that endpoint remain properties of the learned SILVA map.

\begin{table}[H]
\centering
\scriptsize
\setlength{\tabcolsep}{4pt}
\caption{SILVA convergence, fidelity, and activation diagnostics. Statistics
use physical-family weighting, with the five Darcy regimes contributing one
family-level unit.}
\label{tab:silva-publication-gates}
\begin{tabular}{@{}lcc@{}}
\toprule
Diagnostic & Observed & Reference level\\
\midrule
Maximum final forward residual & \(9.98\times10^{-8}\) & \(<10^{-5}\)\\
Maximum implicit/CMR/Phi adjoint residual & \(9.99\times10^{-9}\) & \(<10^{-6}\)\\
Structured-method realizations improving from initialization & 57.0\% & Descriptive\\
Families with collective activation & 10/10 & 10/10\\
Phi-CMR error ratio $\leq1.05$ & 48/50 pairs & 40/50 pairs\\
Phi-CMR lower error than Tikhonov & 27/50 pairs & 25/50 pairs\\
\bottomrule
\end{tabular}
\end{table}

The residuals lie below the stated numerical tolerances, while the Phi-CMR test-error
ratio to the exact-implicit training baseline is at most 1.05 in 48 of 50 family--seed
pairs and the structured mechanism activates in every family. Its error is
lower than Tikhonov in 27 of 50 paired comparisons, all dataset-level paired
95\% intervals against implicit include zero, and 12 of 14 intervals against Tikhonov include
zero, with Reaction--Diffusion 1D and Shallow Water 2D resolving in opposite
directions. Numerical convergence, mechanism activation, and predictive
fidelity therefore support distinct parts of the argument rather than serving
as interchangeable success measures.

\subsection{Datasets and Reproducibility}
\label{app:data-specification}

The learning tasks consume stored benchmark fields rather than regenerating the
PDE solutions, with governing equations, domains, boundary conditions,
numerical discretizations, and generation parameters following the cited
dataset releases \citep{takamoto2022pdebench,ohana2024well,gupta2023pdearena,
dulny2023dynabench,herde2024poseidon,luo2023cfdbench,toshev2023lagrangebench}.
PDEBench supplied five Darcy coefficient-to-solution regimes and temporal
state-to-next-state tasks for Advection, Burgers, Reaction--Diffusion,
Diffusion--Sorption, Diffusion--Reaction, Shallow Water, compressible CFD, and
incompressible Navier--Stokes, while The Well supplied the turbulent radiative-layer
multichannel transition. The cross-benchmark evaluation used PDEArena fields,
DynaBench trajectories, PDEGym Poisson--Gauss source-to-solution pairs,
CFDBench geometry-conditioned velocity fields, and LagrangeBench particle
trajectories. For temporal data, ``next'' denotes the subsequent frame in the
released trajectory at the selected offset rather than a newly computed PDE solution.

Dataset-provided training, validation, and test partitions define the splits when
available, with PDEBench and PDEGym using deterministic disjoint sample or time
partitions and CFDBench using disjoint case-level 80/10/10 partitions, while
normalization is fitted only on training data. Each experiment specifies the
dataset version, selected variables, temporal offsets, method, seed, response
parameters, architecture, optimization budget, and forward/backward tolerances.
Hardware configurations and library versions are unavailable for a subset of
the native and cross-benchmark experiments. Accordingly, the reproducibility claim
concerns the numerical conclusions
under the stated design rather than bitwise cross-platform identity. The five
independent seeds are 123, 456, 789, 1011, and 2027, and none of the SILVA
experiments uses stochastic data augmentation, gradient clipping, a
learning-rate schedule, or early stopping.

The matched native and cross-benchmark studies use \(m_0=0.08\),
\(\kappa=0.12\), collective rank one, local and global initialization scales
0.65 and 0.34, and \(\mu=0.05\) for Tikhonov. Phi-CMR applies the per-mode rule
in Eq.~\eqref{eq:main-phi-mode-law} to the singular values of \(\Gamma\)
(\(\gamma=\Phi_{\rm mode}\) in Algorithm~\ref{alg:silva-cmr-exact}).
Adam uses learning rate \(10^{-3}\), and the native and cross-benchmark hidden
widths are 48 and 32. The native study allows 200 forward and 200 backward
iterations, while the cross-benchmark study allows 120 of each, and both use
forward and backward tolerances \(10^{-7}\) and \(10^{-8}\). Minibatches
contain four or eight samples and the held-out evaluations contain 16 or 32
samples, according to the field representation.

\begin{figure}[!t]
\centering
\includegraphics[width=0.92\textwidth]{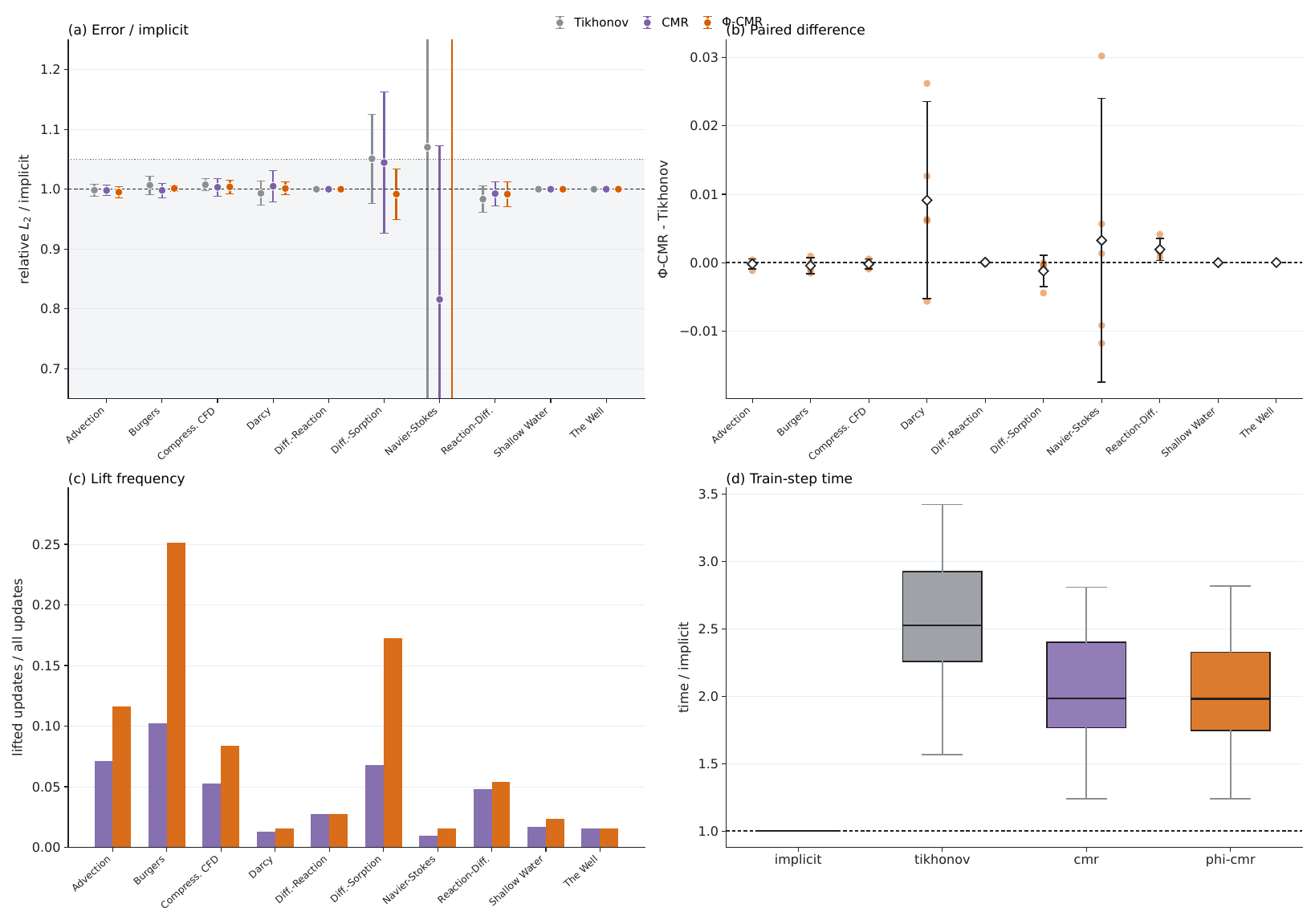}
\small
\caption{Native SILVA multiphysics evaluation with (a) final inverse-standardized
sample-relative Euclidean error normalized by exact implicit differentiation and paired-seed
95\% intervals, (b) paired Phi-CMR-minus-Tikhonov differences, (c) the fraction
of training updates with a lifted collective response, and (d) measured
forward-solve plus backward wall time relative to implicit differentiation,
with the Darcy regimes aggregated as one physical family. Shading in panel (a)
denotes error ratios at most 1.05. The Darcy confidence
interval in (a) extends beyond the displayed range.}
\label{fig:silva-multiphysics}
\end{figure}

\begingroup
\small
\setlength{\parskip}{0pt}
\setlength{\abovedisplayskip}{3pt plus 1pt minus 1pt}
\setlength{\belowdisplayskip}{3pt plus 1pt minus 1pt}
\setlength{\textfloatsep}{4pt plus 1pt minus 1pt}
\setlength{\floatsep}{4pt plus 1pt minus 1pt}

\section{Implicit Gradients, Poles, and Baseline Filters}
\label{app:implicit-derivation}
\label{app:compatibility}
\label{app:baseline-equations}

Applying the implicit function theorem to
\(R_\theta(z^\star,x)=0\) \citep{krantz2002implicit,bai2019deep} gives
the sensitivity relation directly. Holding \(x\) fixed and varying one scalar
parameter \(\theta_a\), the total differential of the equilibrium condition is
\[
 0=dR_\theta
 =\partial_zR_\theta\,dz^\star
  +\partial_{\theta_a}R_\theta\,d\theta_a
 =K\,dz^\star-B_a\,d\theta_a,
\]
because \(B_a=\partial_{\theta_a}f_\theta\) and therefore
\(\partial_{\theta_a}R_\theta=-B_a\). Dividing by \(d\theta_a\) gives
\begin{equation}
 K\frac{\partial z^\star}{\partial\theta_a}=B_a,
 \qquad
 \frac{\partial z^\star}{\partial\theta_a}=K^{-1}B_a.
 \label{eq:forward-sensitivity-scalar}
\end{equation}
The chain rule then yields
\(d\mathcal L/d\theta_a=
\partial_{\theta_a}\mathcal L+g^\top K^{-1}B_a\). Defining the adjoint by
\(K^\top v=g\) moves the inverse outside the parameter dimension and gives
\begin{equation}
 \frac{d\mathcal L}{d\theta_a}
 =\partial_{\theta_a}\mathcal L+v^\top B_a.
 \label{eq:parameter-gradient-adjoint}
\end{equation}
Here \(\partial_{\theta_a}\mathcal L\) is the direct derivative at fixed
equilibrium state, whereas \(d\mathcal L/d\theta_a\) also contains the change
of \(z^\star\) induced by \(\theta_a\). The adjoint solve is shared by all
parameters, and the remaining parameter dependence enters through the
contractions \(v^\top B_a\).
In singular coordinates, \(K^\top v=g\) becomes
\(\Sigma U^\top v=V^\top g\), which gives each nonsingular coefficient as
\(u_i^\top v=(v_i^\top g)/\sigma_i\), and hence
\begin{equation}
 \frac{d\mathcal L}{d\theta_a}
 =\partial_{\theta_a}\mathcal L+
 \sum_i\frac{(v_i^\top g)(u_i^\top B_a)}{\sigma_i}.
 \label{eq:two-sided-parameter-gradient}
\end{equation}
At an exact pole, solvability of the adjoint requires
\begin{equation}
 V_0^\top g=0,
 \label{eq:compatibility-condition}
\end{equation}
where \(V_0\) spans \(\operatorname{Null}(K)\). When this condition holds,
the generalized inverse \citep{benisrael2003generalized} gives the
minimum-norm response \((K^\top)^+g=U\Sigma^+V^\top g\). When it fails,
the unchanged deterministic derivative is not finite, which separates a
source-visible pole from a removable null direction.

The exact, Tikhonov, and denominator-lifted modal responses are
\begin{equation}
 v_{\rm exact}=U\diag(\sigma_i^{-1})V^\top g,
 \label{eq:baseline-exact}
\end{equation}
\[
 v_\mu=U\diag\!\left(\frac{\sigma_i}{\sigma_i^2+\mu^2}\right)V^\top g,
\quad
 v_{\rm TSVD}=U\diag\!\left(\frac{\mathbf 1_{\sigma_i\geq\kappa}}{\sigma_i}\right)V^\top g,
 \qquad
 v_R=U\diag\!\left(\frac{1}{\widehat\sigma_i}\right)V^\top g.
\]
The Tikhonov expression \citep{tikhonov1963solution} follows by differentiating
\(\|K^\top v-g\|_2^2+\mu^2\|v\|_2^2\), obtaining
\((KK^\top+\mu^2I)v=Kg\), and substituting the SVD, whereas CMR sets
\(\widehat\sigma_i=\sigma_i\) on \(\mathcal S\) and
\(\widehat\sigma_i=\max(\sigma_i,m_i)\) on \(\mathcal C\), thereby preserving
every resolved stable channel exactly.

\paragraph{Worked two-mode check.}
Let \(K=\diag(1,10^{-4})\), \(g=(1,1)^\top\), and parameter sources
\(B_1=(1,0)^\top\), \(B_2=(0,1)^\top\), for which the exact adjoint is
\((1,10^4)^\top\) and Eq.~\eqref{eq:two-sided-parameter-gradient} gives
parameter contributions \((1,10^4)\). A CMR mass \(m=0.05\) changes these to
\((1,20)\), whereas TSVD with \(\kappa=10^{-3}\) gives \((1,0)\), with both
biased rules retaining the first resolved mode. If instead
\(B_2=(1,0)^\top\), the critical loss projection remains nonzero but its
contribution to parameter 2 vanishes, illustrating why \(v_i^\top g\) alone is
insufficient.
For the stated CMR mass, the counterterm is
\(\Delta K=\diag(0,0.05-10^{-4})=\diag(0,0.0499)\). Consequently,
\(K^R=K+\Delta K=\diag(1,0.05)\), and direct substitution into
\((K^R)^\top v^R=g\) returns \(v^R=(1,20)^\top\). This verifies in one
calculation the denominator lift, the counterterm, the lifted adjoint, and its
parameter-source contributions. If the second denominator lies outside the
critical set or already exceeds its target mass, then \(\Delta K=0\) and the
same calculation recovers the exact implicit adjoint.

\paragraph{Exact modified model and backward-only rule.}
\label{app:objective-boundary}
Let
\[
 R^R_{\theta,\kappa}(z,x)=R_\theta(z,x)
 +\Delta K_{\mathcal A}(z-z_{{\rm ref},\mathcal A}),
 \qquad K^R=K+\Delta K_{\mathcal A},
\]
with the anchor \(\mathcal A\) fixed during one derivative. The lifted equation
\begin{equation}
 (K^R)^\top v^R=g^R
 \label{eq:objective-consistent-adjoint}
\end{equation}
is the exact adjoint of the modified equilibrium, and because
\(\partial_{\theta_a}R^R=-B_a\) at a frozen anchor, implicit differentiation
repeats the preceding steps and gives
\begin{equation}
 \frac{d\mathcal L^R}{d\theta_a}
 =\partial_{\theta_a}\mathcal L^R+(v^R)^\top B_a.
 \label{eq:appendix-modified-parameter-gradient}
\end{equation}
Keeping the original forward equilibrium while solving the lifted adjoint
produces a backward surrogate, with the original and modified residuals
distinguishing these two uses.

\section{Matrix-Free, Causal, and Renormalized Response}
\label{app:matrix-free}
\label{app:response-field}

\paragraph{Dense and matrix-free realizations.}
The matrix-free formulation changes how the required linear-algebra objects
are obtained but leaves the response operator unchanged. When the state
dimension permits, one may form \(J\) and \(K=I-J\) explicitly, compute a dense
SVD, and compare the modal reconstruction with a direct solve of the lifted
adjoint equation. In the matrix-free formulation, automatic differentiation
supplies only the actions \(w\mapsto Kw\) and \(w\mapsto K^\top w\). An
iterative singular solver then uses both actions to recover the smallest
triplets, which must satisfy the two triplet-residual conditions in
Algorithm~\ref{alg:cmr-backward}. The requested rank is sufficient when the
largest returned small singular value is at or above the cutoff, with the rank
increased until this condition is met. This ensures that every resolved denominator
below \(\kappa\) has been included and avoids replacing the required singular
geometry by eigenvalue information from \(J\).

Automatic differentiation \citep{baydin2018automatic} supplies the actions
\[
 Kw=w-Jw,\qquad K^\top w=w-J^\top w,
\]
These identities allow partial singular solvers and GMRES
\citep{golub1965calculating,lehoucq1998arpack,saad1986gmres} to operate
without materializing \(J\). For a SILVA operator \citep{silva2026silvanetworks}
\(K=K_L-AB^\top\), the Woodbury identity
\citep{woodbury1950inverting,sherman1950adjustment} gives
\[
 K^{-1}=K_L^{-1}+K_L^{-1}A
 (I-B^\top K_L^{-1}A)^{-1}B^\top K_L^{-1}.
\]
Writing \(\Gamma=I-B^\top K_L^{-1}A\) and transposing the identity yields
\begin{equation}
 v=K_L^{-\top}g+K_L^{-\top}B\,\Gamma^{-\top}A^\top K_L^{-\top}g.
 \label{eq:appendix-woodbury-adjoint}
\end{equation}
Only the small collective denominator \(\Gamma\) is decomposed and lifted.
If the state dimension is \(d\) and the structured correction has rank \(r\),
then
\[
 K_L\in\mathbb R^{d\times d},\quad
 A,B,X,Y\in\mathbb R^{d\times r},\quad
 \Gamma\in\mathbb R^{r\times r},\quad
 b,v_R\in\mathbb R^d,\quad h,x_R\in\mathbb R^r,
\]
where \(X=K_L^{-1}A\), \(Y=K_L^{-\top}B\),
\(b=K_L^{-\top}g\), \(h=A^\top b\), and
\(x_R=\Gamma_{\rm eff}^{-\top}h\). These dimensions give
\(v_R=b+Yx_R\) and expose the reduction from a \(d\)-dimensional adjoint to an
\(r\)-dimensional collective solve. The factorization can be checked without
forming the full operator by verifying
\(Kw=K_Lw-A(B^\top w)\) on independent probe vectors, together with the local
and collective residuals.
The determinant lemma gives
\[
 \det K=\det K_L\,\det\Gamma,
\]
but it does not equate the singular spectra. A useful conditioning bound is
\[
 \|K^{-1}\|_2\leq\|K_L^{-1}\|_2
 \left(1+\|A\|_2\|\Gamma^{-1}\|_2\|B\|_2\|K_L^{-1}\|_2\right).
\]
Accordingly, collective, local, and full-\(K\) diagnostics are evaluated
separately. For factor-gauge stability, compute
\(AB^\top=P\Sigma Q^\top\) and use balanced factors
\(\bar A=P\Sigma^{1/2}\), \(\bar B=Q\Sigma^{1/2}\). This leaves \(K\)
unchanged and prevents arbitrary rescaling of raw factor norms.

To derive the response interpretation, perturb the fixed-point iteration by a
source \(s_n\),
\begin{equation}
 z_{n+1}=f_\theta(z_n,x)+\epsilon s_n.
 \label{eq:appendix-sourced-iteration}
\end{equation}
Set \(z_n=z^\star+\epsilon\phi_n+O(\epsilon^2)\). Taylor expansion at the
equilibrium gives the causal recurrence
\begin{equation}
 \phi_{n+1}=J\phi_n+s_n,
 \label{eq:appendix-causal-response-recurrence}
\end{equation}
hence
\[
 \phi_n=\sum_{m<n}J^{n-1-m}s_m,\qquad
 \chi_{n,m}=\frac{\partial z_n}{\partial s_m}=J^{n-1-m}\mathbf 1_{n>m}.
\]
For a constant source and \(\rho(J)<1\), summing the geometric series produces
\begin{equation}
 G_0(0)=\sum_{\ell=0}^{\infty}J^\ell=(I-J)^{-1}=K^{-1}.
 \label{eq:appendix-zero-frequency-response}
\end{equation}
The discrete Fourier transform gives the same limit through
\(G_0(\omega)=(e^{\mathrm i\omega}I-J)^{-1}\).

The following construction supplies a response-theoretic interpretation of the
same operator identities. Its generating functional and diagrammatic expansion
are not required to evaluate the dense or matrix-free denominator lift. A
response-field representation
\citep{martin1973statistical,janssen1976lagrangean,dedominicis1976techniques}
inserts the trajectory constraint with a Fourier multiplier
\(\widetilde z_n\),
\[
 1=\int D\widetilde z\,
 \exp\!\left[-\mathrm i\sum_n\widetilde z_n^\top
 (z_{n+1}-f_\theta(z_n,x)-s_n)\right].
\]
Differentiating the resulting generating functional once with respect to a
state probe and once with respect to the source returns the mixed correlator
\(\langle z_n\widetilde z_m^\top\rangle=\chi_{n,m}\). Expanding the static
equation \(K\phi=\epsilon s+\frac12F^{(2)}[\phi,\phi]+\cdots\) gives
\begin{equation}
 \phi=\epsilon G_0s+
 \frac{\epsilon^2}{2}G_0F^{(2)}[G_0s,G_0s]+O(\epsilon^3),
 \qquad G_0=K^{-1}.
 \label{eq:appendix-static-source-series}
\end{equation}
Thus every perturbative line carries a factor of \(G_0\), matching the
propagator structure of path-integral and diagrammatic perturbation theory
\citep{feynman1948space,feynman1949qed}, and nonlinear
insertions dress the inverse through the Dyson identity \citep{dyson1949smatrix}
\[
 G=G_0+G_0\Pi G,\qquad G^{-1}=G_0^{-1}-\Pi.
\]
Projecting onto a selected singular channel replaces the bare denominator by
\(\sigma_i-\pi_i(0)\). Response renormalization imposes the finite matching
condition
\begin{equation}
 v_i^\top(K+\Delta K)^\top u_i
 =\sigma_i+\delta_i=\widehat\sigma_i>0,
 \label{eq:appendix-cmr-matching-condition}
\end{equation}
which is realized by the minimum-rank counterterm
\begin{equation}
 \Delta K
 =U_{\mathcal C}\diag\!\left([\widehat\sigma_i-\sigma_i]_+\right)
 V_{\mathcal C}^\top.
 \label{eq:appendix-cmr-counterterm}
\end{equation}
Substitution into \((K+\Delta K)^\top v_R=g\) gives
\begin{equation}
 v_R=U_{\mathcal C}\diag(\widehat\sigma_i^{-1})V_{\mathcal C}^\top g
 +U_{\mathcal S}\diag(\sigma_i^{-1})V_{\mathcal S}^\top g.
 \label{eq:appendix-cmr-substitution-chain}
\end{equation}
This chain connects the causal DEQ response, the zero-frequency pole, the
Dyson denominator, and the low-rank lift used here.

\section{Coordinate Metric and Threshold Stability}
\label{app:coordinate-metric}

Under a nonsingular state change \(y=Sz\), use
\[
 R_y=S R,\quad K_y=SKS^{-1},\quad g_y=S^{-\top}g,\quad B_{y,a}=SB_a.
\]
Although \(g^\top K^{-1}B_a=g_y^\top K_y^{-1}B_{y,a}\), Euclidean singular
values of \(K\) are not similarity invariants. To declare the state geometry,
let \(H\succ0\) be a metric, \(H=L^\top L\), and apply CMR to
\(\widetilde K=LKL^{-1}\) and \(\widetilde g=L^{-\top}g\). Under the coordinate
change, \(H_y=S^{-\top}HS^{-1}\), and mapping the adjoint back preserves
\(v^\top B_a\), which makes the construction coordinate-consistent under the
declared metric.

For threshold stability, select a subspace rather than an arbitrary basis in a
near-degenerate cluster. Modes below \(\kappa-\eta\) are critical, those above
\(\kappa+\eta\) stable, and the intervening band is classified as unresolved.
Perturbation checks vary \(\kappa\), \(\eta\), the requested rank, solver
tolerance, and floating-point precision, while discontinuous decisions inside the
unresolved interval are not interpreted as physical transitions.

\section{Notation and Response Identities}
\label{app:notation}

For an equilibrium \(z^\star=f_\theta(z^\star,x)\), define the residual,
Jacobian denominator, loss source, and parameter source by
\[
 R_\theta=z-f_\theta(z,x),\qquad K=\partial_zR_\theta=I-J,
 \qquad g=\nabla_{z^\star}\mathcal L,\qquad
 B_a=\partial_{\theta_a}f_\theta(z^\star,x).
\]
The singular-value decomposition is \(K=U\Sigma V^\top\), with
\(0\leq\sigma_1\leq\cdots\leq\sigma_d\). A critical set
\(\mathcal C=\{i:\sigma_i<\kappa\}\) has bases \(U_{\mathcal C}\) and
\(V_{\mathcal C}\), while \(\mathcal S\) denotes its stable complement. A
critical mode is loss-visible when \(v_i^\top g\neq0\) and parameter-visible
for \(a\) when \(u_i^\top B_a\neq0\). The remaining notation is
\([a]_+=\max(a,0)\),
\(\mathcal P_{[\ell,u]}(a)=\min\{\max(a,\ell),u\}\), and
\(\diag(a_i)\) for the diagonal matrix with entries \(a_i\).
The unscripted \(v\) denotes the adjoint, while \(v_i\) denotes the \(i\)-th
right singular vector. For a vector-valued parameter \(\theta\), the scalar
sources \(B_a\) define the components of the parameter derivative but need not
be assembled as a matrix, because their contraction is the single reverse-mode
product \((\partial_\theta f_\theta)^\top v\).

\begin{table}[H]
\centering
\scriptsize
\setlength{\tabcolsep}{4pt}
\caption{Response-renormalization notation.}
\label{tab:notation}
\begin{tabular}{@{}llll@{}}
\toprule
Symbol & Meaning & Symbol & Meaning\\
\midrule
\(J\) & equilibrium Jacobian & \(K=I-J\) & response denominator\\
\(g\) & loss source & \(v\) & implicit adjoint\\
\(\sigma_i,u_i,v_i\) & singular triplet of \(K\) & \(\kappa\) & critical cutoff\\
\(m_0,m_i\) & base and effective masses & \(a_{\mathcal C}\) & critical source fraction\\
\(K_L,A,B\) & local and low-rank factors & \(\Gamma\) & collective denominator\\
\(\rho_0,\rho_R\) & original and lifted residuals & \(\mathcal P_I\) & projection onto interval \(I\)\\
\bottomrule
\end{tabular}
\end{table}

\section{Response-Renormalization Algorithms}
\label{app:algorithm}

The procedures use \(\gamma\) as the gate selector, \(r\) as the requested number
of singular triplets, \(\kappa>0\) as the critical cutoff, and
\(\tau_{\rm svd},\tau_K>0\) as singular-triplet and linear-solve tolerances.

\paragraph{Reference dense construction.}
The complete dense calculation can be summarized by the operator sequence
\[
 z^\star
 \longrightarrow (J,K,g)
 \longrightarrow (U,\Sigma,V)
 \longrightarrow (\mathcal C,m_i,\widehat\sigma_i)
 \longrightarrow \Delta K
 \longrightarrow v_R
 \longrightarrow \nabla_\theta^R\mathcal L.
\]
After solving the forward equilibrium, one evaluates
\(J=\partial_zf_\theta(z^\star,x)\), \(K=I-J\), and
\(g=\nabla_{z^\star}\mathcal L\). An SVD of \(K\), with its singular values
arranged in ascending order, identifies \(\mathcal C=\{i:\sigma_i<\kappa\}\).
The chosen gate determines the target masses, and
\(\widehat\sigma_i=\max(\sigma_i,m_i)\) is applied only to selected modes.
The counterterm then follows from
\(\Delta K=U_{\mathcal C}\diag(\widehat\sigma_i-\sigma_i)
V_{\mathcal C}^\top\), after which the corresponding adjoint equation is
solved and the parameter gradient is obtained by one reverse-mode product.
The dense calculation is internally verified by the forward residual, the two
singular-triplet residuals, the lifted-adjoint residual, and exact recovery
when the active set is empty. The two-mode calculation in
Appendix~\ref{app:implicit-derivation} supplies a closed-form reference case.

The forward and backward choices determine which objective is differentiated:
\[
\begin{array}{ccl}
z^\star\text{ from }R_\theta=0, & K^\top v=g,
& \text{exact gradient of the original DEQ},\\
z^\star\text{ from }R_\theta=0, & (K+\Delta K)^\top v_R=g,
& \text{backward-response surrogate},\\
z^R\text{ from }R^R_{\theta,\kappa}=0, & (K^R)^\top v^R=g^R,
& \text{exact gradient of the frozen-anchor modified DEQ}.
\end{array}
\]
When \(\Delta K=0\), these alternatives coincide with the original implicit
adjoint. When the lift is active, retaining both the original-equation and
lifted-equation residuals distinguishes the intended response change from
linear-solve error.

\begin{algorithm2e}[H]
\LinesNumbered
\DontPrintSemicolon
\footnotesize
\caption{Critical-subspace implicit backward pass for CMR, Phi-CMR, and Delta-Phi}
\label{alg:cmr-backward}
\KwIn{Equilibrium \(z^\star\), actions \(w\mapsto Kw\) and
\(w\mapsto K^\top w\), source \(g\), requested rank \(r\), cutoff
\(\kappa\), base mass \(m_0\), gate
\(\gamma\in\{\mathrm{CMR},\Phi,\Delta\Phi\}\), parameters
\((\alpha_{\max},\lambda,c_{\max},\varepsilon_{\rm den})\), tolerances
\((\tau_{\rm svd},\tau_K)\), and optional exact adjoint \(v_{\rm exact}\)}
\KwOut{Gated adjoint \(v_{\rm gate}\), selected triplets, effective
denominators, residuals, and response diagnostics}
Set \(d=\dim(g)=\dim(z^\star)\)\;
Estimate the \(r\) smallest triplets \((u_i,\sigma_i,v_i)\) satisfying
\(Kv_i=\sigma_i u_i\) and \(K^\top u_i=\sigma_i v_i\) by dense or iterative
partial SVD\;
Compute \(\eta_{\rm svd}=\max_i\{\|Kv_i-\sigma_i u_i\|_2,
\|K^\top u_i-\sigma_i v_i\|_2\}\)\;
\If{\(\eta_{\rm svd}>\tau_{\rm svd}\)}{
  Return an unresolved-triplet diagnostic and tighten the singular solver\;
}
Sort \(0\leq\sigma_1\leq\cdots\leq\sigma_r\) and set
\(\mathcal C=\{i\leq r:\sigma_i<\kappa\}\)\;
\If{\(r<d\) and \(\sigma_r<\kappa\)}{
  Return an incomplete-rank diagnostic and repeat with larger \(r\)\;
}
\eIf{\(\mathcal C=\varnothing\)}{
  Solve \(K^\top v_{\rm gate}=g\) to tolerance \(\tau_K\)\;
  Set \(\widehat\sigma_i=\sigma_i\) and
  \(\rho_0=\rho_R=\|K^\top v_{\rm gate}-g\|_2\), then return\;
}{
  Form \(U_{\mathcal C}=[u_i]_{i\in\mathcal C}\),
  \(V_{\mathcal C}=[v_i]_{i\in\mathcal C}\), and
  \(c_{\mathcal C}=V_{\mathcal C}^\top g\)\;
  Set \(P_U=I-U_{\mathcal C}U_{\mathcal C}^\top\),
  \(P_V=I-V_{\mathcal C}V_{\mathcal C}^\top\),
  \(g_{\mathcal C}=V_{\mathcal C}c_{\mathcal C}\), and
  \(g_{\mathcal S}=P_Vg\)\;
  Compute \(a_{\mathcal C}=\|c_{\mathcal C}\|_2^2/
  (\|g\|_2^2+\varepsilon_{\rm den})\),
  \(\sigma_{\mathcal C}=\min_{i\in\mathcal C}\sigma_i\), and
  \(p_{\mathcal C}=\mathcal P_{[0,1]}((\kappa-\sigma_{\mathcal C})/\kappa)\)\;
  \eIf{\(\gamma=\mathrm{CMR}\)}{
    Set \(m_i=m_0\) for every \(i\in\mathcal C\)\;
  }{
    Set \(m_i^\Phi=m_0(1+m_0/(\sigma_i+m_0))\) for every
    \(i\in\mathcal C\)\;
    \eIf{\(\gamma=\Phi\)}{
      Set \(m_i=m_i^\Phi\)\;
    }{
      Set \(s_{\mathcal C}=\mathcal P_{[-1,1]}(2a_{\mathcal C}-1)\) and
      \(m_{\mathcal C}^\Phi=m_0[1+(\alpha_{\max}-1)p_{\mathcal C}]\)\;
      Set \(m^{\Delta\Phi}=\mathcal P_{[m_0,c_{\max}m_0]}
      (m_{\mathcal C}^\Phi+m_0\lambda p_{\mathcal C}s_{\mathcal C})\)\;
      Set \(m_i=\max(m_i^\Phi,m^{\Delta\Phi})\) for every
      \(i\in\mathcal C\)\;
    }
  }
  Set \(\widehat\sigma_i=\max(\sigma_i,m_i)\),
  \(\delta_i=\widehat\sigma_i-\sigma_i\), and
  \(\Delta K=U_{\mathcal C}\diag(\delta_i)V_{\mathcal C}^\top\)\;
  Solve \(P_VK^\top P_Uv_{\mathcal S}=g_{\mathcal S}\), constrained by
  \(v_{\mathcal S}=P_Uv_{\mathcal S}\), to tolerance \(\tau_K\)\;
  Reconstruct \(v_{\mathcal C}=U_{\mathcal C}
  \diag(\widehat\sigma_i^{-1})c_{\mathcal C}\) and set
  \(v_{\rm gate}=v_{\mathcal S}+v_{\mathcal C}\)\;
  Compute \(\rho_0=\|K^\top v_{\rm gate}-g\|_2\) and
  \(\rho_R=\|(K+\Delta K)^\top v_{\rm gate}-g\|_2\)\;
  Retain \((\sigma_i,\widehat\sigma_i,\delta_i)\), \(\eta_{\rm svd}\),
  \(a_{\mathcal C}\), lifted rank, Krylov iterations, \(\rho_0\), and \(\rho_R\)\;
  \If{\(v_{\rm exact}\) is available}{
    Compute the norm ratio, cosine, relative error, signed projection, and
    critical magnitude suppression defined in
    Appendix~\ref{app:quantitative-metrics}\;
  }
  Return \(v_{\rm gate}\) and the diagnostics\;
}
\end{algorithm2e}

\clearpage
\begin{algorithm2e}[H]
\LinesNumbered
\DontPrintSemicolon
\normalsize
\caption{Dense or matrix-free DEQ response-renormalization procedure}
\label{alg:deq-cmr-exact}
\KwIn{Map \(f_\theta\), input \(x\), target \(y\), requested rank \(r\),
cutoff \(\kappa\), base mass \(m_0\), gate
\(\gamma\in\{\mathrm{CMR},\Phi,\Delta\Phi\}\), gate parameters
\((\alpha_{\max},\lambda,c_{\max},\varepsilon_{\rm den})\), and forward,
singular-triplet, and Krylov tolerances}
\KwOut{Equilibrium \(z^R\), gated adjoint \(v_{\rm gate}\), parameter
gradient, and numerical diagnostics}
Solve \(z_{\rm ref}=f_\theta(z_{\rm ref},x)\) and verify the forward residual\;
Define \(K_{\rm ref}w=w-[\partial_zf_\theta(z_{\rm ref},x)]w\) by a JVP\;
Define \(K_{\rm ref}^\top w=w-[\partial_zf_\theta(z_{\rm ref},x)]^\top w\)
by a VJP\;
Estimate the \(r\) smallest triplets of \(K_{\rm ref}=U\Sigma V^\top\) with
dense or iterative partial SVD\;
Sort the selected denominators and collect
\(U_r=[u_1,\ldots,u_r]\), \(V_r=[v_1,\ldots,v_r]\)\;
Apply the triplet-residual and rank-coverage conditions from
Algorithm~\ref{alg:cmr-backward}, stopping if either condition is unresolved\;
\eIf{\(\gamma=\mathrm{CMR}\)}{
  Set \(m_i=m_0\) for \(i=1,\ldots,r\)\;
}{
  Set \(m_i^\Phi=m_0(1+m_0/(\sigma_i+m_0))\)\;
  \eIf{\(\gamma=\Phi\)}{
    Set \(m_i=m_i^\Phi\)\;
  }{
    Compute \(g_{\rm ref}=\nabla_z\mathcal L(z_{\rm ref},y)\),
    \(\mathcal C_r=\{j\leq r:\sigma_j<\kappa\}\), and
    \(\widehat a_{\mathcal C}=\sum_{j\in\mathcal C_r}
    (v_j^\top g_{\rm ref})^2/
    (\|g_{\rm ref}\|_2^2+\varepsilon_{\rm den})\)\;
    Set \(p_{\mathcal C}=\mathcal P_{[0,1]}
    ((\kappa-\min_{j\in\mathcal C_r}\sigma_j)/\kappa)\) when
    \(\mathcal C_r\neq\varnothing\), otherwise set
    \(p_{\mathcal C}=\widehat a_{\mathcal C}=0\)\;
    Set \(s_{\mathcal C}=\mathcal P_{[-1,1]}(2\widehat a_{\mathcal C}-1)\)\;
    Set \(m^{\Delta\Phi}=\mathcal P_{[m_0,c_{\max}m_0]}
    (m_0[1+(\alpha_{\max}-1)p_{\mathcal C}]
    +m_0\lambda p_{\mathcal C}s_{\mathcal C})\)\;
    Set \(m_i=\max(m_i^\Phi,m^{\Delta\Phi})\) for the field gate\;
  }
}
Set \(\widehat\sigma_i=m_i\) when \(\sigma_i<\kappa\) and
\(\sigma_i<m_i\), otherwise set \(\widehat\sigma_i=\sigma_i\)\;
Set \(\delta_i=\widehat\sigma_i-\sigma_i\geq0\) and
\(\Delta K=U_r\diag(\delta_i)V_r^\top\)\;
\eIf{the modified-forward route is active}{
  Solve \(z^R=f_\theta(z^R,x)-\Delta K(z^R-z_{\rm ref})\)\;
}{
  Set \(z^R=z_{\rm ref}\) and label the backward response as a surrogate\;
}
Compute \(g=\nabla_z\mathcal L(z^R,y)\), define
\(Kw=w-[\partial_zf_\theta(z^R,x)]w\), and set \(K^R=K+\Delta K\)\;
Solve \((K^R)^\top v_{\rm gate}=g\) using dense algebra or GMRES with
\((K^R)^\top w=K^\top w+V_r\diag(\delta_i)U_r^\top w\)\;
Return \(\nabla_\theta^R\mathcal L=\partial_\theta\mathcal L+
(\partial_\theta f_\theta(z^R,x))^\top v_{\rm gate}\)\;
Compute the original and modified residuals, denominators, shifts, solver
counts, and, when available, exact-adjoint response diagnostics\;
Refresh the frozen anchor \((z_{\rm ref},U_r,V_r,\delta)\) only after the
optimizer update\;
\end{algorithm2e}

\clearpage
\begin{algorithm2e}[H]
\LinesNumbered
\DontPrintSemicolon
\caption{SILVA local--global CMR, Phi-CMR, or Delta-Phi adjoint}
\label{alg:silva-cmr-exact}
\KwIn{Local matrix \(K_L\), factors \(A,B\), source \(g\), cutoff \(\kappa\),
base mass \(m_0\), gate
\(\gamma\in\{\mathrm{CMR},\Phi_{\rm mode},\Phi_{\rm coll},\Delta\Phi\}\), parameters
\((\alpha_{\max},\lambda,c_{\max},\varepsilon_{\rm den})\), and local-solve
tolerances}
\KwOut{SILVA-shaped gated response \(v_{\rm gate}\), collective spectrum,
effective denominators, residuals, and solver diagnostics}
\For{\(j=1,\ldots,r\)}{
  Solve \(K_LX_{:j}=A_{:j}\) by local GMRES and record its residual\;
}
Form the collective denominator \(\Gamma=I_r-B^\top X\)\;
Compute the dense SVD \(\Gamma=P\diag(s_i)Q^\top\) and verify the triplet residuals\;
Solve \(K_L^\top b=g\) by local GMRES and set \(h=A^\top b\)\;
Set \(\mathcal C=\{i:s_i<\kappa\}\) and
\(p_{\mathcal C}=\mathcal P_{[0,1]}
((\kappa-\min_{i\in\mathcal C}s_i)/\kappa)\) when
\(\mathcal C\neq\varnothing\), otherwise set \(p_{\mathcal C}=0\)\;
Compute \(a_{\mathcal C}=\sum_{i\in\mathcal C}(q_i^\top h)^2/
(\|h\|_2^2+\varepsilon_{\rm den})\), where \(q_i=Q_{:i}\)\;
\For{\(j=1,\ldots,r\)}{
  Solve \(K_L^\top Y_{:j}=B_{:j}\) by local GMRES and record its residual\;
}
\uIf{\(\gamma=\mathrm{CMR}\)}{
  Set \(m_i=m_0\) for \(i=1,\ldots,r\)\;
}
\uElseIf{\(\gamma=\Phi_{\rm mode}\)}{
  Set \(m_i=m_0(1+m_0/(s_i+m_0))\) for \(i=1,\ldots,r\)\;
}
\Else{
  Set \(m^\Phi=m_0[1+(\alpha_{\max}-1)p_{\mathcal C}]\)\;
  \eIf{\(\gamma=\Phi_{\rm coll}\)}{
    Set \(m_i=m^\Phi\) for \(i=1,\ldots,r\)\;
  }{
    Set \(s_{\mathcal C}=\mathcal P_{[-1,1]}(2a_{\mathcal C}-1)\)\;
    Set \(m^{\Delta\Phi}=\mathcal P_{[m_0,c_{\max}m_0]}
    (m^\Phi+m_0\lambda p_{\mathcal C}s_{\mathcal C})\)\;
    Set \(m_i=m^{\Delta\Phi}\) for \(i=1,\ldots,r\)\;
  }
}
Set \(\widehat s_i=m_i\) only when \(s_i<\kappa\) and \(s_i<m_i\), otherwise
set \(\widehat s_i=s_i\)\;
Form \(\Gamma_{\rm eff}=P\diag(\widehat s_i)Q^\top\) and
\(x_R=\Gamma_{\rm eff}^{-\top}h\)\;
Set \(v_{\rm gate}=b+Yx_R\)\;
Compute the original residual
\(\rho_0=\|(K_L-AB^\top)^\top v_{\rm gate}-g\|_2\)\;
Compute the lifted collective residual
\(\rho_R=\|\Gamma_{\rm eff}^\top x_R-h\|_2\) and verify
\(\rho_R\) and every local-solve residual against their tolerances\;
Retain \(s_i\), \(\widehat s_i\), lifted rank, susceptibility
\(1/\min_i\widehat s_i\), \(p_{\mathcal C}\), \(a_{\mathcal C}\), \(m_i\),
local residuals, Krylov iterations, solve count, and matrix--vector products\;
\If{no collective denominator is lifted}{
  Verify that \(v_{\rm gate}\) matches the unmodified SILVA adjoint\;
}
Backpropagate \(v_{\rm gate}\) through the common SILVA equilibrium map and
return the response and diagnostics\;
\end{algorithm2e}
\clearpage
\section{Phi and Source-Conditioned Finite Response}
\label{app:phi-derivation}
\label{app:delta-phi-derivation}

\paragraph{Finite-response condition.}
The renormalization condition fixes a target susceptibility
\(G(0)=\chi_R>0\). Since a scalar static response is \(G(0)=m_R^{-1}\), the
response mass is \(m_R=\chi_R^{-1}\). To make the lift-only consequence
explicit, consider one singular response channel with denominator
\(\sigma\geq0\). We restrict the correction to a nonnegative lift, with
\([x]_+=\max(x,0)\), which gives
\[
 m_R=\chi_R^{-1},\qquad
 \delta m=[m_R-\sigma]_+,\qquad
 \sigma^{\rm eff}=\max(\sigma,m_R).
\]
The resulting response is
\[
 G_R(0)=\frac{1}{\sigma^{\rm eff}}
 =\min\!\left(\frac{1}{\sigma},\chi_R\right).
\]
The target therefore acts as an upper bound on susceptibility. A channel whose
original response already lies below this bound is unchanged, whereas a channel
whose response exceeds it is reduced to the finite target. This distinction is
why the construction controls selected amplification without imposing damping
on every response direction.

\paragraph{Bounded Phi-CMR parameterization.}
Rather than choosing a separate susceptibility for every channel, Phi-CMR
uses the bounded monotone target that we prescribe as
\[
 m_i^\Phi=m_0\left(1+\frac{m_0}{\sigma_i+m_0}\right),
 \qquad m_0\leq m_i^\Phi\leq2m_0,
 \qquad \frac{dm_i^\Phi}{d\sigma_i}<0.
\]
Here \(m_0>0\) sets the base denominator scale and has the same units as
\(\sigma_i\). The dimensionless factor \(m_0/(\sigma_i+m_0)\) is largest near
the pole and decreases smoothly as the channel becomes better conditioned. In
particular,
\[
 m_i^\Phi(0)=2m_0,\qquad
 m_i^\Phi(m_0)=\frac32m_0,\qquad
 \lim_{\sigma_i\to\infty}m_i^\Phi=m_0.
\]
For \(\sigma_i\geq0\), the bounds and monotonicity follow directly from
\[
 0<\frac{m_0}{\sigma_i+m_0}\leq1,
 \qquad
 \frac{\partial m_i^\Phi}{\partial\sigma_i}
 =-\frac{m_0^2}{(\sigma_i+m_0)^2}<0.
\]
It therefore applies the strongest finite lift to the smallest selected
denominator and approaches the base mass continuously away from the pole.
This rational form is a prescribed smooth interpolation satisfying positivity,
boundedness, and stronger action near smaller denominators. It is not a unique
choice and its parameters are not automatically learned. Moreover, a target
mass does not by itself imply a modification. A channel is lifted only when
\(\sigma_i<\kappa\) and \(m_i^\Phi>\sigma_i\). The threshold \(\kappa\) identifies the
candidate critical set while the positive-part rule determines whether a
nonzero correction is actually applied.

\paragraph{Collective pole pressure.}
For a collective pole,
\[
 p_{\mathcal C}=\mathcal P_{[0,1]}
 \left(\frac{\kappa-\sigma_{\mathcal C}}{\kappa}\right),
 \qquad
 m_{\mathcal C}^\Phi=m_0[1+(\alpha_{\max}-1)p_{\mathcal C}].
\]
Here \(\mathcal P_{[0,1]}\) denotes projection onto \([0,1]\), and
\(p_{\mathcal C}\) is a prescribed dimensionless measure of proximity to the
critical threshold that is zero when \(\sigma_{\mathcal C}\geq\kappa\),
increases as \(\sigma_{\mathcal C}\) decreases within the critical region, and
approaches one as \(\sigma_{\mathcal C}\) approaches zero. The collective
response mass consequently interpolates from \(m_0\) at the critical threshold
to \(\alpha_{\max}m_0\) as the collective denominator approaches zero.
Normalizing by \(\kappa\) expresses the depth of the pole relative to the same
resolution used to define the critical set. The parameter
\(\alpha_{\max}\geq1\) bounds the adaptive range, and
\(\alpha_{\max}=1\) recovers the fixed collective mass \(m_0\). When no
collective singular value lies below \(\kappa\), the convention
\(p_{\mathcal C}=0\) is used and no collective channel is lifted.

\paragraph{Hartree motivation and its scope.}
The finite-mass interpretation follows from a Hartree stationary response
\citep{baym1962self,cornwall1974effective}. In a scalar quartic closure, let
\(m\) denote the bare denominator, \(g_4\geq0\) the quartic coupling,
\(D\geq0\) the fluctuation scale, and \(\phi\) the scalar fluctuation. The
self-consistent relations are
\(m_R=m+3g_4\langle\phi^2\rangle\) and
\(\langle\phi^2\rangle=D/m_R\). Substitution of the second relation into the
first and multiplication by \(m_R\) give
\[
 m_R^2-mm_R-3g_4D=0,
 \qquad
 m_R=\frac{m+\sqrt{m^2+12g_4D}}{2}>0.
\]
The second quadratic root is nonpositive and is excluded by the requirement of
a positive susceptibility. At \(m=0\), the admissible root remains finite at
\(m_R=\sqrt{3g_4D}\), illustrating how self-consistency can replace a vanishing
bare denominator by a positive effective one. This calculation supplies the
structural motivation for a positive response mass. It does not identify
\(g_4\) or \(D\) for a DEQ, nor does it uniquely determine the rational
Phi-CMR law, \(m_0\), \(\kappa\), or \(\alpha_{\max}\).

\paragraph{Source-conditioned extension.}
Source conditioning follows from the critical source energy. Phi-CMR depends
on the denominator geometry but not on whether the loss excites the
corresponding directions, and Delta-Phi adds this information through the
critical source fraction
\[
 a_{\mathcal C}=
 \frac{\sum_{i\in\mathcal C}(v_i^\top g)^2}
 {\sum_i(v_i^\top g)^2+\varepsilon_{\rm den}},
 \qquad s_{\mathcal C}=\mathcal P_{[-1,1]}(2a_{\mathcal C}-1).
\]
Because the right singular vectors are orthonormal, the numerator is the
squared norm of the loss source projected onto the critical right-singular
subspace. The denominator is the corresponding total modal source energy, with
\(\varepsilon_{\rm den}>0\) preventing division by zero. Hence
\(a_{\mathcal C}\) is close to zero when the critical subspace is nearly
invisible to the loss and close to one when most source energy lies in that
subspace. The centered quantity \(s_{\mathcal C}\) is negative below one-half,
zero at one-half, and positive above one-half, with projection protecting the
admissible interval against numerical roundoff.

Let \(\lambda\geq0\) control the source-dependent adjustment and let
\(c_{\max}\geq1\) cap the admissible mass relative to \(m_0\).
Define the constrained potential
\[
 \Delta\Phi_{\mathcal C}(m)=\frac12
 [m-m^\Phi-m_0\lambda p_{\mathcal C}s_{\mathcal C}]^2,
 \qquad m\in[m_0,c_{\max}m_0].
\]
Its unconstrained stationary point is
\(m_{\rm unc}=m^\Phi+m_0\lambda p_{\mathcal C}s_{\mathcal C}\), and projection
onto the admissible interval yields
\[
 m^{\Delta\Phi}=\mathcal P_{[m_0,c_{\max}m_0]}(m_{\rm unc}).
\]
Before projection, increasing source visibility raises the target mass when
\(p_{\mathcal C}>0\), whereas low visibility lowers it relative to the
source-independent Phi-CMR target. The adjustment vanishes when the pole
pressure is zero or when \(a_{\mathcal C}=1/2\). Projection prevents either
case from crossing the positive floor or the prescribed upper cap. This
quadratic potential is a transparent parameterization of the gate, not an
independently derived physical effective action.
The correction vanishes when the pole pressure is zero, changes monotonically
with source visibility before clipping, and cannot remove the positive response
floor.

\paragraph{Lift and no-lift cases.}
The same shift has the equivalent closed form
\[
 \delta_i=\mathbf 1_{\{\sigma_i<\kappa\}}[m_i-\sigma_i]_+,\qquad
 K^R=K+U_{\mathcal C}\diag(\delta_i)V_{\mathcal C}^\top.
\]
Hence \(\delta_i=0\) leaves the corresponding singular channel unchanged, and
an empty active set recovers \(K^R=K\) and the exact implicit adjoint.

The Delta-Phi mass is the solution of a one-dimensional constrained
minimization. With
\(m_{\rm unc}=m^\Phi+m_0\lambda p_{\mathcal C}s_{\mathcal C}\), the
Lagrangian for \(\min_m\frac12(m-m_{\rm unc})^2\) on
\([m_0,c_{\max}m_0]\) is
\[
 \mathcal J(m,\eta_-,\eta_+)=\frac12(m-m_{\rm unc})^2
 +\eta_-(m_0-m)+\eta_+(m-c_{\max}m_0).
\]
Stationarity, complementary slackness, and nonnegative multipliers require
\[
 m-m_{\rm unc}-\eta_-+\eta_+=0,
 \quad \eta_-(m_0-m)=0,
 \quad \eta_+(m-c_{\max}m_0)=0,
 \quad \eta_-,\eta_+\geq0.
\]
The Karush--Kuhn--Tucker conditions for the constrained minimization give
\[
 m^{\Delta\Phi}=
 \begin{cases}
 m_0,&m_{\rm unc}<m_0,\\
 m_{\rm unc},&m_0\leq m_{\rm unc}\leq c_{\max}m_0,\\
 c_{\max}m_0,&m_{\rm unc}>c_{\max}m_0.
 \end{cases}
\]
The interior case retains the unconstrained target, while the other two cases
select the nearest admissible endpoint. This establishes that the clipping rule
is the exact minimizer of the stated one-dimensional problem.

Because every active effective denominator is at least \(m_0\), the critical
response obeys the finite bound
\[
 \|v_{\mathcal C}^R\|_2^2
 =\sum_{i\in\mathcal C}\frac{|v_i^\top g|^2}{\widehat\sigma_i^2}
 \leq\frac{\|V_{\mathcal C}^\top g\|_2^2}{m_0^2},
\]
while the stable contribution remains
\(U_{\mathcal S}\diag(\sigma_i^{-1})V_{\mathcal S}^\top g\).
The bound separates the two ingredients needed for problematic amplification.
Large pole pressure indicates a small denominator, while large
\(a_{\mathcal C}\) indicates that the loss excites its singular subspace. A
near-pole direction with negligible source projection contributes little to the
adjoint, and a strongly aligned but well-conditioned direction does not require
a pole correction. The strongest intervention is reserved for cases in which
both quantities are large. Increasing \(m_0\) lowers the maximum critical gain
but can introduce more critical-sector bias, while increasing \(\kappa\)
admits more candidate directions and increasing \(\alpha_{\max}\), \(\lambda\),
or \(c_{\max}\) enlarges the available adaptive range.

The corresponding operator dimensions are
\[
 z^\star,g,v\in\mathbb R^d,\qquad
 U_{\mathcal C},V_{\mathcal C}\in\mathbb R^{d\times r_{\mathcal C}},
 \qquad \Delta K\in\mathbb R^{d\times d}.
\]
The matrix-free representation stores only the two bases and the
\(r_{\mathcal C}\) shifts. A residual JVP returns \(Kw\), a residual VJP
returns \(K^\top w\), and the counterterm actions are
\(\Delta Kw=U_{\mathcal C}\diag(\delta_i)V_{\mathcal C}^\top w\) and
\(\Delta K^\top w=V_{\mathcal C}\diag(\delta_i)U_{\mathcal C}^\top w\).
Thus the dense and matrix-free procedures evaluate the same lifted operator
when the requested triplets and Krylov solves meet their stated tolerances,
although this storage identity is not an end-to-end complexity or runtime-scaling result.

The projected stable solve in Algorithm~\ref{alg:cmr-backward} follows directly
from the singular decomposition. With
\(P_U=I-U_{\mathcal C}U_{\mathcal C}^\top\) and
\(P_V=I-V_{\mathcal C}V_{\mathcal C}^\top\),
\[
 P_VK^\top P_U
 =V_{\mathcal S}\diag(\sigma_i)U_{\mathcal S}^\top.
\]
Consequently,
\[
 v_{\mathcal S}
 =U_{\mathcal S}\diag(\sigma_i^{-1})V_{\mathcal S}^\top g,
 \qquad
 v_{\mathcal C}
 =U_{\mathcal C}\diag(\widehat\sigma_i^{-1})
 V_{\mathcal C}^\top g.
\]
Substitution of \(v_R=v_{\mathcal S}+v_{\mathcal C}\) into the lifted
adjoint equation gives
\[
 (K+\Delta K)^\top v_R
 =V_{\mathcal S}V_{\mathcal S}^\top g+
 V_{\mathcal C}V_{\mathcal C}^\top g=g.
\]
The lifted residual therefore vanishes to linear-solve tolerance. The
residual against the original equation is
\[
 K^\top v_R-g
 =-V_{\mathcal C}
 \diag\!\left(\frac{\delta_i}{\widehat\sigma_i}\right)
 V_{\mathcal C}^\top g,
\]
which measures the intended change in the selected response rather than a
failed solve.
Consequently, the lifted-equation residual measures numerical solution
accuracy, whereas the residual against the original equation measures the
strength of the deliberate response modification. In the result panels, stable
retention near one verifies selectivity, reduced critical amplitude records the
intended control of the pole, and a nonzero original-equation residual should
not be interpreted as failure of the renormalized solve.

\paragraph{Connection to the reported parameter scales.}
The controlled sweeps use target gaps \(0.005\), \(0.01\), and \(0.02\), a
representative stable denominator \(0.125\), \(m_0=0.03\), and
\(\kappa=0.08\). The ordering
\[
 0.005\leq\sigma_{\mathcal C}\leq0.02<m_0<\kappa<0.125
\]
places every constructed critical denominator inside the candidate set and
keeps the stable denominator outside it. At
\(\sigma_{\mathcal C}=0.005\) and \(\alpha_{\max}=2\),
\[
 p_{\mathcal C}=\frac{0.08-0.005}{0.08}=0.9375,
 \qquad
 m_{\mathcal C}^\Phi=0.03(1+0.9375)=0.058125,
\]
which is the \(0.0581\) effective denominator reported in the mechanism
experiment. The native and cross-benchmark SILVA studies instead hold
\(m_0=0.08\) and \(\kappa=0.12\) fixed across datasets and seeds. Their
per-mode Phi-CMR targets lie between \(0.08\) and \(0.16\), but a measured
collective mode is modified only when it lies below \(0.12\) and its target
mass exceeds its original denominator. Sparse activation therefore means that
Phi-CMR often recovers the implicit adjoint exactly under the observed
conditioning, rather than indicating that the response calculation failed.
These numerical scales specify the evaluated operating regimes and are not
claimed to be universal or uniquely optimal.

\paragraph{Numerical parameter interpretation.}
For a new operator, the response parameters are determined relative to a
resolved small-singular-value spectrum rather than transferred as universal
constants. The requested rank \(r\) is increased until the largest returned
small singular value lies at or above the proposed cutoff. The cutoff
\(\kappa\) is then placed above the resolved near-critical cluster, with the
unresolved interval of Appendix~\ref{app:coordinate-metric} used when classifications
change under tolerance or precision perturbations. Because every active
effective denominator is at least \(m_0\), the base mass imposes a maximum
selected gain \(1/m_0\). It therefore expresses the largest critical response
retained by the calculation. The parameters
\(\alpha_{\max},\lambda,c_{\max}\) control the bounded adaptive range and may
be evaluated on calibration cases through stable-sector retention, critical
response, prediction error, and activation frequency. A valid numerical
realization requires the forward, singular-triplet, local-solve, and
lifted-equation residuals to satisfy their stated tolerances and must recover
the exact implicit adjoint whenever no denominator is lifted. These conditions
separate a deliberate response modification from unresolved spectral or
linear-solve error.
\section{Metrics, Protocol, and Quantitative Evidence}
\label{app:quantitative-metrics}
\label{app:protocols}

For candidate adjoint \(v_m\) and exact adjoint \(v_\star\), define the
amplitude, direction, projection, error, and residual by
\[
 R_m=\frac{\|v_m\|}{\|v_\star\|+\epsilon},\quad
 C_m=\frac{v_m^\top v_\star}{\|v_m\|\|v_\star\|+\epsilon},\quad
 P_m=\frac{v_m^\top v_\star}{\|v_\star\|^2+\epsilon},
\]
\[
 E_m=\frac{\|v_m-v_\star\|}{\|v_\star\|+\epsilon},\qquad
 \rho_m=\|K^\top v_m-g\|.
\]
Stable and critical amplitudes are computed after projection onto
\(U_{\mathcal S}\) and \(U_{\mathcal C}\). After inverting train-only
standardization, the sample score is
\[
 e_i=\frac{\|\widehat y_i-y_i\|_2}
 {\max(\|y_i\|_2,10^{-12})},
\]
with all stored channels and spatial entries flattened before tables average
samples and equally weight family--seed units. Field panels use pointwise
absolute error, while mixed-unit channels make \(e_i\) a descriptive Euclidean
score, not a physical invariant. The descriptive upper-threshold count records
whether the ratio to the paired implicit score is at most 1.05, including cases
where the candidate error is lower. It is not a prespecified noninferiority or
equivalence test. Paired intervals use the five declared seeds, never training
updates, horizons, channels, or examples as independent replicates.

For the SILVA response, let
\(x_R=\Gamma_{\rm eff}^{-\top}h\) and \(v_R=b+Yx_R\). Its residual against the
original full operator is
\(\rho_0=\|(K_L-AB^\top)^\top v_R-g\|_2\), while
\(\rho_R=\|\Gamma_{\rm eff}^\top x_R-h\|_2\) verifies the lifted collective solve.
The local solves for \(X\), \(Y\), and \(b\) are checked separately, preventing
these two quantities from concealing an inaccurate local inverse.

The controlled response sweeps use \(m_0=0.03\) and \(\kappa=0.08\), with the
collective and source-rotation sweeps evaluated at seeds 123 and 456. The collective
sweep sets \(\alpha_{\max}=2\) in Eq.~\eqref{eq:main-phi-collective-law}
(\(\gamma=\Phi_{\rm coll}\) in Algorithm~\ref{alg:silva-cmr-exact}). The
source-rotation sweep sets \(\alpha_{\max}=2.5\) and uses critical source fractions
\(0,\allowbreak 0.05,\allowbreak 0.10,\allowbreak 0.25,\allowbreak
0.50,\allowbreak 0.75,\allowbreak 0.90,\allowbreak 1\). The Delta-Phi
selection uses \(0,\allowbreak 0.10,\allowbreak 0.50,\allowbreak 1\), with
\(\lambda\in\{0,\allowbreak 0.25,\allowbreak 0.50,\allowbreak
0.75,\allowbreak 1\}\),
\(\alpha_{\max}=2.5\), \(c_{\max}=3\), and
\(\varepsilon_{\rm den}=10^{-14}\). It selects \(\lambda\) using seeds 123
and 456 at gaps 0.005 and 0.01, and every configuration with seed 789 or gap 0.02
forms the 20-unit held-out set. The field ablation uses
\(m_0=0.08\), \(\kappa\in\{0.08,0.12\}\),
\(\lambda\in\{0,1,4\}\), and \(c_{\max}\in\{3,5\}\), with the per-mode Phi
mass combined with the scalar Delta-Phi mass as stated after
Eq.~\eqref{eq:delta-phi}.

\begin{table}[H]
\centering
\scriptsize
\setlength{\tabcolsep}{3.5pt}
\caption{Evaluation blocks and statistical units.}
\label{tab:protocol-grid}
\begin{tabular}{@{}P{0.19\textwidth}P{0.22\textwidth}P{0.52\textwidth}@{}}
\toprule
Block & Unit & Fixed design\\
\midrule
Controlled response & 36 collective, 48 source-rotation, 20 held-out Delta-Phi configurations &
Dimensions 12--24, ranks 1--2, gaps 0.005--0.02, prescribed source fractions,
double precision, triplet tolerance \(10^{-10}\), and GMRES relative tolerance
\(10^{-11}\).\\
Native and cross-benchmark SILVA & 70 and 65 family/regime--seed model units, 1,080 method evaluations &
Five seeds, eight backward modes, disjoint dataset-provided splits, identical forward
models and ordered minibatches within every paired comparison.\\
Capacity extension & 50 exploratory aligned settings and 400 evaluations, plus 40 transient-aligned evaluations &
Six high-error families, hidden widths 96 or 128, resolutions up to \(64^2\)
or \(16^3\), and the same eight backward modes.\\
Transient evaluation & 2,880 physical-time and 945 solver-time repeated evaluations &
Free autoregressive rollouts without retraining and solver horizons
\(1,2,4,8,16,32,64\), compared with converged equilibria and GMRES adjoints.\\
\bottomrule
\end{tabular}
\end{table}

\begin{table}[H]
\centering
\scriptsize
\setlength{\tabcolsep}{3pt}
\caption{Extended quantitative evidence. Negative field-error differences
favor the first method.}
\label{tab:extended-quantitative-evidence}
\begin{tabular}{@{}P{0.25\textwidth}P{0.27\textwidth}rrrr@{}}
\toprule
Block & Metric & \(n\) & A & B & A--B\\
\midrule
Phi-CMR vs Tikhonov & response-norm ratio & 36 & 0.303 & 0.231 & 0.073\\
CMR vs Tikhonov, source sweep & response-norm ratio & 48 & 0.471 & 0.242 & 0.229\\
High vs low source alignment & CMR gain over Tikhonov & 36 & 0.263 & 0.172 & 0.091\\
Phi-CMR vs Tikhonov, Darcy & relative field error & 1 & 1.211 & 1.558 & -0.347\\
Delta-Phi vs Phi-CMR & low-source norm ratio & 20 & 0.668 & 0.623 & 0.045\\
Delta-Phi vs Phi-CMR & high-source suppression & 20 & 0.816 & 0.758 & 0.057\\
Delta/Phi-CMR vs Tikhonov & relative field error & 8 & 2.345 & 2.562 & -0.218\\
\bottomrule
\end{tabular}
\end{table}

\paragraph{Validity boundary.}
\label{app:evidence-boundary}
The method requires a converged equilibrium, accurate local products, resolved
critical subspaces, the relevant loss and parameter projections, and a positive
bounded effective denominator, with no-lift cases recovering the exact implicit
adjoint. Lifted backward-only cases are identified as surrogates, frozen-anchor
modified-forward cases are exact gradients of the stated modified residual,
and physical rollout time remains distinct from DEQ solver time.

\endgroup

\section{Supplementary Pole and Source Diagnostics}
\label{app:extra-figures}

\begin{figure}[H]
\centering
\includegraphics[width=.88\textwidth]{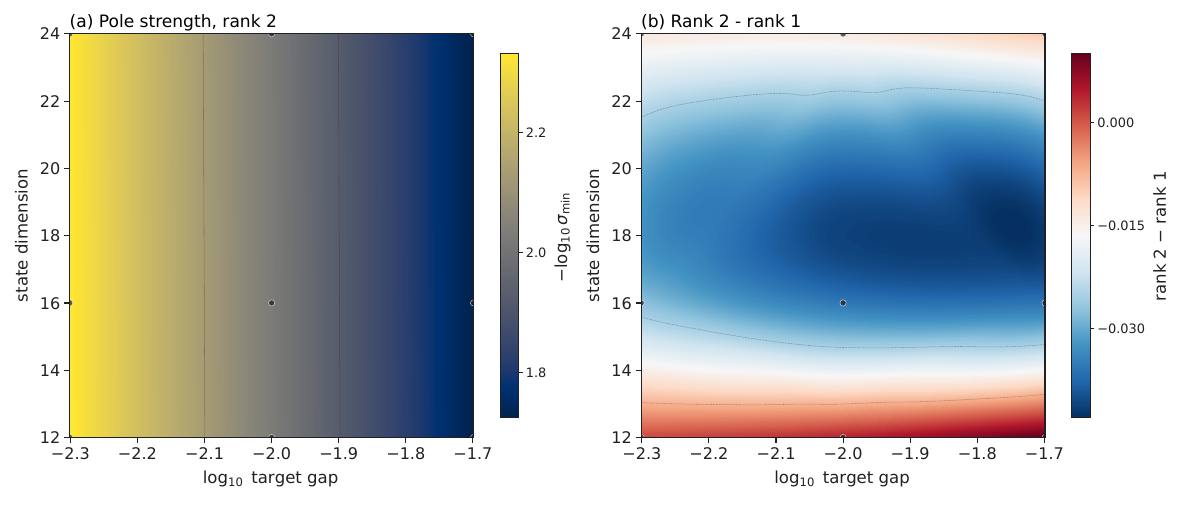}
\caption{Rank-two response with (a) pole strength and (b) the rank-two minus
rank-one Phi-CMR advantage over Tikhonov.}
\label{fig:rank2-pole-extra}
\end{figure}

\begin{figure}[H]
\centering
\includegraphics[width=.88\textwidth]{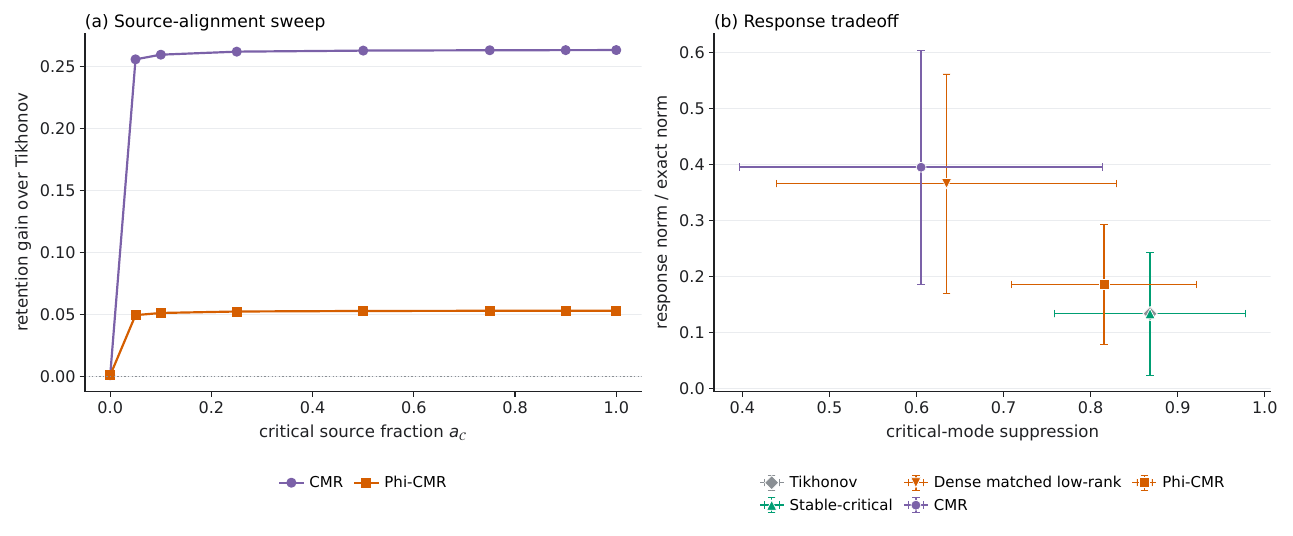}
\caption{Source-alignment experiment with (a) retention gain over Tikhonov and
(b) response norm versus critical-mode suppression.}
\label{fig:source-falsification}
\end{figure}

\begin{figure}[H]
\centering
\includegraphics[width=.88\textwidth]{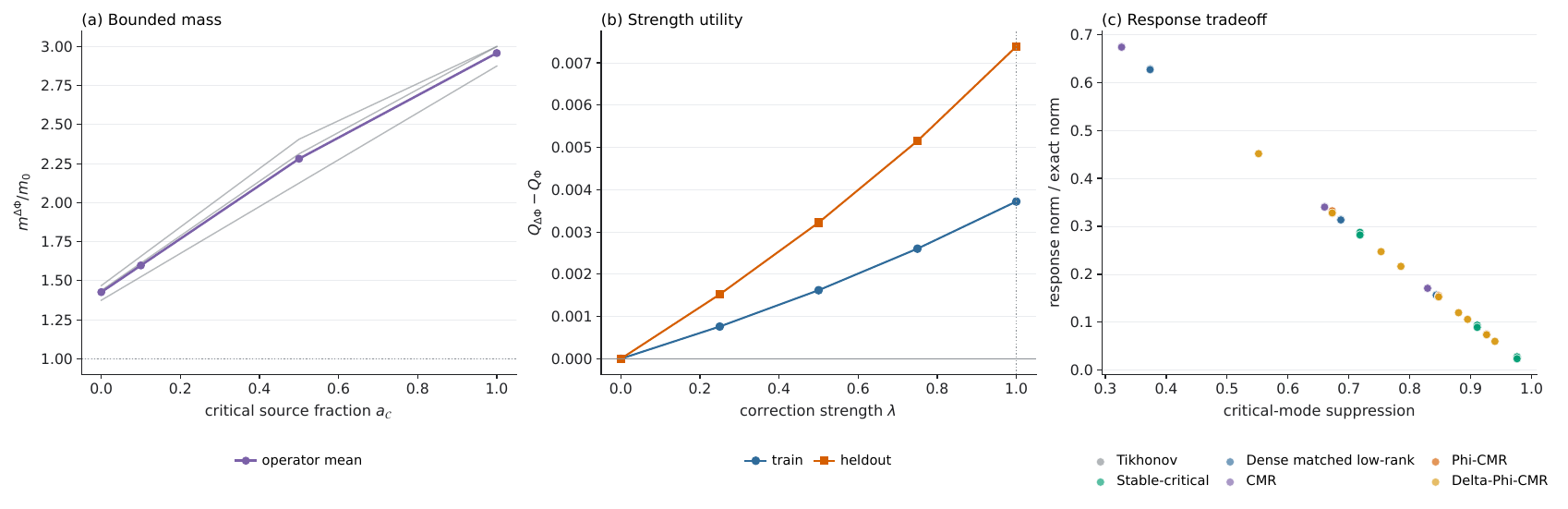}
\caption{Delta-Phi sweep with (a) bounded mass, (b) strength utility, and (c)
response norm versus critical-mode suppression.}
\label{fig:delta-phi-extra}
\end{figure}

\begin{figure}[H]
\centering
\includegraphics[width=.88\textwidth]{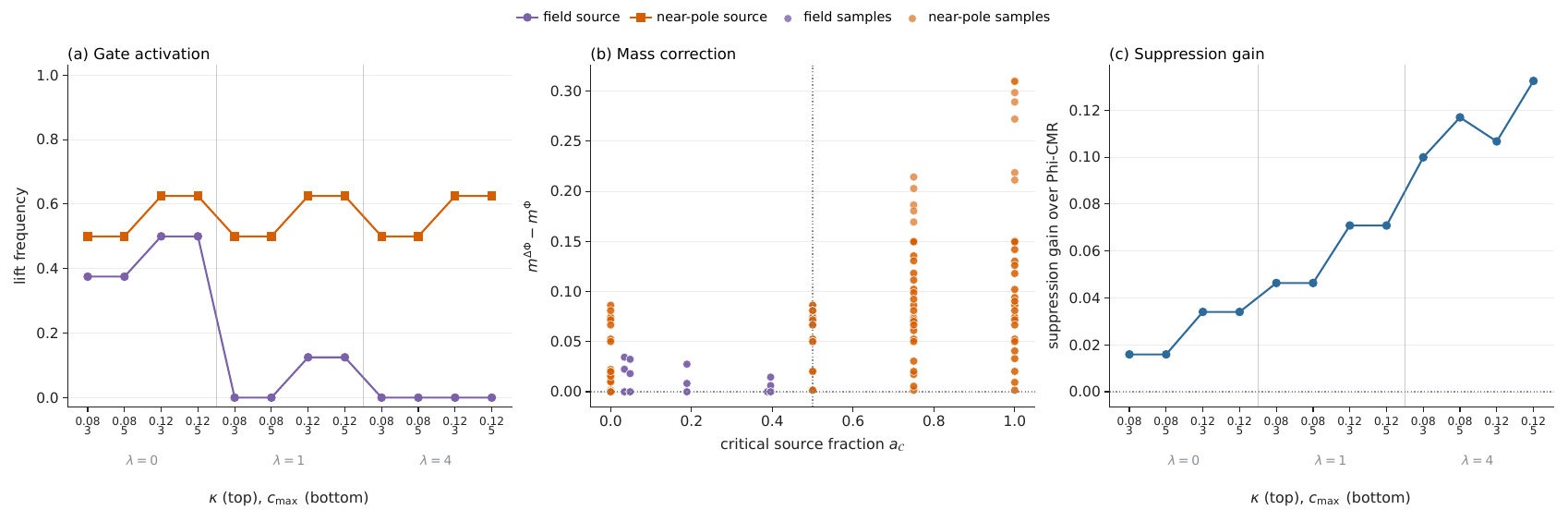}
\caption{Delta-Phi ablation with (a) gate activation, (b) mass correction, and
(c) suppression gain.}
\label{fig:delta-ablation-extra}
\end{figure}
\section{Cross-Benchmark Evaluation of SILVA Response Renormalization}
\label{app:silva-cross-suite}

\newcommand{\silvaatlascaption}[2]{#1 with (a) input, (b) target, (c) Phi-CMR
prediction, (d) absolute error, (e) Tikhonov-minus-Phi-CMR error, (f)
five-seed all-method relative-$L_2$, (g) original and effective collective
denominators with active lifts, and (h) source-weighted singular response. #2}

The cross-benchmark analysis tests whether response renormalization depends on a
particular PDE family or representation by repeating the matched eight-method
SILVA protocol on 13 additional public benchmark datasets comprising four PDEArena
systems \citep{gupta2023pdearena}, six DynaBench systems
\citep{dulny2023dynabench}, PDEGym Poisson--Gauss
\citep{herde2024poseidon}, CFDBench cylinder geometry
\citep{luo2023cfdbench}, and LagrangeBench Taylor--Green vortex particles
\citep{toshev2023lagrangebench}.
This extension changes the governing equations, spatial dimension, operator
type, geometry, and discretization while retaining five seeds, 16 training
minibatches, the same 64-update direct or 128-update residual budget, eight
backward methods, train-only normalization,
disjoint held-out splits, and the same SILVA forward solver and numerical
tolerances at evaluation.

\begin{table}[H]
\centering
\scriptsize
\setlength{\tabcolsep}{4pt}
\caption{Cross-benchmark evaluation. Dynamic entries use one-step training
and seven-step free physical-time rollouts, while Poisson--Gauss is evaluated as a
static source-to-solution operator.}
\label{tab:silva-cross-suite-coverage}
\begin{tabular}{@{}P{0.12\textwidth}P{0.21\textwidth}P{0.32\textwidth}P{0.22\textwidth}@{}}
\toprule
Source & Dataset & Physical setting & Representation\\
\midrule
PDEArena & Navier--Stokes 2D & incompressible flow & multi-field grid\\
PDEArena & Shallow Water 2D & free-surface flow & multi-field grid\\
PDEArena & Maxwell 3D & electromagnetic waves & vector-field volume\\
PDEArena & Kuramoto--Sivashinsky 1D & chaotic dissipative dynamics & scalar grid\\
DynaBench & Advection & linear transport & structured grid\\
DynaBench & Burgers & nonlinear conservation law & structured grid\\
DynaBench & Gas Dynamics & compressible flow & structured grid\\
DynaBench & Kuramoto--Sivashinsky & chaotic dissipative dynamics & structured grid\\
DynaBench & Reaction--Diffusion & coupled reaction and diffusion & structured grid\\
DynaBench & Wave & hyperbolic wave dynamics & structured grid\\
PDEGym & Poisson--Gauss & elliptic boundary-value problem & operator map\\
CFDBench & Cylinder geometry & external incompressible flow & velocity grid\\
Lagrange\-Bench & Taylor--Green vortex 2D & vortex dynamics & moving particles\\
\bottomrule
\end{tabular}
\end{table}

Table~\ref{tab:silva-cross-suite-results} gives the family-weighted static and
horizon-seven comparisons for all eight backward methods, comprising 520
static method evaluations and 3,360 repeated physical-time evaluations. Phi-CMR has mean
error ratios 0.9999 statically and 1.0016 at horizon seven, with ratios at most
1.05 in all 65 static and 57 of 60 horizon-seven paired evaluations, and lower
static error than Tikhonov in 31 of 65 paired comparisons, while CMR gives
ratios 0.9989 and 0.9978 under the same evaluation. Near-identity error ratios
across grids, a vector-field volume, an
operator map, geometry-conditioned flow, and moving particles show that the
response rule transfers without requiring a representation-specific notion of
the critical denominator.

\begin{table}[H]
\centering
\scriptsize
\setlength{\tabcolsep}{4pt}
\caption{All-method cross-benchmark results. Static values use 65 family--seed
pairs, while horizon seven uses 60 pairs from the 12 dynamic datasets. Error and
measured solve-plus-backward time are normalized by exact implicit
differentiation.}
\label{tab:silva-cross-suite-results}
\begin{tabular}{@{}lccccc@{}}
\toprule
Method & Static error & Ratio $\leq1.05$ & Time & Horizon-7 error & Ratio $\leq1.05$\\
\midrule
Implicit & 1.0000 & 65/65 & 1.00 & 1.0000 & 60/60\\
Phantom & 0.9994 & 64/65 & 0.59 & 0.9969 & 60/60\\
Neumann & 0.9975 & 64/65 & 0.54 & 0.9922 & 60/60\\
JFB & 0.9982 & 63/65 & 0.48 & 0.9962 & 60/60\\
SHINE & 0.9818 & 62/65 & 0.52 & 0.9922 & 55/60\\
Tikhonov & 0.9989 & 65/65 & 2.48 & 0.9987 & 58/60\\
CMR & 0.9989 & 65/65 & 1.85 & 0.9978 & 59/60\\
Phi-CMR & 0.9999 & 65/65 & 1.86 & 1.0016 & 57/60\\
\bottomrule
\end{tabular}
\end{table}

\begin{figure}[!htbp]
\centering
\includegraphics[width=.90\textwidth]{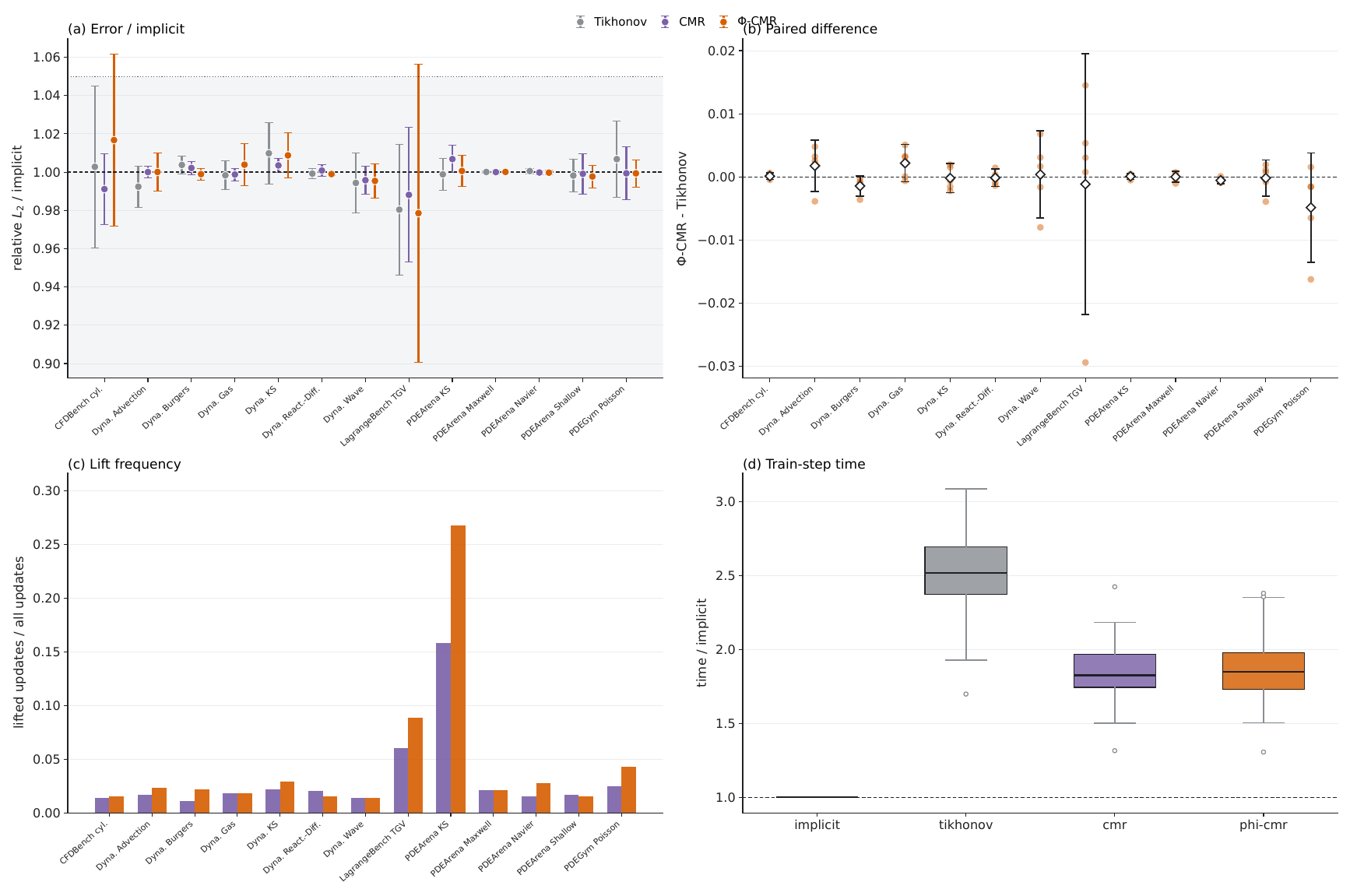}
\caption{Cross-benchmark evaluation of SILVA response renormalization with (a) final inverse-standardized
sample-relative Euclidean error
normalized by exact implicit differentiation, (b) paired
Phi-CMR-minus-Tikhonov differences across seeds, (c) the fraction of training
measurements with an active collective lift, and (d) measured equilibrium-solve
plus backward time relative to implicit differentiation. Shading in panel (a)
denotes error ratios at most 1.05.}
\label{fig:silva-cross-suite-aggregate}
\end{figure}

Figure~\ref{fig:silva-cross-suite-aggregate} disaggregates these averages by
dataset to test whether aggregate fidelity conceals representation-specific
failures. Panel (a) shows task error relative to implicit, panel (b) gives the
paired Phi-CMR--Tikhonov differences and seed intervals, panel (c) shows the
fraction of training measurements on which the collective denominator was
lifted, and panel (d) shows measured cost. The lift occurs at least once in
each dataset while remaining sparse over updates, matching the selective
counterterm in Eq.~\eqref{eq:cmr-adjoint} and showing that near-identity error
ratios do not arise from an always-active global correction.

\begin{figure}[p]
\centering
\includegraphics[width=.90\textwidth]{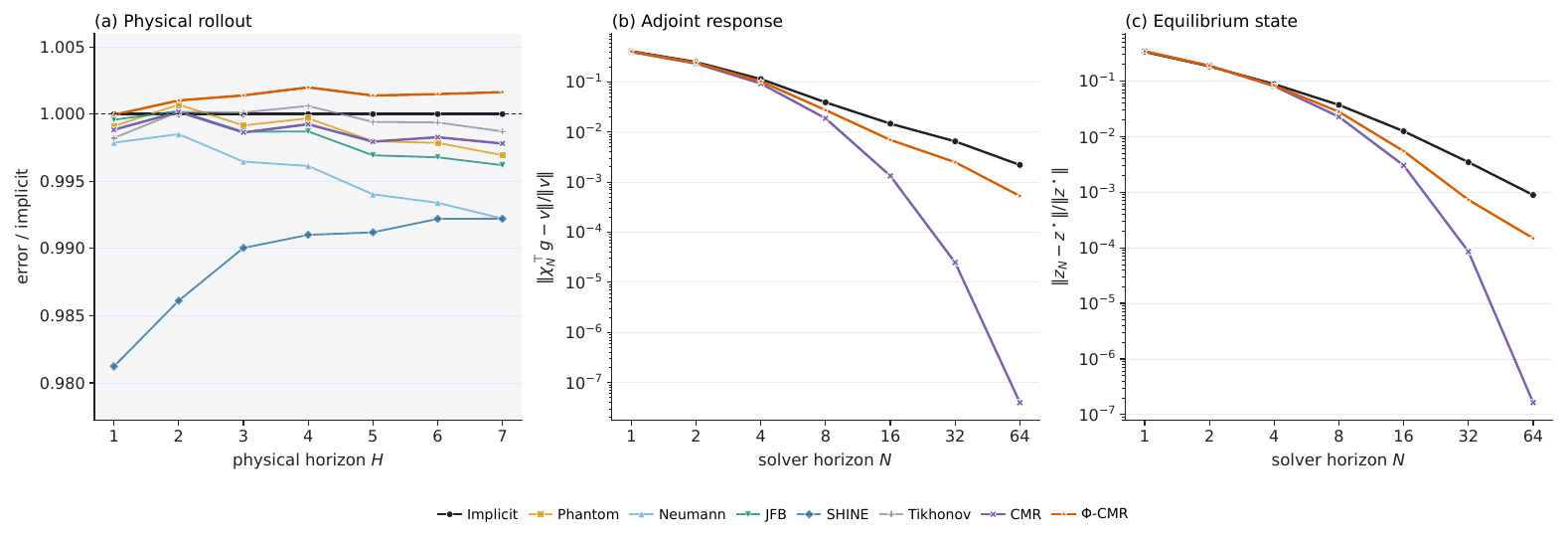}
\caption{Cross-benchmark transient evidence with (a) seven-step free physical-time
rollouts for all eight backward methods on 12 dynamic datasets, (b) finite
causal-response error relative to the static implicit adjoint, and (c) original
DEQ state error relative to equilibrium, where panels (b,c) use implicit, CMR,
and Phi-CMR models at solver horizons
\(N\in\{1,2,4,8,16,32,64\}\). Shading in panel (a) denotes error ratios at most
1.05.}
\label{fig:silva-cross-suite-transient}
\end{figure}

The two time variables derived in Appendix~\ref{app:silva-transient} are tested
separately in Fig.~\ref{fig:silva-cross-suite-transient}, where panel (a)
propagates prediction errors through seven physical PDE steps for all eight
methods and panels (b,c) vary the DEQ solver horizon
\(N\in\{1,2,4,8,16,32,64\}\), reaching at \(N=64\) a Phi-CMR causal-response
error of $5.27\times10^{-4}$ relative to the static adjoint and a state error of
\(1.49\times10^{-4}\) relative to equilibrium over 60 trained models.

\subsection{Forward-capacity extension for the highest-error families}
\label{app:silva-forward-capacity}

The matched compact predictor controls the forward design but leaves visible
approximation error in PDEArena Navier--Stokes, Shallow Water, and Maxwell,
PDEBench Diffusion--Reaction 2D and Darcy, and LagrangeBench Taylor--Green
vortex, motivating a forward-capacity extension that tests whether this error
belongs to the shared representation rather than to response renormalization.
Because held-out trajectories, offsets, equilibrium solver and tolerances, backward
methods, response parameters, and five seeds remain fixed while every method
receives the same larger family-specific forward model, the extension does not
alter CMR or Phi-CMR. Darcy and Navier--Stokes use 128-state two-layer encoders
and readouts with three-layer width-32 2D adapters, Shallow Water uses 96 states
with a three-layer width-24 dilated adapter, and Maxwell reconstructs two
three-component vector fields at $16^3$ with a three-layer width-16 Conv3D
adapter rather than folding depth and vector components into a 2D channel axis.
LagrangeBench uses a 128-state two-layer particle readout, while the fixed
$t_0\!\to t_1$ Diffusion--Reaction stress test uses 96 states with a two-layer
width-16 adapter. All profiles selected on seed 123 remain fixed throughout the
subsequent five-seed, eight-method evaluation, and because the same seed
participates in both model selection and evaluation, this extension is
exploratory and its paired intervals are descriptive rather than confirmatory.

\begin{table}[H]
\centering
\scriptsize
\setlength{\tabcolsep}{4pt}
\caption{Phi-CMR inverse-standardized sample-relative Euclidean error in the forward-capacity
extension, with capacity-to-baseline ratios and paired 95\% intervals for the
capacity-minus-baseline difference.}
\label{tab:silva-capacity-results}
\begin{tabular}{@{}lccccc@{}}
\toprule
Family & Baseline & Capacity & Ratio & Paired delta 95\% CI & Seeds improved\\
\midrule
Darcy 2D & 1.1119 & 0.6483 & 0.583 & $[-0.6957,-0.2314]$ & 5/5\\
Diffusion--Reaction 2D & 1.0170 & 0.9931 & 0.977 & $[-0.0258,-0.0219]$ & 5/5\\
PDEArena Navier--Stokes & 0.6405 & 0.4496 & 0.702 & $[-0.1951,-0.1866]$ & 5/5\\
PDEArena Shallow Water & 0.3629 & 0.3287 & 0.906 & $[-0.0536,-0.0147]$ & 5/5\\
PDEArena Maxwell 3D & 0.9674 & 0.6244 & 0.645 & $[-0.3588,-0.3272]$ & 5/5\\
LagrangeBench TGV & 0.2204 & 0.1252 & 0.573 & $[-0.1235,-0.0668]$ & 5/5\\
\bottomrule
\end{tabular}
\end{table}

\begin{figure}[p]
\centering
\includegraphics[width=.78\textwidth]{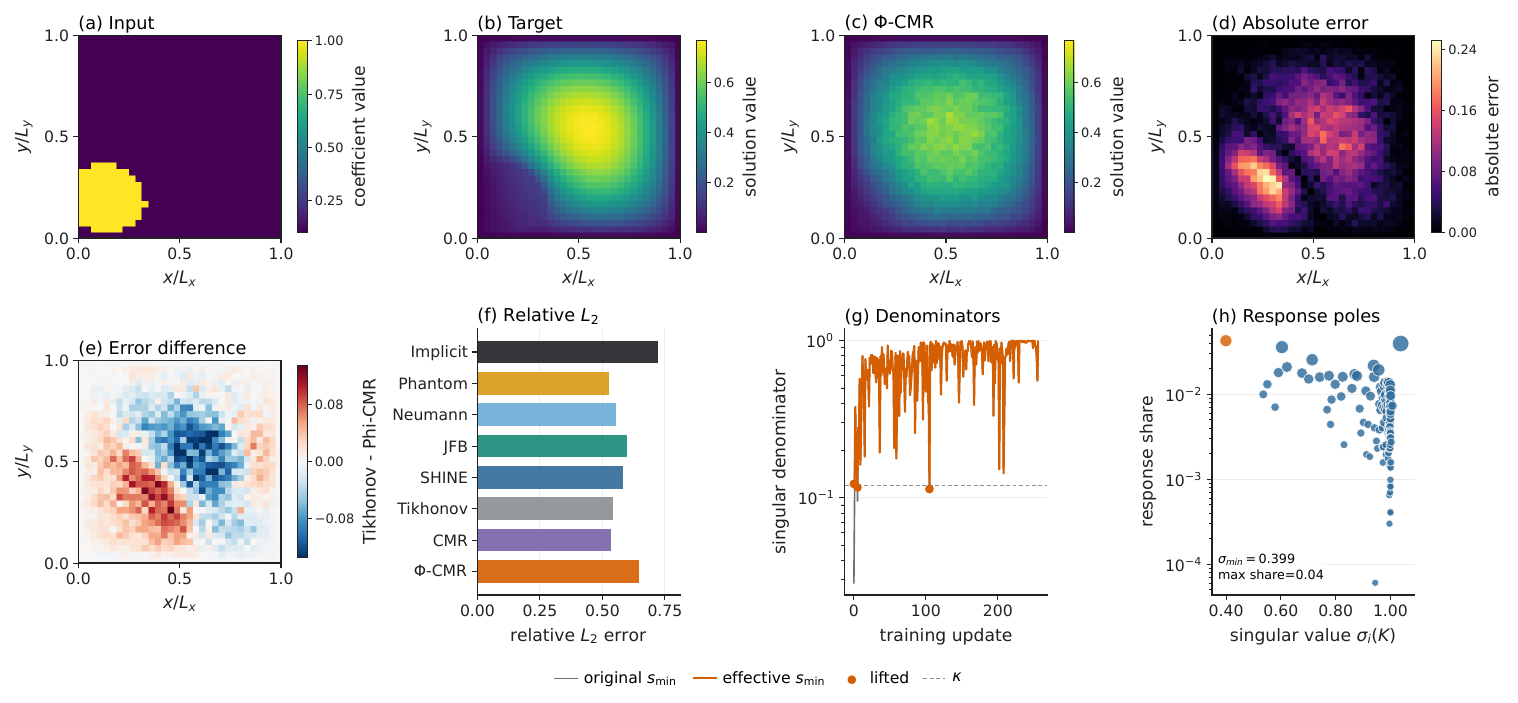}
\caption{Higher-capacity \silvaatlascaption{Darcy 2D}{Panels (a--e) use the
median-error $\beta=1$ example, while panel (f) aggregates all five Darcy regimes.}}
\label{fig:silva-capacity-darcy}
\end{figure}

\begin{figure}[p]
\centering
\includegraphics[width=.78\textwidth]{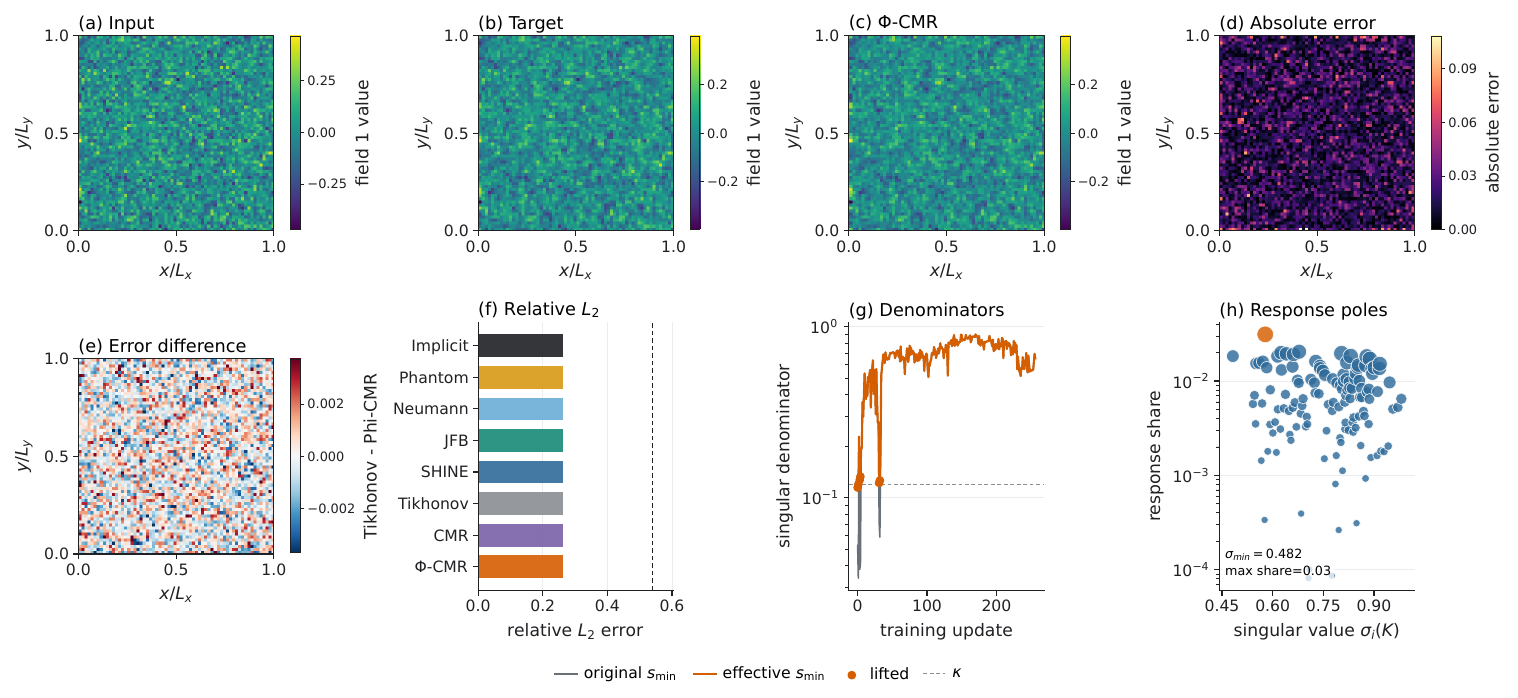}
\caption{Transient-aligned \silvaatlascaption{Diffusion--Reaction 2D}{Field
panels display the first physical channel for the separately evaluated
$t_1\!\to t_2$ task.}}
\label{fig:silva-capacity-diffusion-reaction}
\end{figure}

\begin{figure}[p]
\centering
\includegraphics[width=.78\textwidth]{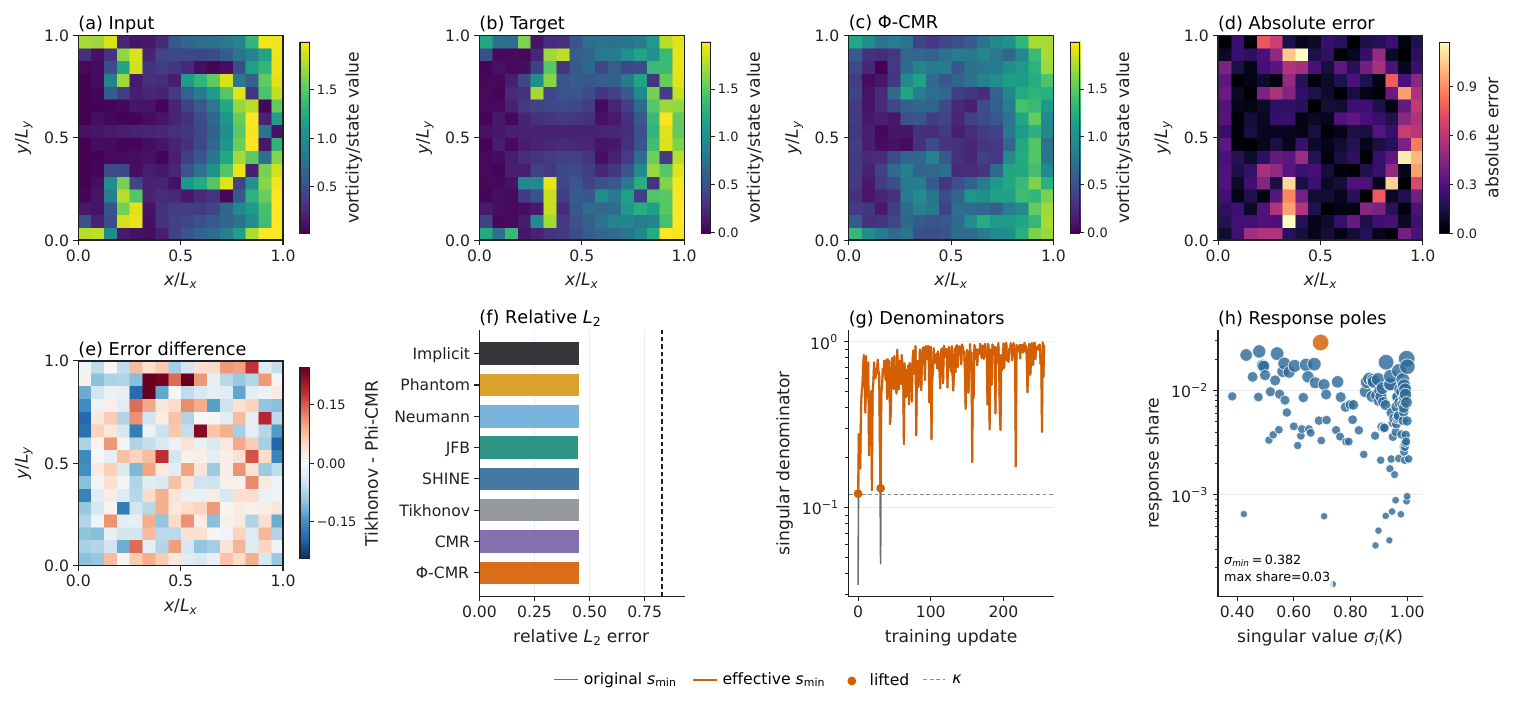}
\caption{Higher-capacity \silvaatlascaption{PDEArena Navier--Stokes 2D}{Field
panels display the first vorticity/state channel.}}
\label{fig:silva-capacity-pdearena-ns}
\end{figure}

\begin{figure}[p]
\centering
\includegraphics[width=.78\textwidth]{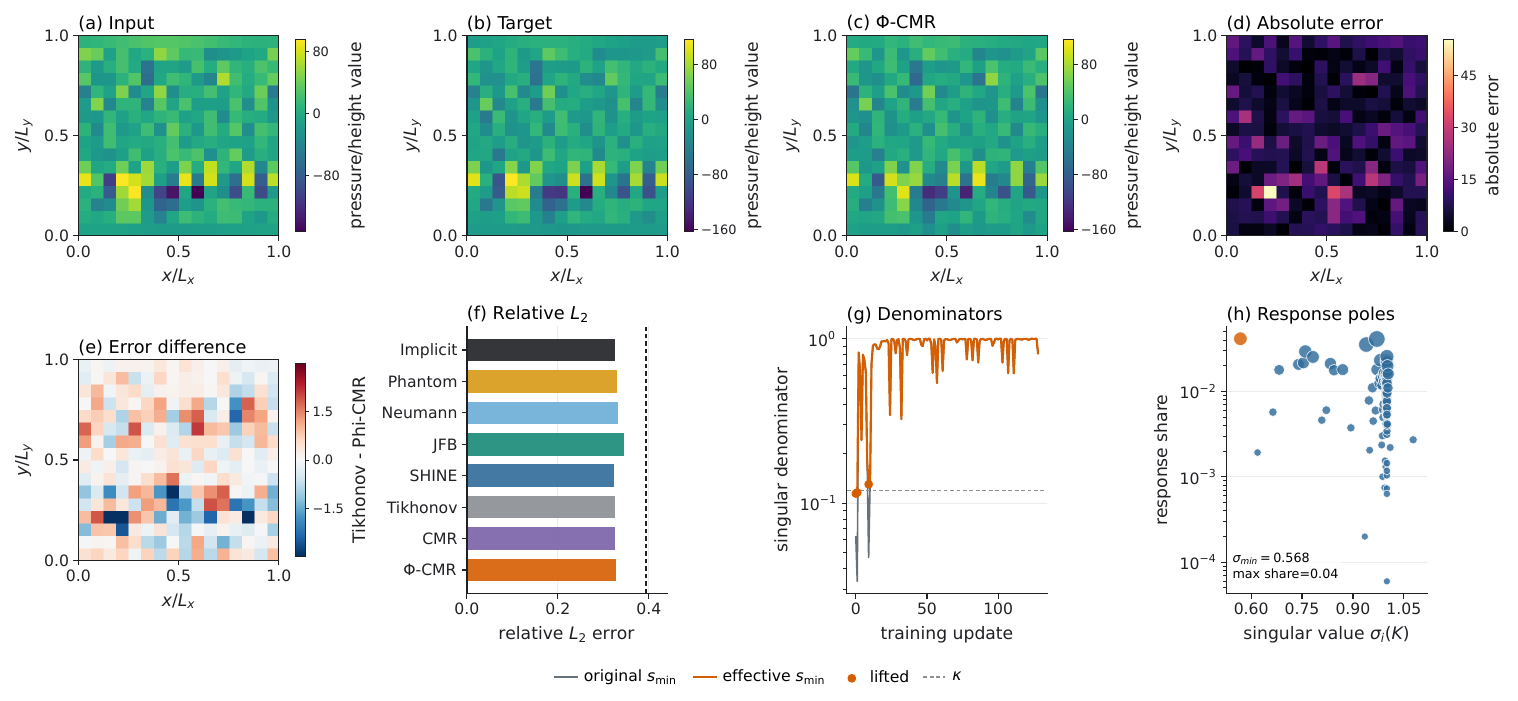}
\caption{Higher-capacity \silvaatlascaption{PDEArena Shallow Water 2D}{Field
panels display the first pressure/height channel.}}
\label{fig:silva-capacity-pdearena-sw}
\end{figure}

\begin{figure}[p]
\centering
\includegraphics[width=.78\textwidth]{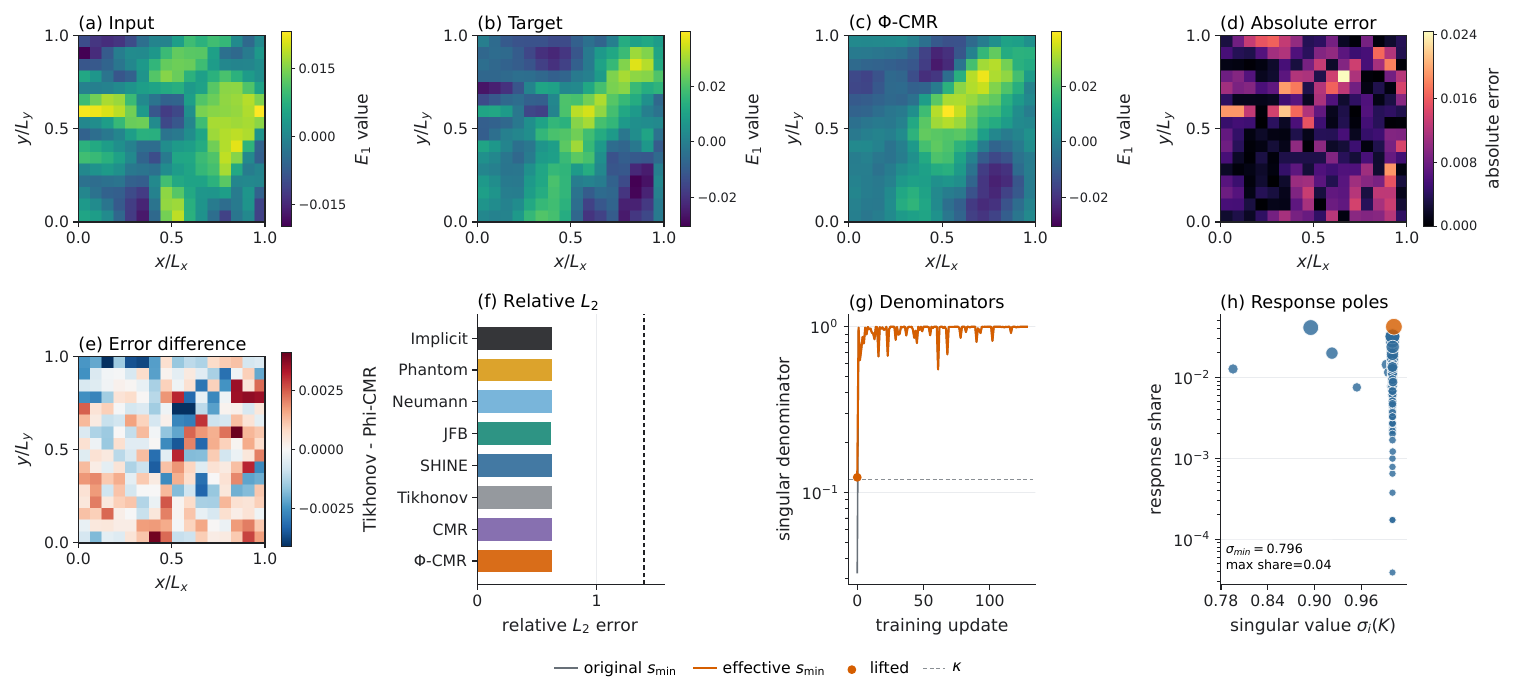}
\caption{Higher-capacity \silvaatlascaption{PDEArena Maxwell 3D}{Field panels
show the central slice of the first electric-displacement component after
vector-aware Conv3D reconstruction.}}
\label{fig:silva-capacity-pdearena-maxwell}
\end{figure}

\begin{figure}[p]
\centering
\includegraphics[width=.88\textwidth]{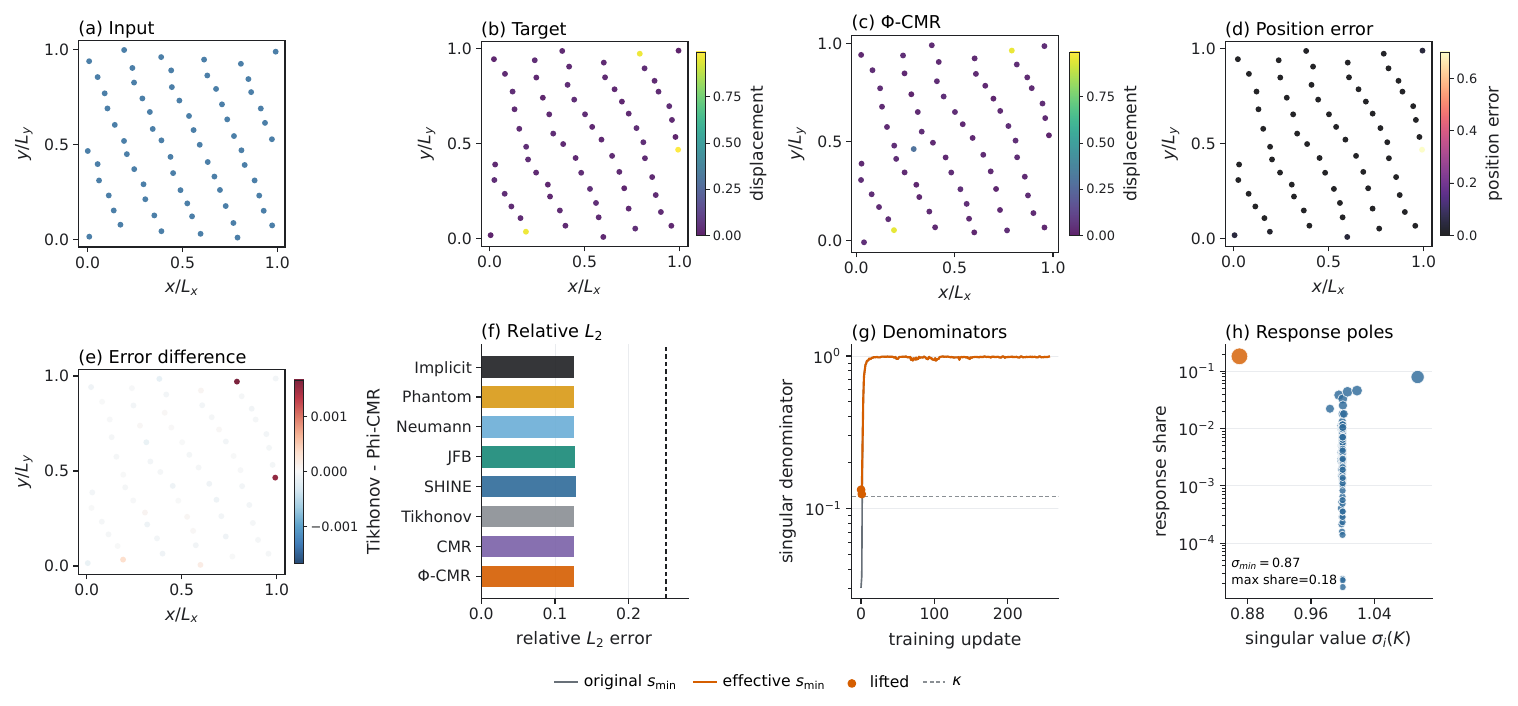}
\caption{Higher-capacity \silvaatlascaption{LagrangeBench Taylor--Green vortex
2D}{Panels (a--e) retain the native particle coordinates.}}
\label{fig:silva-capacity-lagrange}
\end{figure}

Phi-CMR improves in all 30 family--seed comparisons in
Table~\ref{tab:silva-capacity-results}, with the corresponding familywise
prediction geometry shown in
Figs.~\ref{fig:silva-capacity-darcy}--\ref{fig:silva-capacity-lagrange}.
The largest
reductions occur for Darcy and LagrangeBench, while the more modest
Diffusion--Reaction reduction indicates a limitation of the selected
physical-time mapping rather than of the backward response law. These results
attribute a substantial part of the largest absolute errors to forward
representation capacity, while the response-renormalization comparison remains
the matched compact benchmark defined by Tables~\ref{tab:silva-main-summary}
and~\ref{tab:silva-cross-suite-results}.

\begin{figure}[p]
\centering
\includegraphics[width=.82\textwidth]{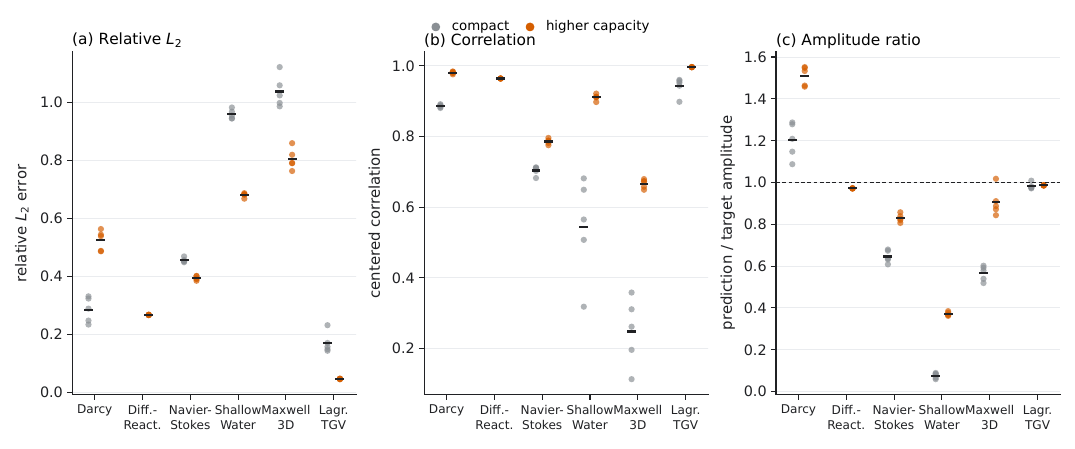}
\caption{Held-out prediction geometry over five seeds with (a)
representative-field relative error, (b) centered spatial correlation, and (c)
prediction-to-target amplitude ratio, where points are individual seeds, short
horizontal marks are means, and Diffusion--Reaction uses the separate
$t_1\!\to t_2$ transient task.}
\label{fig:silva-prediction-alignment}
\end{figure}

Prediction geometry in Fig.~\ref{fig:silva-prediction-alignment} distinguishes
shape recovery from amplitude calibration, which a single relative-error
number cannot separate. On the same first held-out example across five seeds,
the capacity model changes Navier--Stokes correlation from $0.704$ to $0.786$
and amplitude ratio from $0.646$ to $0.830$, while the corrected Maxwell Conv3D
changes them from $0.248$ to $0.666$ and from $0.566$ to $0.906$ and
LagrangeBench reaches correlation $0.996$. Because Darcy improves pattern
correlation to $0.980$ but overpredicts the amplitude of that first example,
the atlas uses the held-out example nearest the median Phi-CMR error rather than the
first or best example. The remaining smoothing of sharp Navier--Stokes
vortical features and the Darcy amplitude mismatch identify forward-resolution
and calibration limits that are not explained by the backward response law.

The unusual Diffusion--Reaction first step remains a stress test because, from
$t_0$ to $t_1$, its two fields have persistence errors of $8.31$ and $20.66$
and input--target correlations of only $0.23$ and $0.10$, whereas the separately
evaluated $t_1\!\to t_2$ transient window has correlations near $0.94$.
A $64^2$, per-channel normalized, four-layer width-32 residual adapter reaches
mean Phi-CMR error $0.2633$, representative-field correlation $0.964$, and
amplitude ratio $0.972$ on the latter window, which is excluded from the paired
capacity ratio in Table~\ref{tab:silva-capacity-results} because it represents
a different physical-time task.

Figures~\ref{fig:silva-capacity-darcy}--\ref{fig:silva-capacity-lagrange}
show held-out fields or particles for higher-capacity models using the same
five-seed all-method aggregate and denominator/pole diagnostics as the compact
atlases, with panels (a--e) selecting under a fixed evaluation seed the example
whose Phi-CMR error is nearest the median of 16 held-out examples.

\begin{figure}[p]
\centering
\includegraphics[width=.88\textwidth]{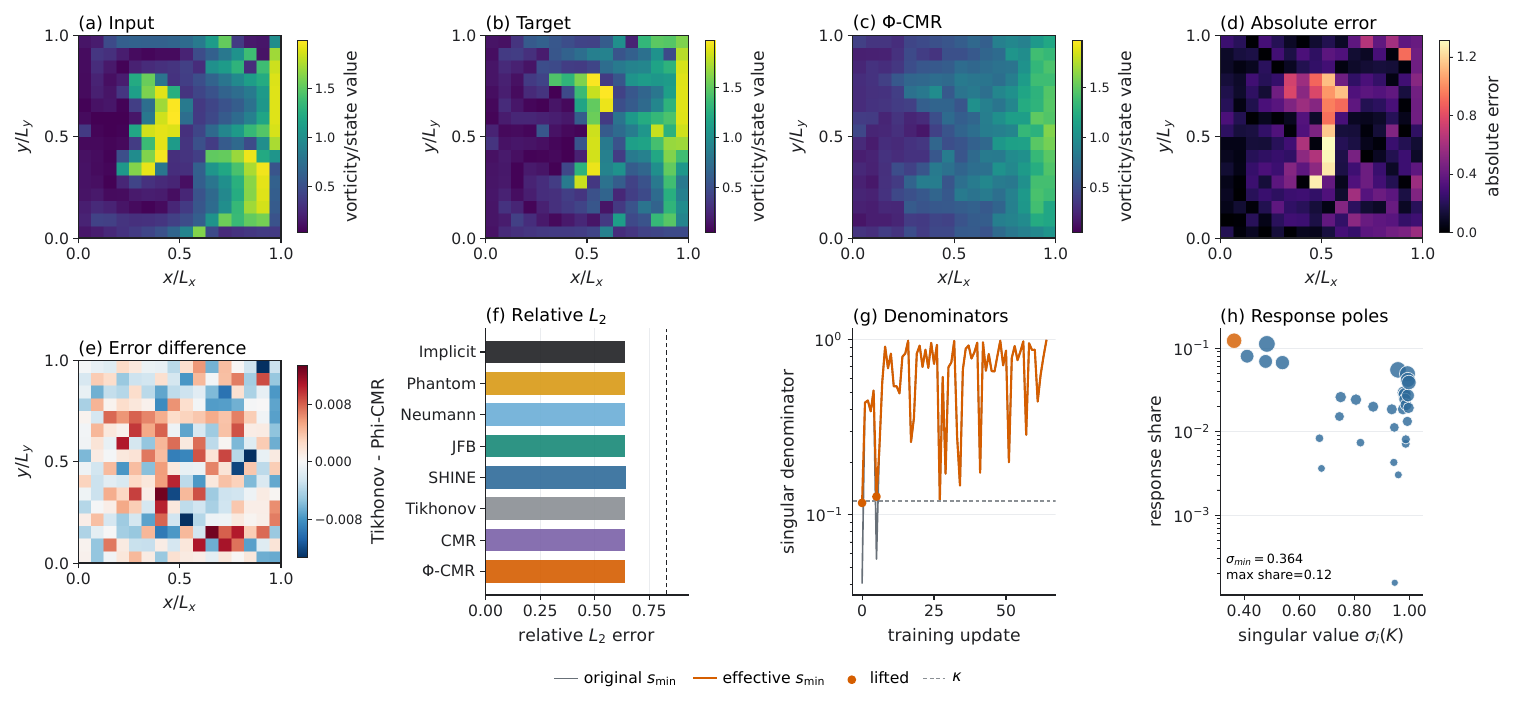}
\caption{\silvaatlascaption{PDEArena Navier--Stokes 2D}{Field panels display
the first vorticity/state channel.}}
\label{fig:silva-extension-pdearena-ns}
\end{figure}

\begin{figure}[p]
\centering
\includegraphics[width=.78\textwidth]{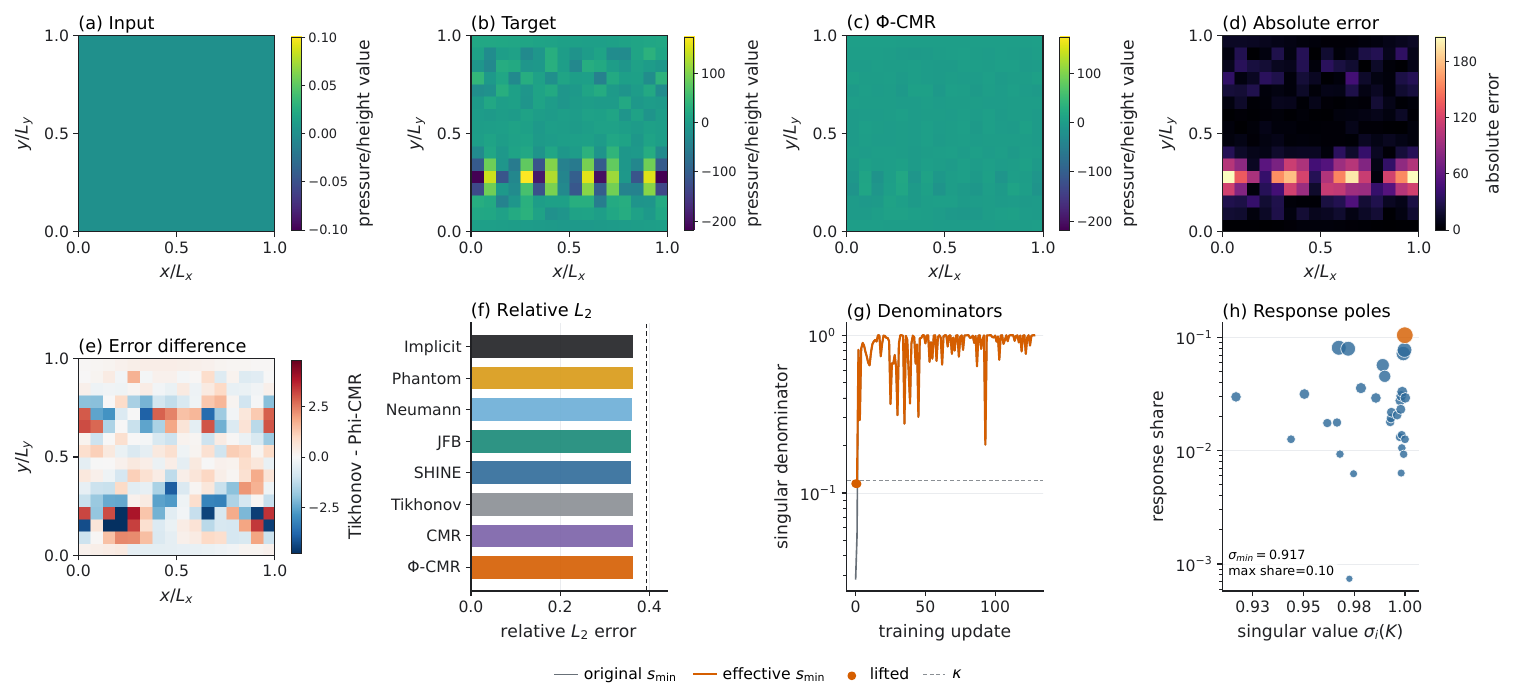}
\caption{\silvaatlascaption{PDEArena Shallow Water 2D}{Field panels display
the first pressure/height channel.}}
\label{fig:silva-extension-pdearena-sw}
\end{figure}

\begin{figure}[p]
\centering
\includegraphics[width=.78\textwidth]{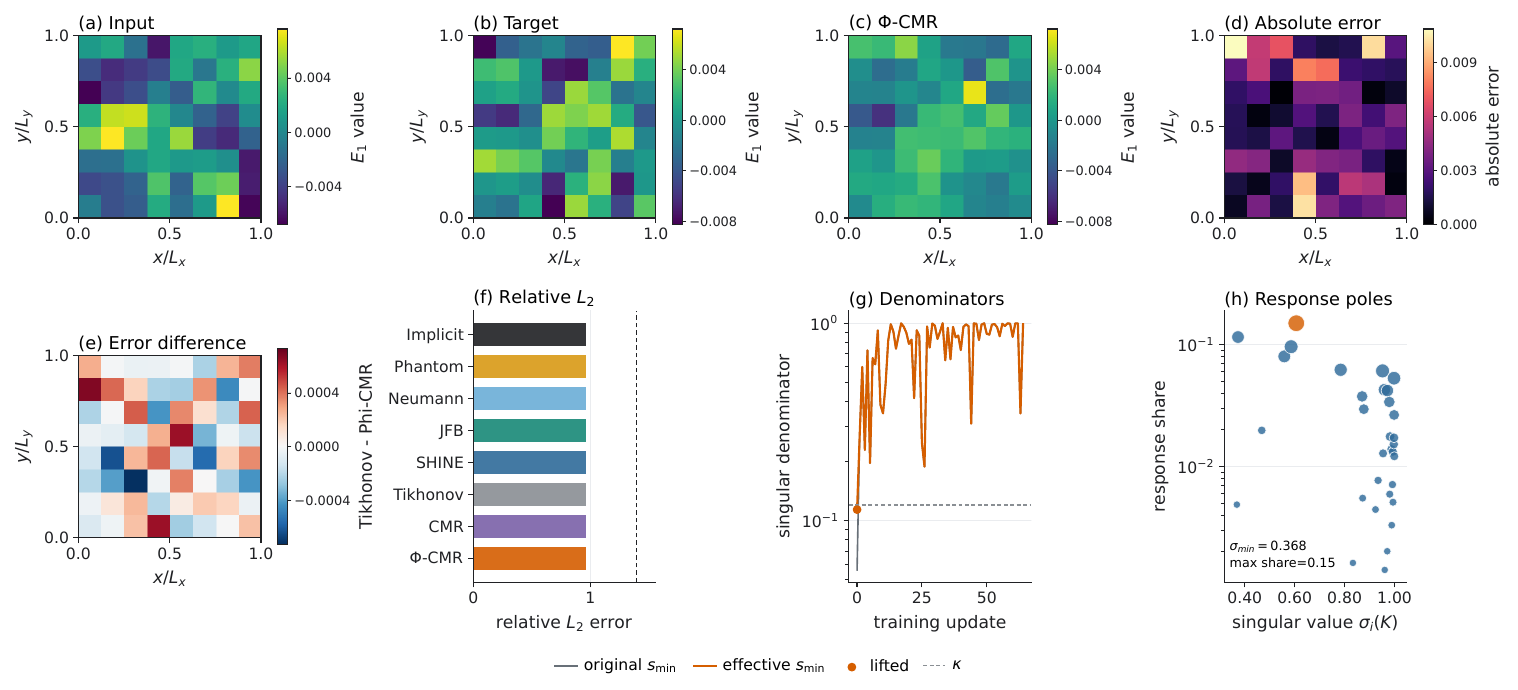}
\caption{\silvaatlascaption{PDEArena Maxwell 3D}{Field panels show the central
slice of the first electric-displacement component.}}
\label{fig:silva-extension-pdearena-maxwell}
\end{figure}

\begin{figure}[p]
\centering
\includegraphics[width=.78\textwidth]{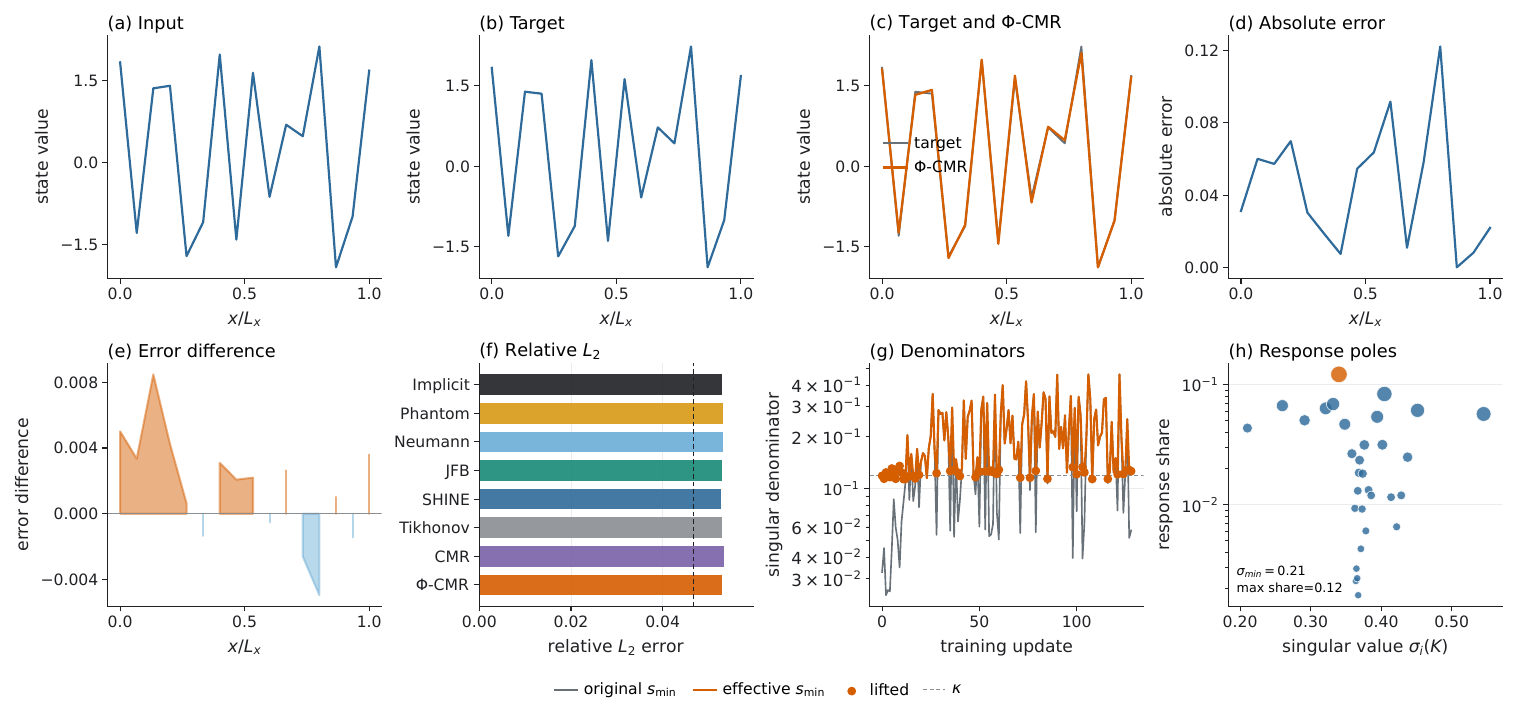}
\caption{\silvaatlascaption{PDEArena Kuramoto--Sivashinsky 1D}{Field panels
display the scalar state.}}
\label{fig:silva-extension-pdearena-ks}
\end{figure}

\begin{figure}[p]
\centering
\includegraphics[width=.78\textwidth]{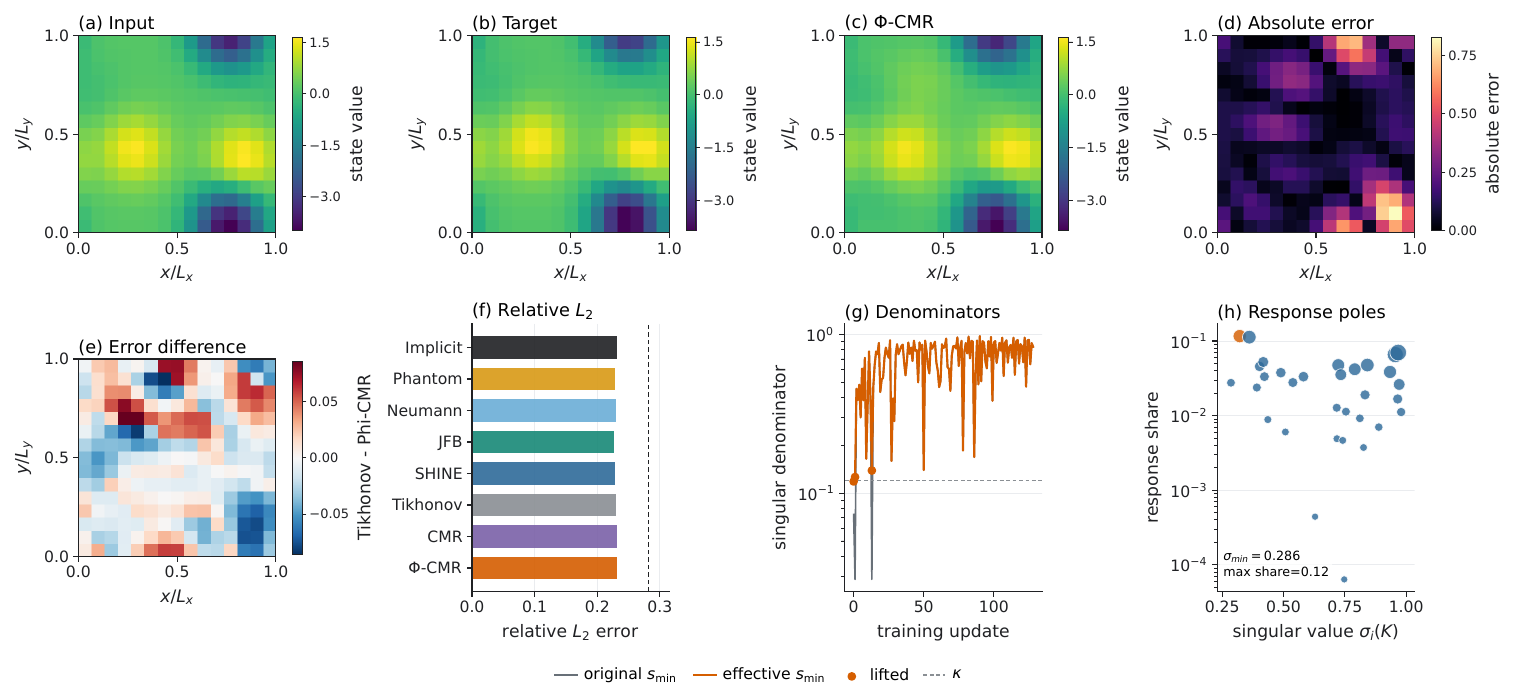}
\caption{\silvaatlascaption{DynaBench Advection}{Field panels display the
advected scalar.}}
\label{fig:silva-extension-dynabench-advection}
\end{figure}

\begin{figure}[p]
\centering
\includegraphics[width=.78\textwidth]{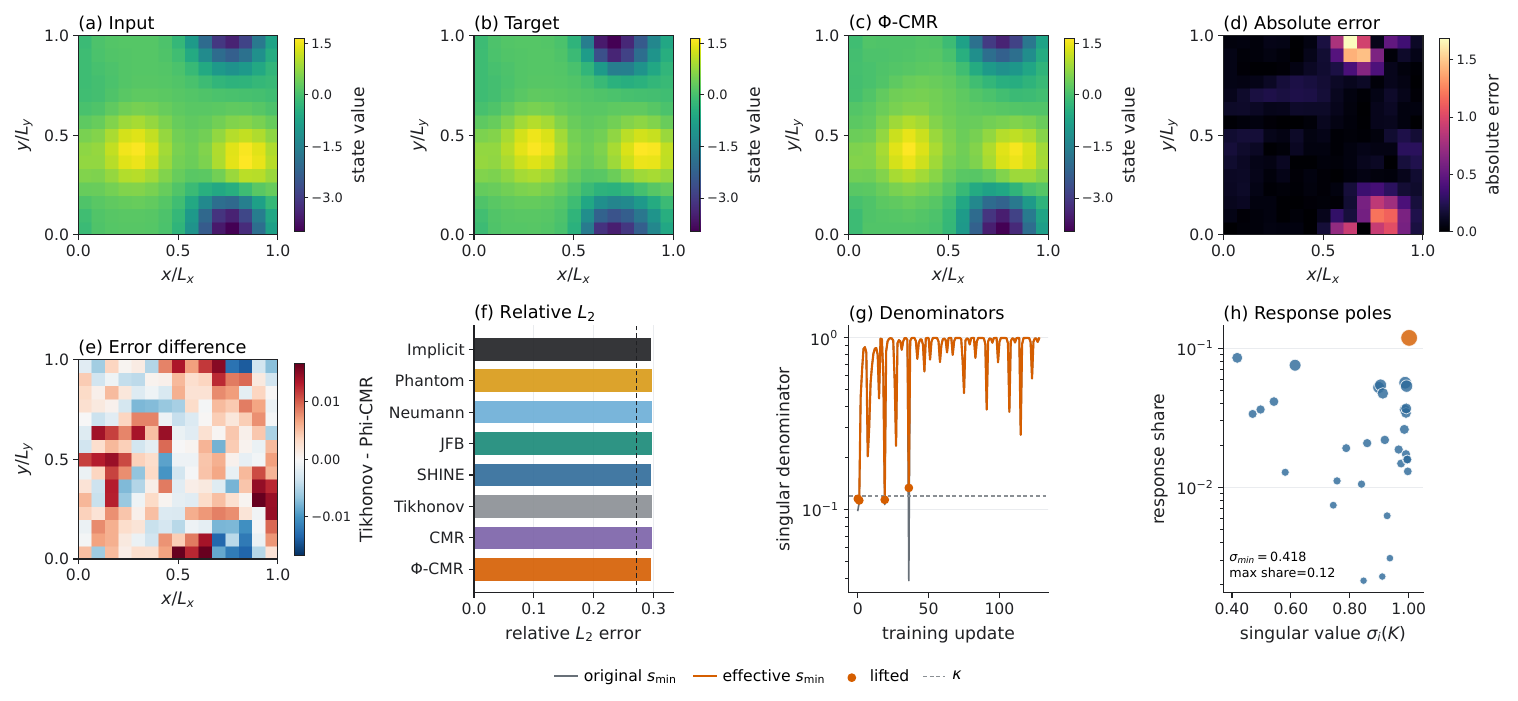}
\caption{\silvaatlascaption{DynaBench Burgers}{Field panels display the
conserved scalar.}}
\label{fig:silva-extension-dynabench-burgers}
\end{figure}

\begin{figure}[p]
\centering
\includegraphics[width=.78\textwidth]{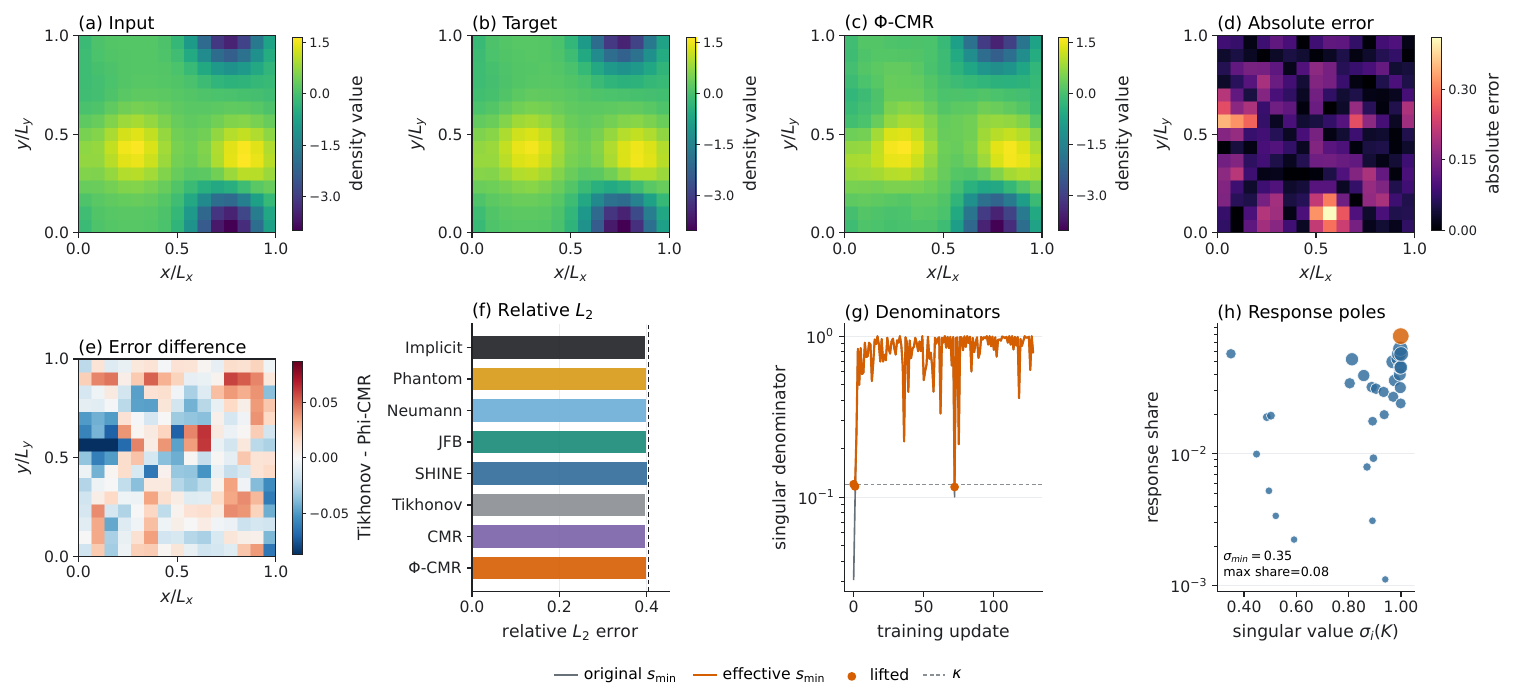}
\caption{\silvaatlascaption{DynaBench Gas Dynamics}{Field panels display
density, while panel (f) aggregates all predicted variables.}}
\label{fig:silva-extension-dynabench-gas}
\end{figure}

\begin{figure}[p]
\centering
\includegraphics[width=.78\textwidth]{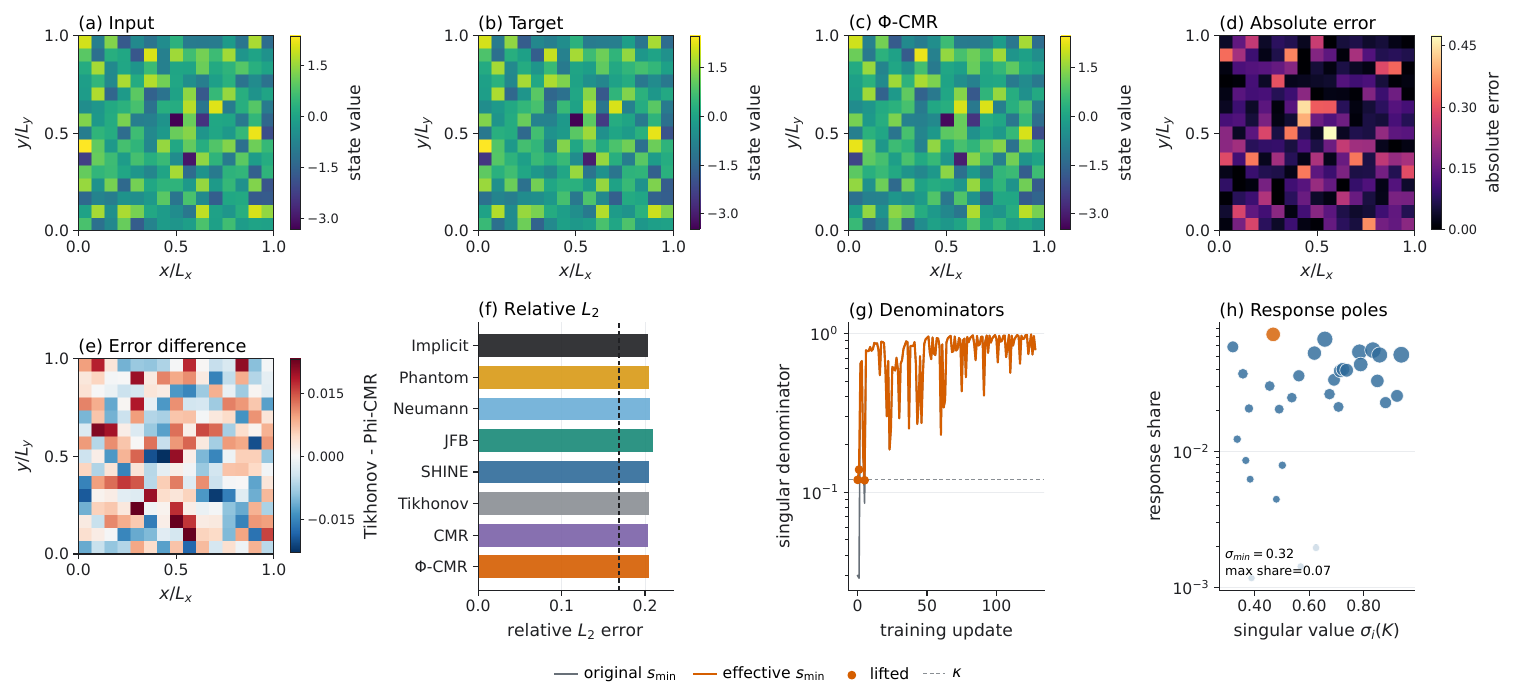}
\caption{\silvaatlascaption{DynaBench Kuramoto--Sivashinsky}{Field panels
display the scalar state for the deterministic example nearest the Phi-CMR
median error (offset 3 of 16).}}
\label{fig:silva-extension-dynabench-ks}
\end{figure}

\begin{figure}[p]
\centering
\includegraphics[width=.78\textwidth]{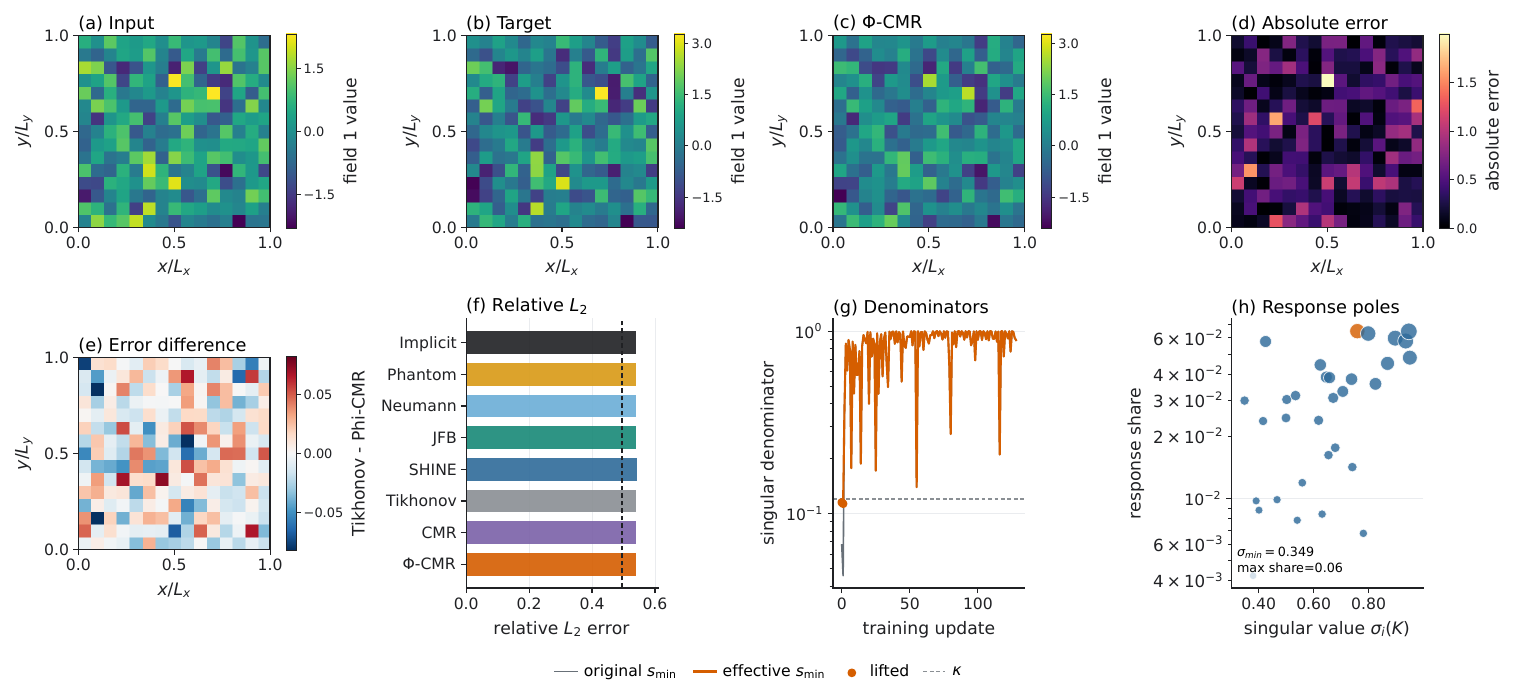}
\caption{\silvaatlascaption{DynaBench Reaction--Diffusion}{Field panels display
the first reaction--diffusion channel.}}
\label{fig:silva-extension-dynabench-rd}
\end{figure}

\begin{figure}[p]
\centering
\includegraphics[width=.78\textwidth]{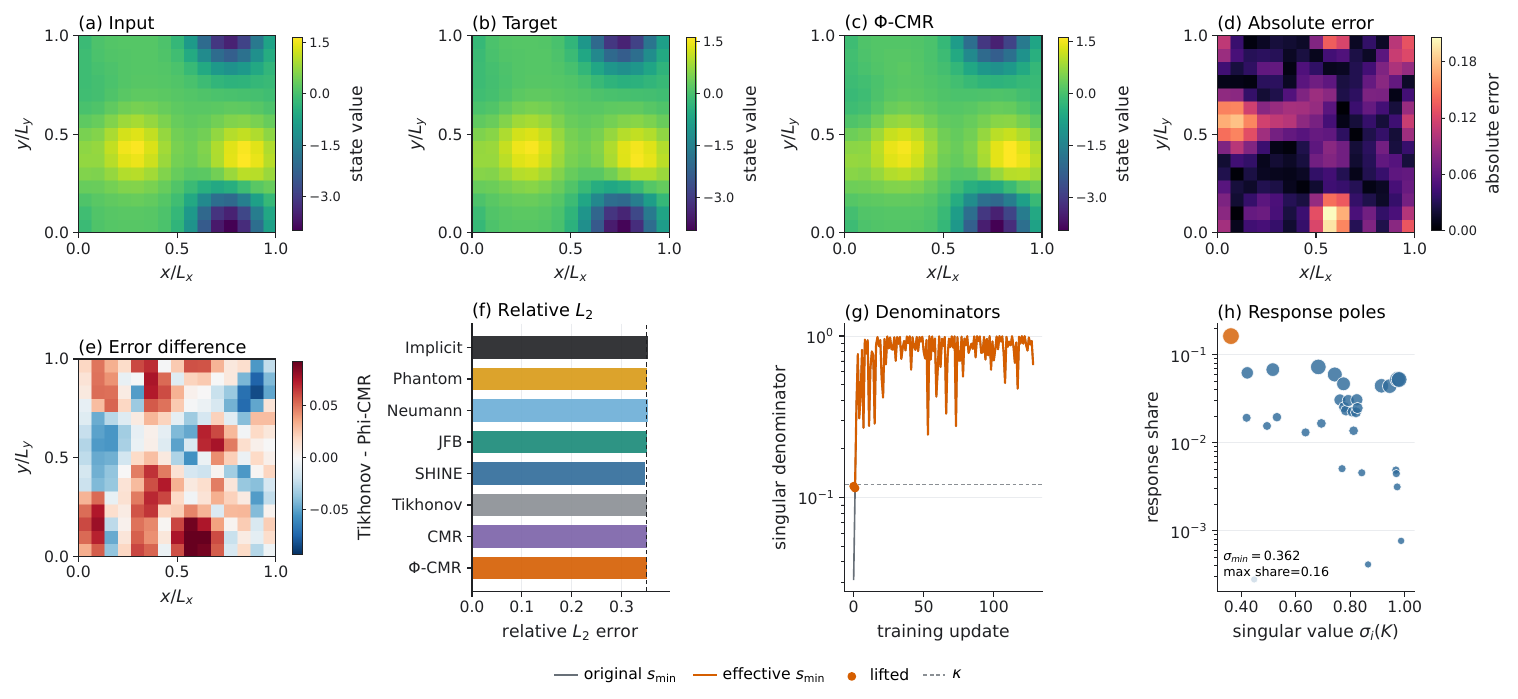}
\caption{\silvaatlascaption{DynaBench Wave}{Field panels display the first
wave-state channel.}}
\label{fig:silva-extension-dynabench-wave}
\end{figure}

\begin{figure}[p]
\centering
\includegraphics[width=.78\textwidth]{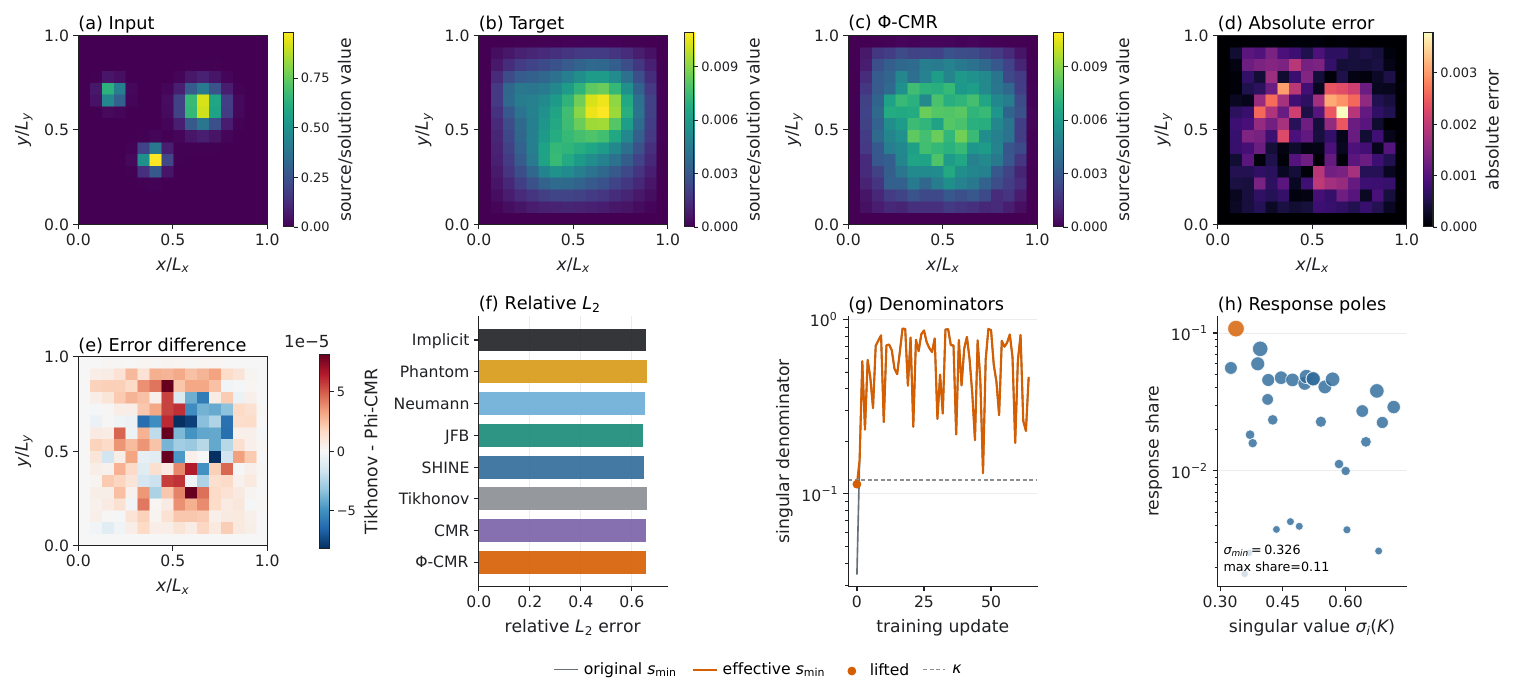}
\caption{\silvaatlascaption{PDEGym Poisson--Gauss}{Panels (a--e) show the
source-to-solution operator example.}}
\label{fig:silva-extension-pdegym}
\end{figure}

\begin{figure}[p]
\centering
\includegraphics[width=.78\textwidth]{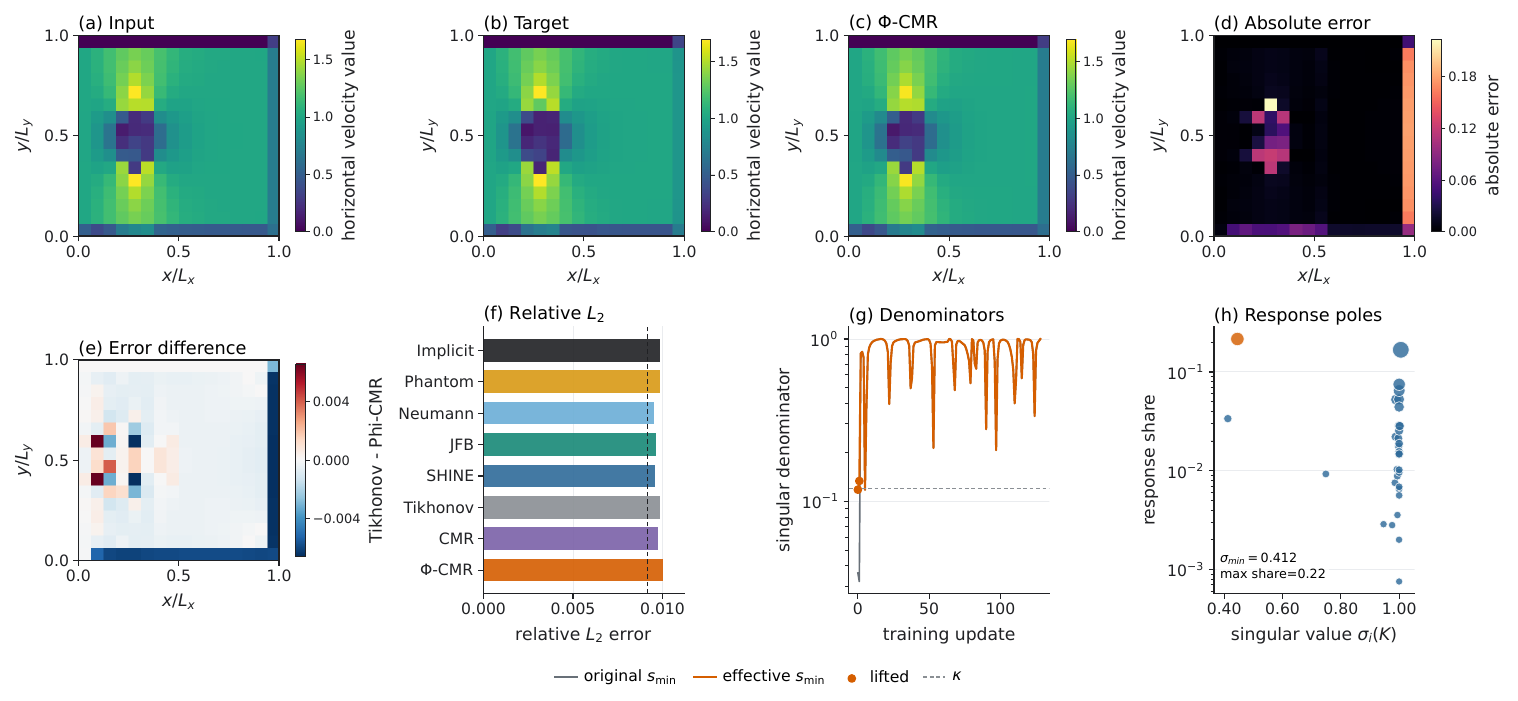}
\caption{\silvaatlascaption{CFDBench cylinder geometry}{Field panels display
velocity for a held-out geometry case.}}
\label{fig:silva-extension-cfdbench}
\end{figure}

\begin{figure}[!htbp]
\centering
\includegraphics[width=.78\textwidth]{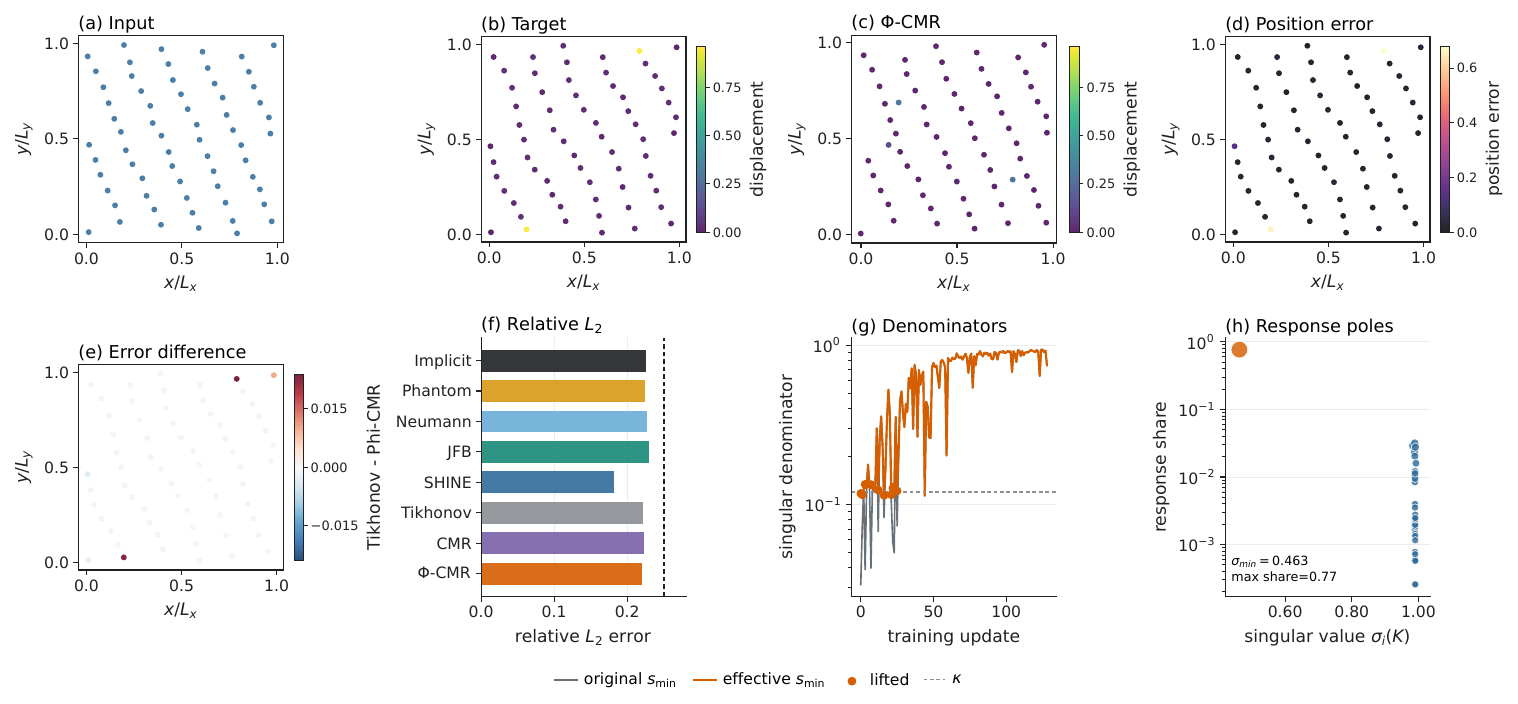}
\caption{\silvaatlascaption{LagrangeBench Taylor--Green vortex 2D}{Panels
(a--e) retain the native particle coordinates, and panel (d) is position error.}}
\label{fig:silva-extension-lagrange}
\end{figure}

Figures~\ref{fig:silva-extension-pdearena-ns}--\ref{fig:silva-extension-lagrange}
use a fixed evaluation seed and the first held-out example, except for
DynaBench Kuramoto--Sivashinsky, which uses the deterministic median-error
example. They retain panels (a--h) from
Appendix~\ref{app:silva-family-atlas}, with Maxwell shown as a central slice and
LagrangeBench in native particle coordinates.

\begingroup
\linespread{0.86}\selectfont
\section{Scope and Validity Conditions}
\label{app:extended-discussion-boundary}

The exact mathematical statement is narrower than a generic training-stability
claim because, for \(K=I-J=U\Sigma V^\top\), parameter \(a\) receives modal
contribution \((v_i^\top g)(u_i^\top B_a)/\sigma_i\), requiring both the loss
source and parameter source. CMR changes a selected denominator and is
therefore exact only for the explicitly modified, frozen-anchor residual,
whereas its application at the original equilibrium is a biased backward surrogate. The
experiments evaluate finite response, selectivity, and predictive fidelity
under conditions in which exact implicit differentiation remains numerically
available as a reference, allowing each response modification to be measured
against a resolved adjoint.

The dense rule requires a resolved critical singular subspace because Euclidean
singular modes depend on state coordinates, and near a repeated singular cluster
only the subspace, not individual vectors, is identifiable. We therefore fix
representation and train-only normalization in every comparison, interpret
threshold crossings within numerical tolerance as ambiguous, and recommend
an unresolved interval \([\kappa-\eta,\kappa+\eta]\) with principal-angle
tracking for repeated clusters, as formalized by the coordinate-consistent
metric in Appendix~\ref{app:coordinate-metric}.

The SILVA specialization instead modifies singular values of the collective
matrix \(\Gamma=I-B^\top K_L^{-1}A\), which need not equal those of the full
\(K\), and therefore assumes accurate local solves and an informative low-rank
factorization. Raw norms of \(A,B\) are gauge dependent under
\(A\mapsto AQ\), \(B\mapsto BQ^{-\top}\), which motivates an SVD-balanced
factorization for diagnostics, while neither the dense nor collective rule repairs poor forward
convergence, architecture mismatch, unresolved subspaces, or rapidly changing
modes.

Finally, the error-ratio threshold of 1.05 is a descriptive upper threshold
rather than a noninferiority or equivalence margin, and the five seeds are independent model
repetitions, whereas updates, horizons, channels, and examples are repeated
measurements. Field errors are computed as inverse-standardized sample-relative
Euclidean errors rather than conservation errors or PDE residuals. Accordingly, the
quantitative conclusions concern response fidelity, selective activation,
convergence, and cost at the evaluated sizes under matched compact
architectures. Governing-equation residuals, invariant
preservation, asymptotic scaling, and architecture-matched comparisons with
monotone, spectral, or reversible DEQ designs require dedicated evaluations
with the corresponding forward models.

Forward parameterization uses training data only, with families whose
training-split persistence error is at most \(0.65\) predicting
\(\widehat u_{t+1}=u_t+r_\theta(u_t)\) and the others using a direct readout.
Shared normalization makes \(r_\theta=0\) recover persistence at initialization,
while projection after each update enforces \(\lVert W\rVert_2\leq0.995\). Consequently,
\(J=\operatorname{diag}(\operatorname{sech}^2(\cdot))W\) satisfies
\(\lVert J\rVert_2\leq0.995\) and \(\sigma_{\min}(I-J)\geq0.005\). The forward
map is therefore contractive while inducing near-critical response, and static
operator maps use direct parameterization. All methods share architecture,
initialization, data order, equilibrium solver, and numerical tolerances, which
allows the atlas to compare predictions within a matched forward design even
though the backward rule may produce distinct trained parameters and equilibria.

\endgroup

\FloatBarrier

\begin{figure}[H]
\centering
\includegraphics[width=.58\textwidth]{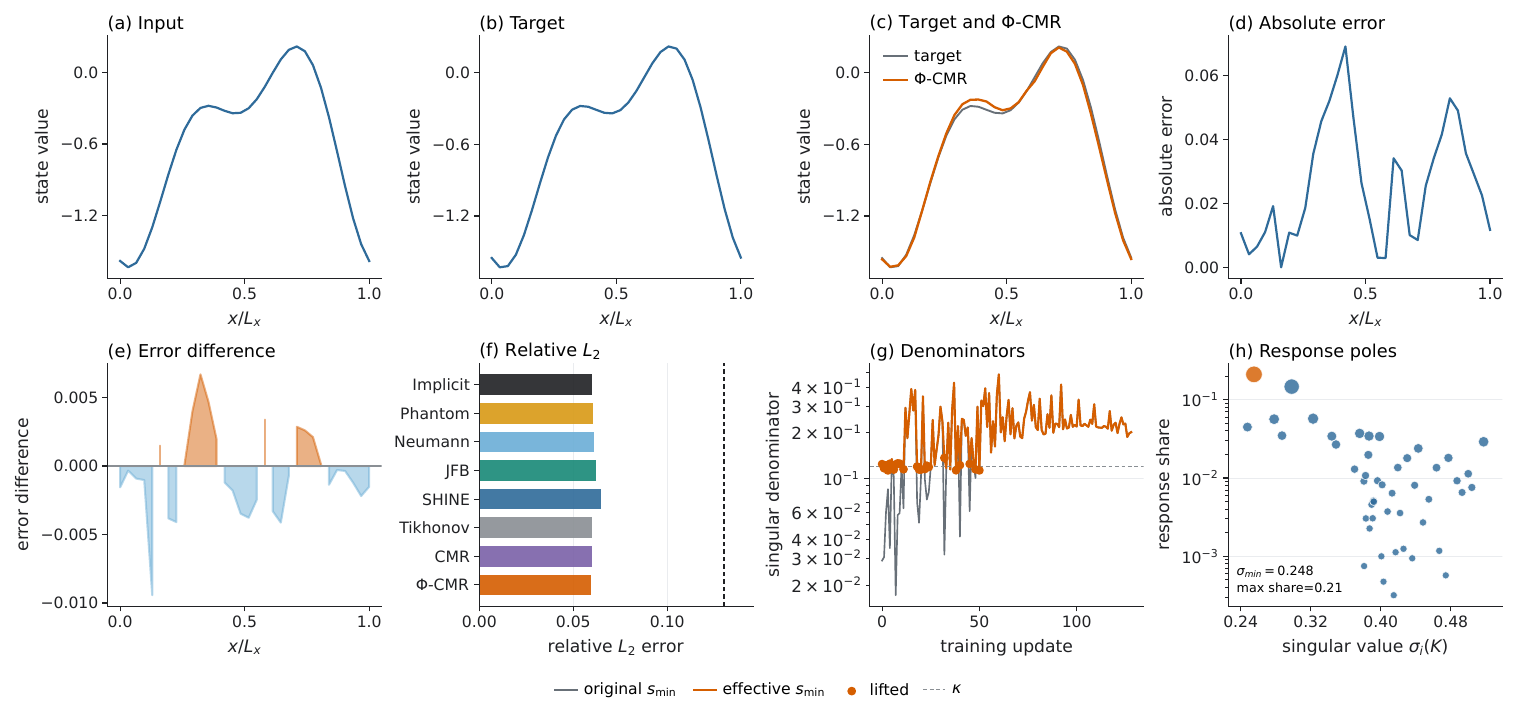}
\footnotesize
\caption{\silvaatlascaption{Advection 1D}{Field panels display the scalar state.}}
\label{fig:silva-atlas-advection}
\end{figure}

\begin{figure}[H]
\centering
\includegraphics[width=.58\textwidth]{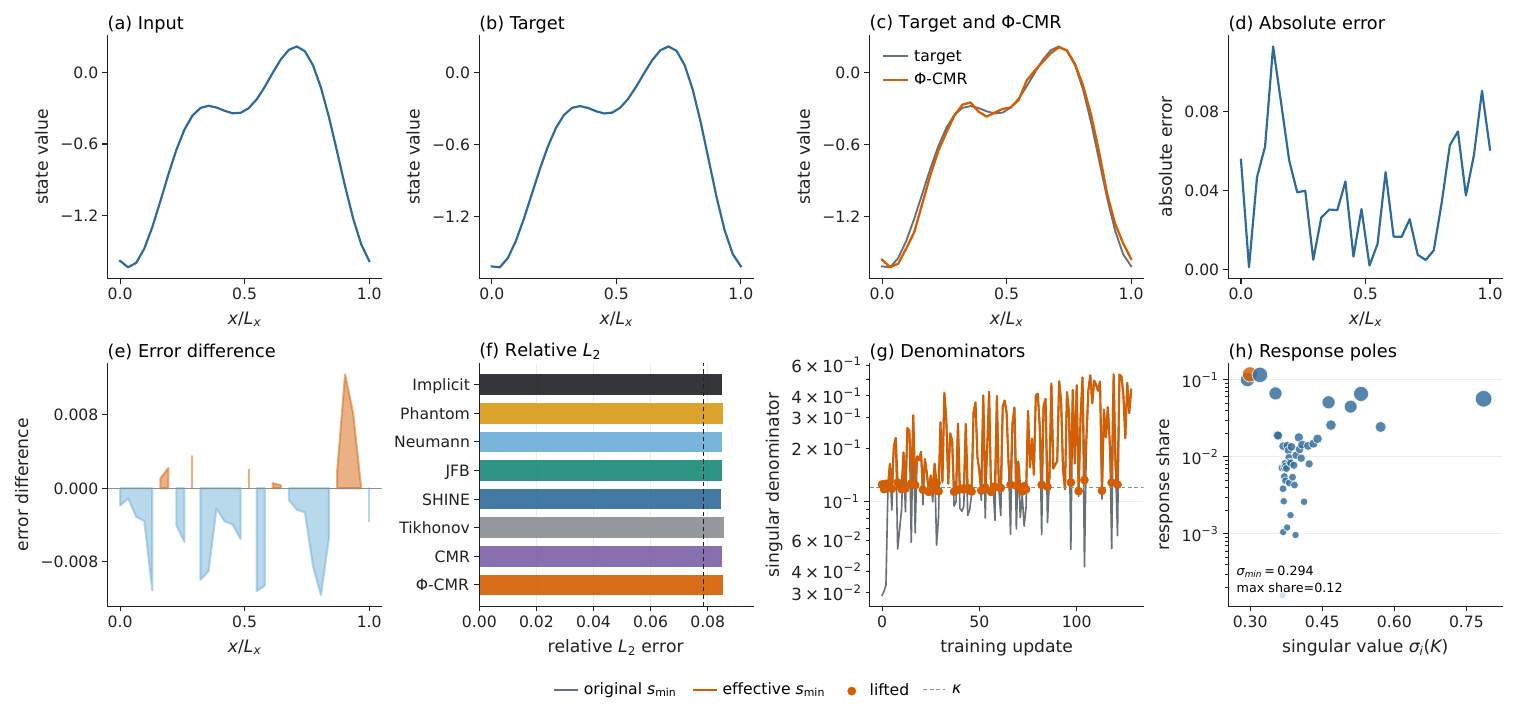}
\footnotesize
\caption{\silvaatlascaption{Burgers 1D}{Field panels display the scalar state.}}
\label{fig:silva-atlas-burgers}
\end{figure}

\begin{figure}[p]
\centering
\includegraphics[width=.78\textwidth]{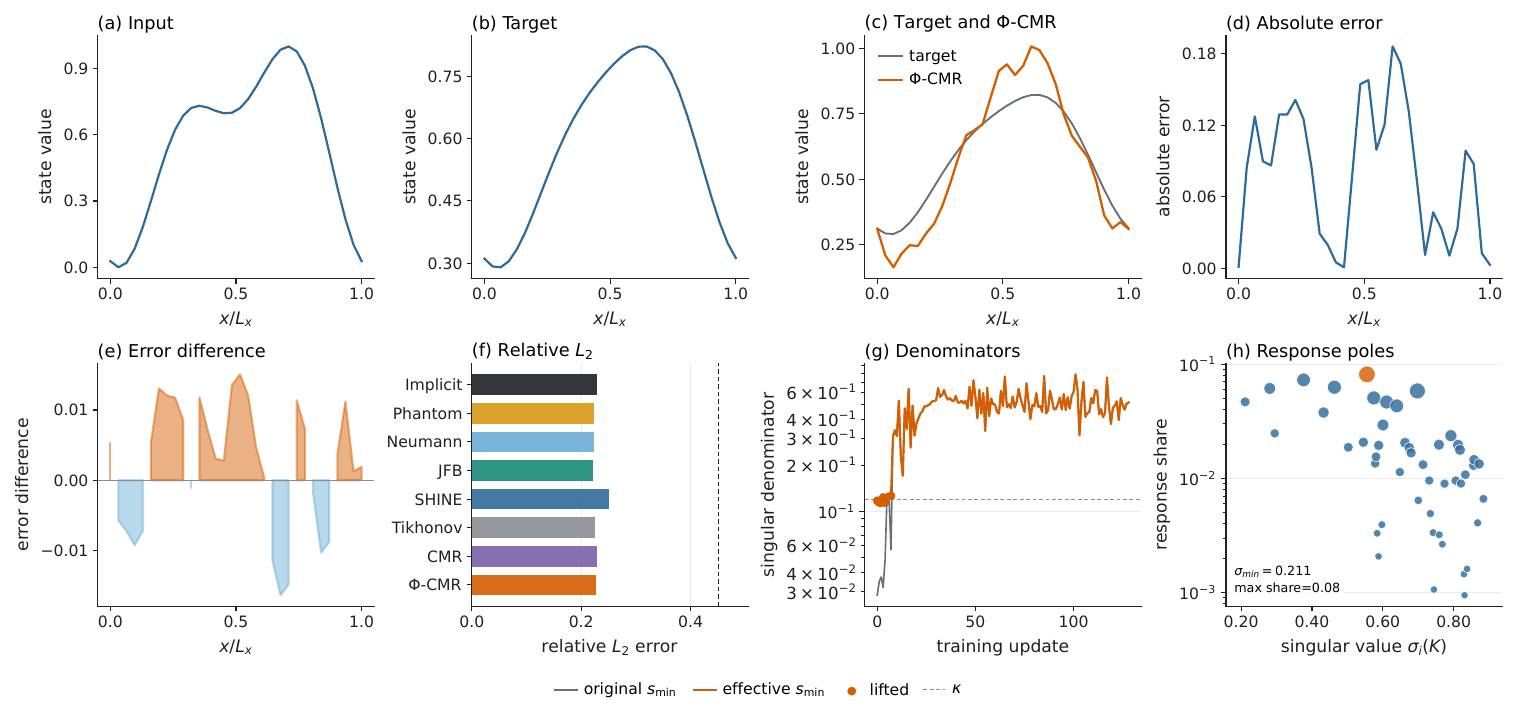}
\caption{\silvaatlascaption{Reaction--Diffusion 1D}{Field panels display the first state channel.}}
\label{fig:silva-atlas-reaction-diffusion}
\end{figure}

\begin{figure}[p]
\centering
\includegraphics[width=.78\textwidth]{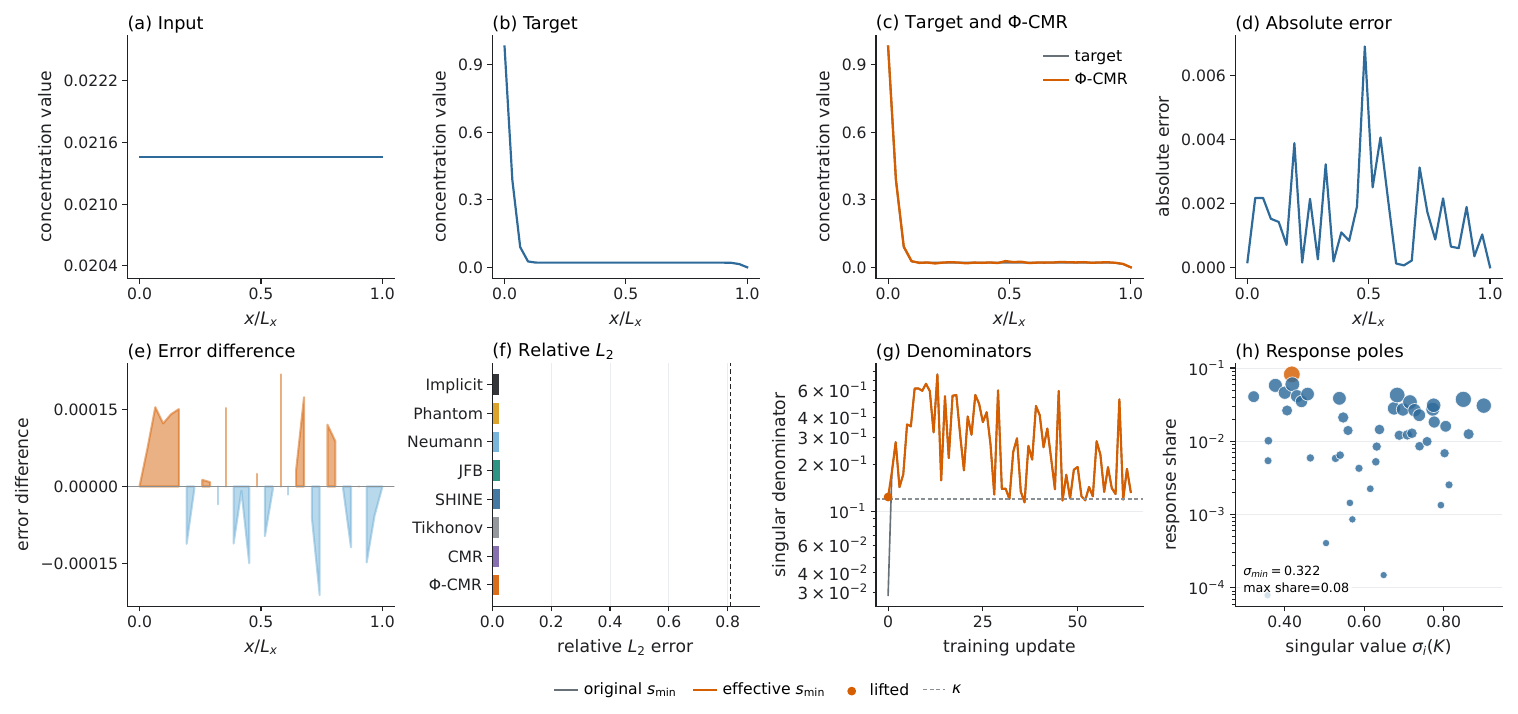}
\caption{\silvaatlascaption{Diffusion--Sorption 1D}{Field panels display concentration.}}
\label{fig:silva-atlas-diffusion-sorption}
\end{figure}

\begin{figure}[p]
\centering
\includegraphics[width=.78\textwidth]{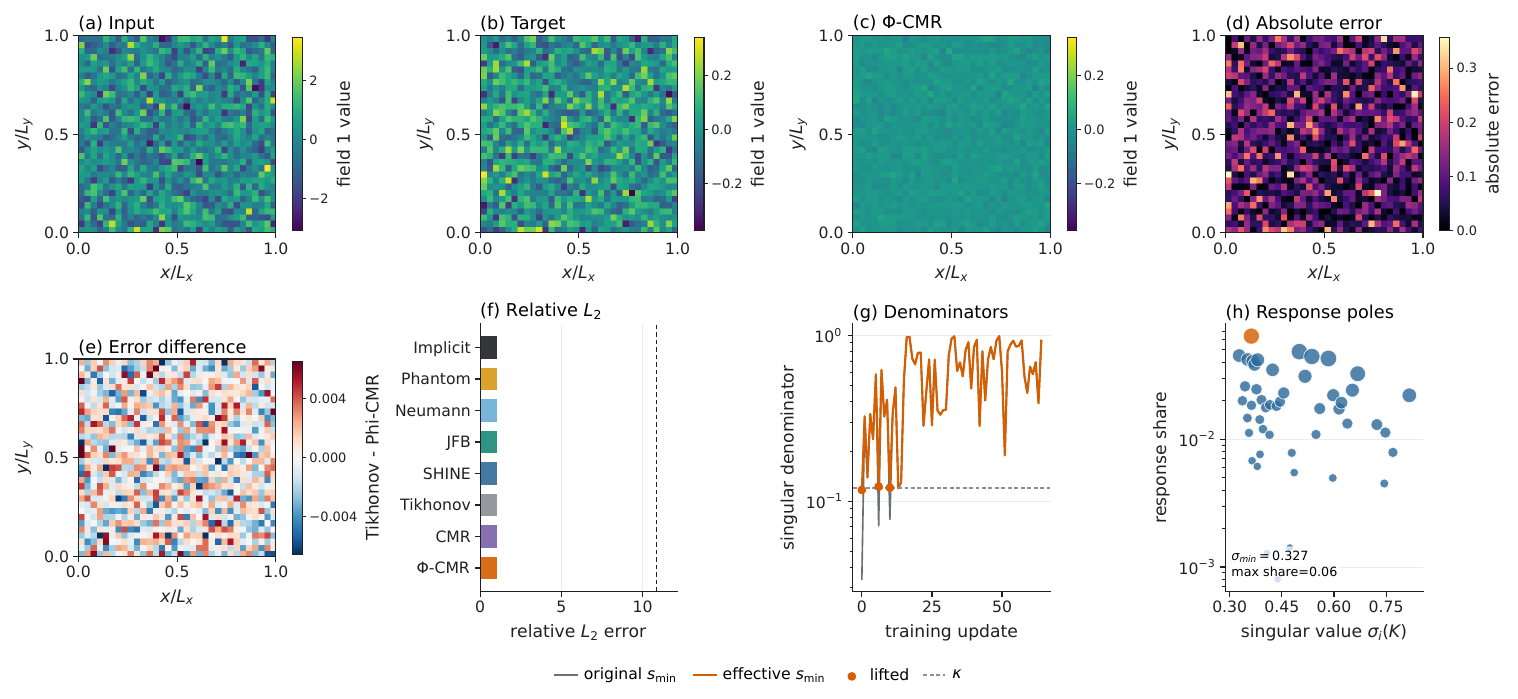}
\caption{\silvaatlascaption{Diffusion--Reaction 2D}{Field panels display the first physical channel.}}
\label{fig:silva-atlas-diffusion-reaction}
\end{figure}

\begin{figure}[p]
\centering
\includegraphics[width=.78\textwidth]{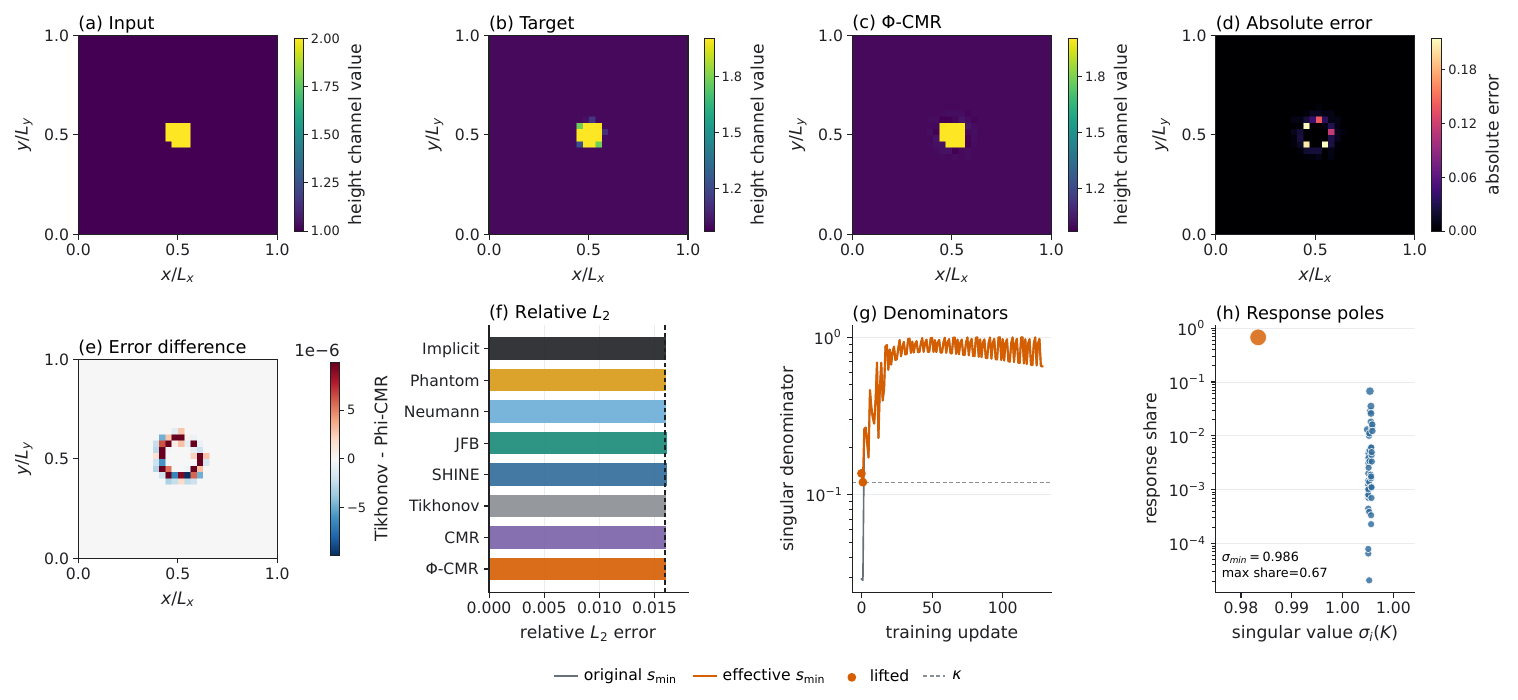}
\caption{\silvaatlascaption{Shallow Water 2D}{Field panels display fluid height.}}
\label{fig:silva-atlas-shallow-water}
\end{figure}

\begin{figure}[p]
\centering
\includegraphics[width=.78\textwidth]{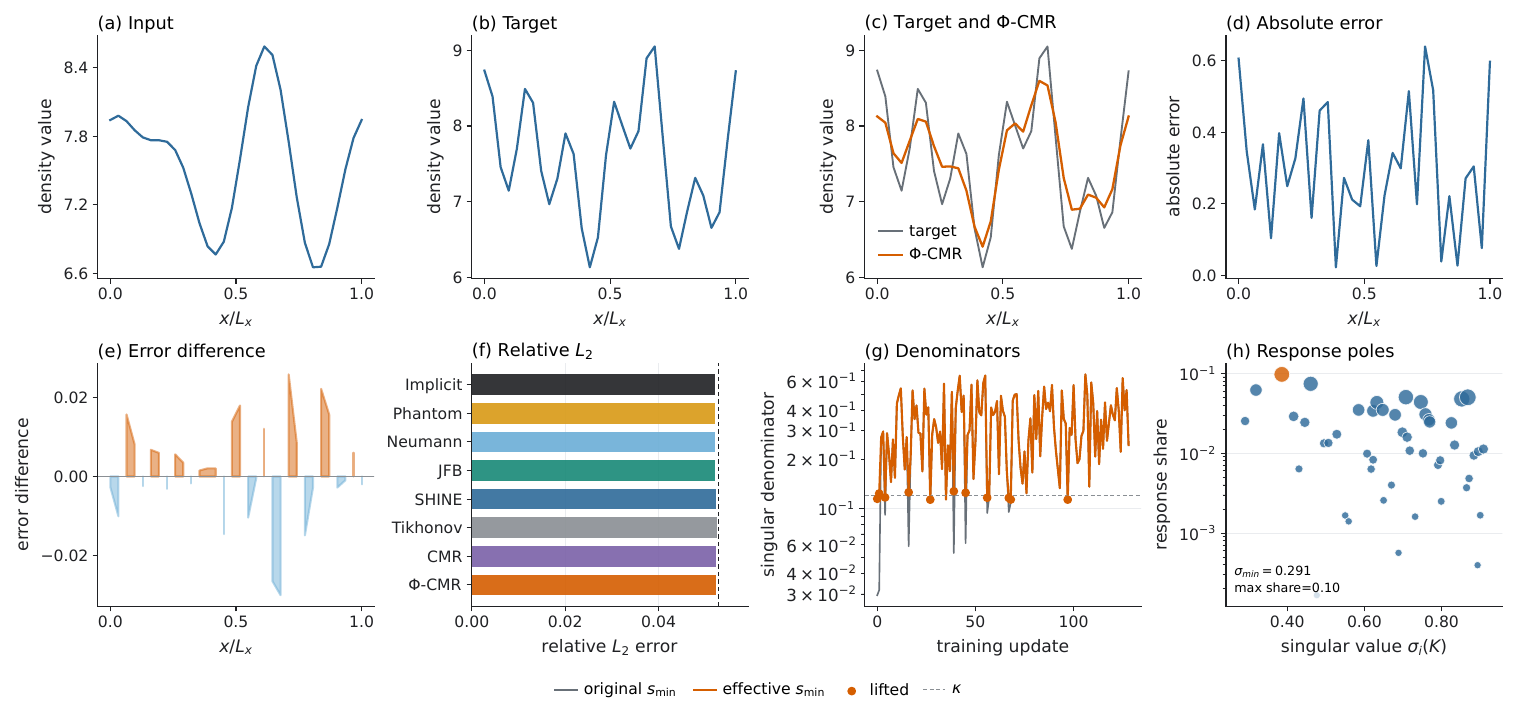}
\caption{\silvaatlascaption{Compressible CFD 1D}{Field panels display density,
while panel (f) aggregates density, pressure, and velocity.}}
\label{fig:silva-atlas-compressible-cfd}
\end{figure}

\begin{figure}[p]
\centering
\includegraphics[width=.78\textwidth]{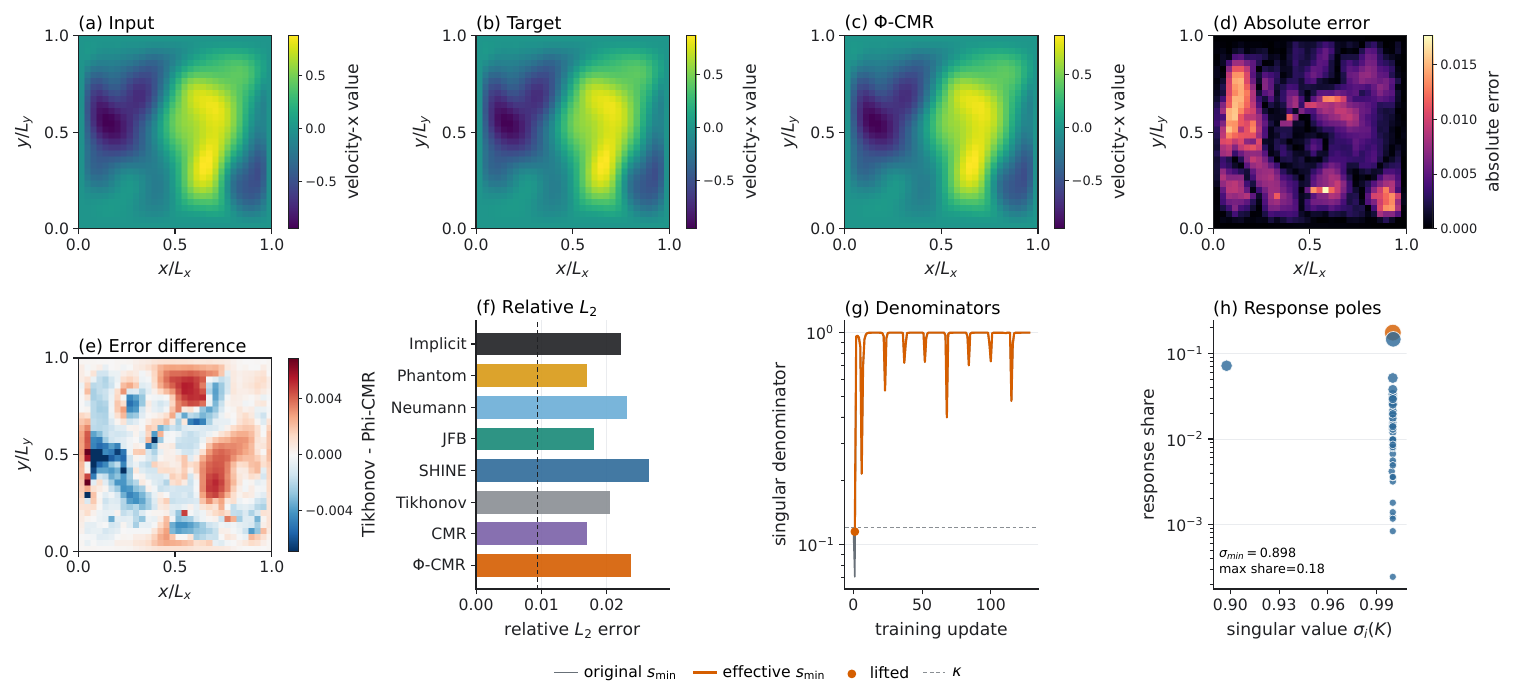}
\caption{\silvaatlascaption{Incompressible Navier--Stokes 2D}{Field panels
display horizontal velocity.}}
\label{fig:silva-atlas-navier-stokes}
\end{figure}

\begin{figure}[H]
\centering
\includegraphics[width=.78\textwidth]{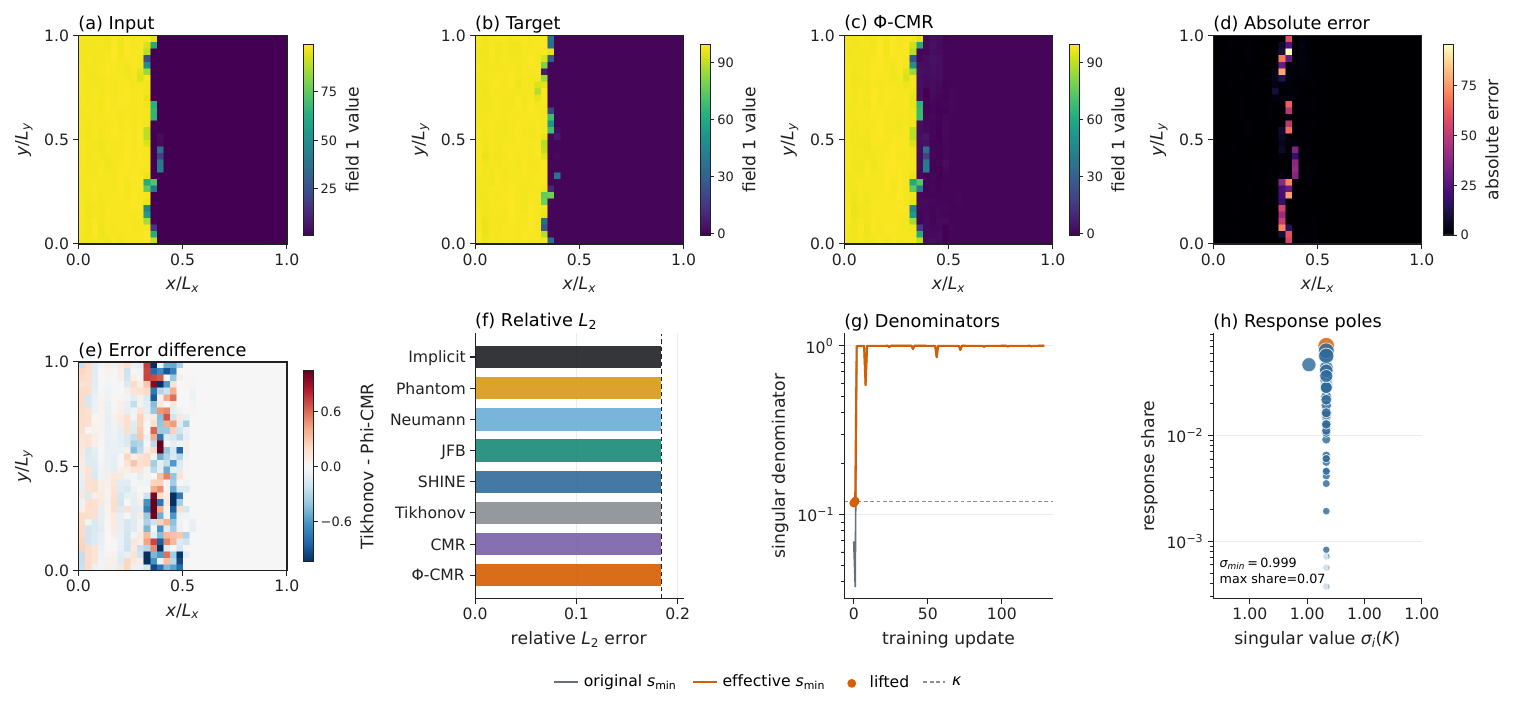}
\caption{\silvaatlascaption{The Well turbulent radiative layer 2D}{Field panels
display the first physical channel from the held-out split.}}
\label{fig:silva-atlas-the-well}
\end{figure}

\begin{figure}[H]
\centering
\includegraphics[width=.78\textwidth]{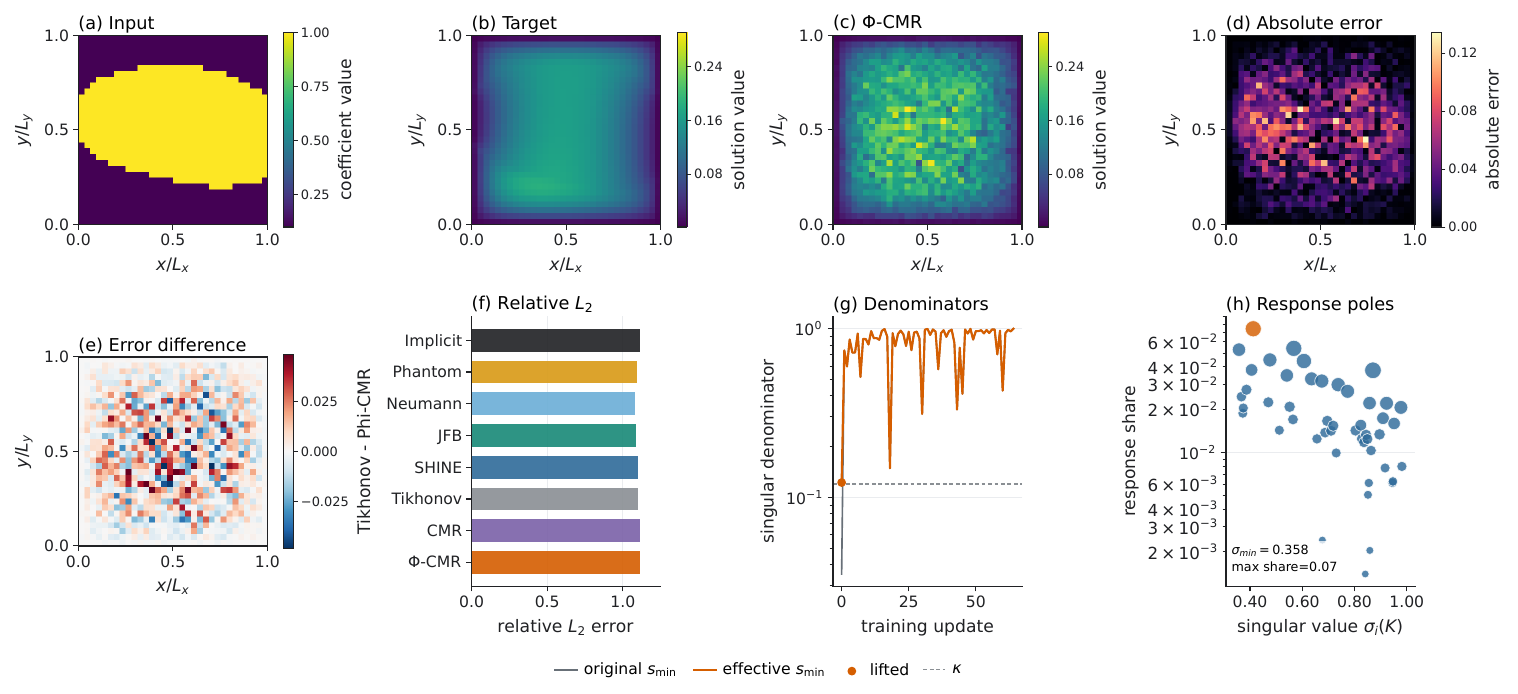}
\caption{\silvaatlascaption{Darcy 2D}{Panels (a--e) use a fixed $\beta=1$
coefficient-to-solution example, while panel (f) aggregates five Darcy regimes.}}
\label{fig:silva-atlas-darcy}
\end{figure}

\begingroup
\linespread{0.94}\selectfont
\section{Native SILVA Per-Family Field and Pole Atlas}
\label{app:silva-family-atlas}

Figures~\ref{fig:silva-atlas-advection}--\ref{fig:silva-atlas-darcy} connect
aggregate error ratios to recognizable physical predictions and the proposed
response mechanism, with each visualization using a fixed seed and an a priori
held-out example. Panels (a--e) show the first physical channel as input or
initial condition, target, Phi-CMR prediction, absolute error, and pointwise
Tikhonov-minus-Phi-CMR error, with positive values in panel (e) favoring
Phi-CMR, while panel (f) displays the all-channel, five-seed aggregate for all
eight methods normalized by exact implicit differentiation and, for Darcy,
also averages five conditioning regimes while the displayed field uses
$\beta=1$.

Panel (g) traces the measured and effective Phi-CMR collective denominators,
the cutoff $\kappa$, and active-lift updates, while panel (h) evaluates the normalized
modal contribution $|v_i^\top g|/\sigma_i$ of the final 48-dimensional hidden
residual $K=I-J$ for the same held-out source. The associated adjoint
amplification is determined by the loss-source projection onto each mode,
while training traces activate the collective lift in every family even when a
displayed final source does not. Panel (g) therefore resolves the SILVA
mechanism and panel (h) resolves full-state source visibility.

\begin{table}[H]
\centering
\scriptsize
\setlength{\tabcolsep}{5pt}
\caption{Physical systems and displayed quantities in the native SILVA
per-family analysis. Darcy is evaluated as an elliptic coefficient-to-solution
map, while the remaining entries are one-step dynamical predictions.}
\label{tab:silva-family-atlas-coverage}
\begin{tabular}{@{}lll@{}}
\toprule
Physical family & Mathematical setting & Displayed quantity\\
\midrule
Advection 1D & linear transport & state\\
Burgers 1D & nonlinear conservation law & state\\
Reaction--Diffusion 1D & coupled reaction and diffusion & state\\
Diffusion--Sorption 1D & reactive transport & concentration\\
Diffusion--Reaction 2D & coupled parabolic system & first field\\
Shallow Water 2D & hyperbolic free-surface flow & height\\
Compressible CFD 1D & compressible flow & density\\
Incompressible Navier--Stokes 2D & incompressible flow & horizontal velocity\\
The Well radiative layer 2D & turbulent radiation hydrodynamics & first field\\
Darcy 2D & elliptic operator map & coefficient / solution\\
\bottomrule
\end{tabular}
\end{table}

\endgroup

\FloatBarrier

\end{document}